\documentclass{article}

\def\titleprefix{}
\def\titlerunningsuffix{}

\usepackage{microtype}
\usepackage{graphicx}
\usepackage{subfigure}
\usepackage{booktabs}
\usepackage{hyperref}
\usepackage{datatool}
\usepackage{siunitx}

\usepackage{amsfonts}
\usepackage{amsmath}
\usepackage{amssymb}
\usepackage{bbm}
\usepackage{bm}

\usepackage[
   accepted
]{icml2021}

\icmltitlerunning{Learning a Universal Template for Few-shot Dataset Generalization \titlerunningsuffix}

\newcommand{\ra}[1]{\renewcommand{\arraystretch}{#1}}

\newcommand*\dtlformat[1]{\DTLifnumerical{#1}{\num{#1}}{#1}}

\DeclareMathOperator*{\argmax}{argmax}
\DeclareMathOperator*{\softmax}{softmax}

\newcommand{\blend}{\textsc{BLEND}}

\DTLloaddb[noheader=true,omitlines=1]{example_table}{data/example_table.csv}
\DTLloaddb[noheader=true,omitlines=1]{all}{data/all.csv}
\DTLloaddb[noheader=true,omitlines=1]{ablations}{data/ablations.csv}

\DTLloaddb[noheader=true,omitlines=1]{all_correct_TS}{data/all_correct_TS.csv}
\DTLloaddb[noheader=true,omitlines=1]{ablations_correct_TS}{data/ablations_correct_TS.csv}
\DTLloaddb[noheader=true,omitlines=1]{more_runs_inits}{data/more_runs_inits.csv}
\DTLloaddb[noheader=true,omitlines=1]{more_runs_inits_correct_TS}{data/more_runs_inits_correct_TS.csv}
\DTLloaddb[noheader=true,omitlines=1]{more_runs_ftune}{data/more_runs_ftune.csv}
\DTLloaddb[noheader=true,omitlines=1]{more_runs_ftune_correct_TS}{data/more_runs_ftune_correct_TS.csv}
\DTLloaddb[noheader=true,omitlines=1]{traffic_signs_shuffled}{data/traffic_signs_shuffled.csv}
\DTLloaddb[noheader=true,omitlines=1]{traffic_signs_unshuffled}{data/traffic_signs_unshuffled.csv}
\DTLloaddb[noheader=true,omitlines=1]{compare_with_sur}{data/compare_with_sur.csv}
\DTLloaddb[noheader=true,omitlines=1]{compare_with_sur_correct_TS}{data/compare_with_sur_correct_TS.csv}
\DTLloaddb[noheader=true,omitlines=1]{hard_blender}{data/hard_blender.csv}
\DTLloaddb[noheader=true,omitlines=1]{hard_blender_correct_TS}{data/hard_blender_correct_TS.csv}

\begin{document}
\begin{filecontents*}{data/example_table.csv}
TASK                   ,  avg1,  std1,  avg2,  std2
ImageNet               , 00.00, 00.00, 00.00, 00.00
Omniglot               , 00.00, 00.00, 00.00, 00.00
Aircraft               , 00.00, 00.00, 00.00, 00.00
Birds                  , 00.00, 00.00, 00.00, 00.00
Textures               , 00.00, 00.00, 00.00, 00.00
Quick Draw             , 00.00, 00.00, 00.00, 00.00
Fungi                  , 00.00, 00.00, 00.00, 00.00
VGG Flower             , 00.00, 00.00, 00.00, 00.00
Traffic Sign           , 00.00, 00.00, 00.00, 00.00
MSCOCO                 , 00.00, 00.00, 00.00, 00.00
MNIST                  , 00.00, 00.00, 00.00, 00.00
CIFAR10                , 00.00, 00.00, 00.00, 00.00
CIFAR100               , 00.00, 00.00, 00.00, 00.00
Average                , 00.00, 00.00, 00.00, 00.00
Average (in-domain)    , 00.00, 00.00, 00.00, 00.00
Average (out-of-domain), 00.00, 00.00, 00.00, 00.00
\end{filecontents*}

\begin{filecontents*}{data/all.csv}
TASK            , avg,   ci,    avg,     ci,    avg,        ci,      avg,       ci,    avg,         ci,    avg,         ci,     avg,        ci,         avg,        ci
ImageNet        ,52.3,	1.0,	50.6,	1.1,\bf 58.6,\bf    1.1,	56.4,	    1.2,	55.5,	    1.1,	56.3,	    1.1,	55.7,	    1.1,	    51.8,	    1.1
Omniglot        ,88.4,	0.7,    90.7,	0.6,	91.7,	    0.6,	88.5,	    0.8,	90.2,	    0.6,	93.1,	    0.5,\bf 94.4,   \bf 0.4,	    93.2,	    0.5
Aircraft        ,80.5,	0.6,	83.8,	0.6,	82.4,	    0.7,	79.5,	    0.8,	79.8,	    0.7,	85.4,	    0.7,	85.8,	    0.6,   \bf 87.2,    \bf 0.5
Birds           ,72.2,	0.9,	74.6,	0.8,	74.9,	    0.8,	76.4,	    0.9,	77.5,	    0.8,	71.4,	    1.0,	76.3,	    0.8,   \bf 79.2,    \bf 0.8
Textures        ,58.3,	0.7,	62.1,	0.7,	67.8,	    0.8,\bf 73.1,   \bf 0.7,\bf 73.5,   \bf 0.7,	71.5,	    0.8,	71.8,	    0.7,	    68.8,	    0.8
Quickdraw       ,72.5,	0.8,	74.8,	0.7,	77.7,	    0.7,	75.7,	    0.7,	75.8,	    0.7,	81.3,	    0.6,\bf 82.5,   \bf 0.6,	    79.5,	    0.7
Fungi           ,47.4,	1.0,    48.7,	1.0,	46.9,	    1.0,	48.2,	    0.9,	48.1,	    0.9,\bf 63.1,   \bf 1.0,\bf 63.5,   \bf 1.0,	    58.1,	    1.1
Flower          ,86.0,	0.5,	89.6,	0.6,	90.7,	    0.5,	90.6,	    0.5,\bf 91.9,   \bf 0.5,	82.8,	    0.7,	88.2,	    0.6,   \bf 91.6,    \bf 0.6
Traffic Signs   ,60.2,	0.9,	67.0,	0.7,	73.5,	    0.7,	65.1,	    0.8,	67.5,	    0.8,	70.4,	    0.8,	69.4,	    0.8,	\bf 74.8,   \bf 0.7
MSCOCO          ,42.6,	1.1,	43.4,	1.0,	46.2,	    1.1,\bf 52.1,   \bf 1.0,\bf 52.1,   \bf 1.0,\bf 52.4,   \bf 1.1,\bf 52.2,   \bf 1.1,	    50.0,	    1.0
MNIST           ,92.7,	0.4,	92.3,	0.4,	93.9,	    0.4,	93.2,	    0.4,	93.9,	    0.4,	94.3,	    0.4,	94.8,	    0.4,   \bf 95.6,    \bf 0.5
CIFAR-10        ,61.5,	0.7,	69.3,	0.8,	74.3,	    0.7,	66.4,	    0.8,	66.1,	    0.8,	66.8,	    0.9,	67.3,	    0.8,   \bf 78.6,    \bf 0.7
CIFAR-100       ,50.1,	1.0,	54.6,	1.1,	60.5,	    1.0,	57.1,	    1.0,	57.3,	    1.0,	56.6,	    1.0,	56.9,	    1.0,   \bf 67.1,    \bf 1.0
Average WG      ,69.7,   ,       71.9,   ,       73.8,       ,       73.6,       ,       74.0,       ,       75.6,       ,   \bf 77.3,       ,       76.2
Average SG      ,61.4,   ,       65.3,   ,       69.7,       ,       66.8,       ,       67.4,       ,       68.1,       ,       68.1,       ,   \bf 73.2
Average all     ,66.5,   ,       69.3,   ,       72.2,       ,       70.9,       ,       71.5,       ,       72.7,       ,       73.8,       ,   \bf 75.0
\end{filecontents*}

\begin{filecontents*}{data/ablations.csv}
TASK                   ,avglt,          cifl,       avgsc,     cisc,        avgim,   ciim,   avgimsc,ciimsc, avgsimim,ciimim,avgsurpf,cisurpf
ImageNet        	   ,51.8,           1.1,        47.7,	    1.1,	\bf 53.9,	1.1,	34.2,	1.0,	46.9,	1.1,    46.9,	1.1
Omniglot        	   ,\bf 93.2,       0.5,	    91.2,	    0.6,	    75.6,	1.1,	57.2,	1.4,	61.6,	1.4,    77.9,	1.1
Aircraft        	   ,\bf 87.2,   \bf 0.5,	    80.6,	    0.8,	    66.0,	0.9,	35.3,	0.8,	48.5,	1.0,    67.3,	0.8
Birds        	       ,\bf 79.2,   \bf 0.8,	    72.0,	    0.9,	    73.2,	0.9,	28.8,	0.8,	47.9,	1.0,    60.0,	0.9
Textures        	   ,68.8,	        0.8,   	    69.8,  	    0.7,	\bf 71.9,	0.7,	55.5,	0.7,	63.8,	0.8,    61.2,	0.7
Quickdraw        	   ,\bf 79.5,	    0.7,   	    78.1,  	    0.7,	    69.3,	0.8,	50.6,	1.0,	57.5,	1.0,    60.9,	0.9
Fungi        	       ,\bf 58.1,	    1.1,   	    51.6,  	    1.1,	    46.0,	1.1,	23.0,	0.9,	31.8,	1.0,    34.3,	1.0
Flower        	       ,\bf 91.6,   \bf 0.6,	\bf 91.0,  	    0.6,	    89.2,	0.6,	65.9,	1.0,	80.1,	0.9,    73.8,	0.8
Traffic Signs          ,\bf 74.8,   \bf 0.7,	\bf 74.6,	    0.7,	    72.1,	0.7,	62.5,	0.8,	69.6,	0.7,    66.3,	0.8
MSCOCO        	       ,\bf 50.0,   	1.0,   	    42.5,	    1.0,	\bf 50.6,	1.0,	29.3,	0.9,	41.4,	1.0,    41.3,	1.0
MNIST        	       ,\bf 95.6,   \bf 0.4,   	\bf 95.5,	    0.5,	    83.4,	0.7,	78.9,	0.7,	80.8,	0.8,    86.9,	0.6
CIFAR-10        	   ,\bf 78.6,   \bf 0.7,	    69.6,	    0.9,	    76.9,	0.7,	47.1,	0.8,	65.4,	0.8,    65.4,	0.8
CIFAR-100        	   ,\bf 67.1,   \bf 1.0,	    58.0,  	    1.1,	\bf 67.5,	0.9,	34.5,	1.0,	52.7,	1.1,    52.2,	1.1
Average WG       	   ,\bf 76.2,           ,       72.8,	    ,	        68.1,	,	    43.8,	,	    54.8,   ,       60.3,
Average SG       	   ,\bf 73.2,           ,       68.0,	    ,	        70.1,	,	    50.5,	,	    62.0 ,  ,       62.4,
Average all       	   ,\bf 75.0,           ,       70.9,	    ,	        68.9,	,	    46.4,	,	    57.5 ,  ,       61.1,
\end{filecontents*}

\begin{filecontents*}{data/all_correct_TS.csv}
TASK            , avg,   ci,    avg,     ci,    avg,        ci,      avg,       ci,    avg,         ci,    avg,         ci,     avg,        ci,         avg,        ci
ImageNet        ,52.3,	1.0,	50.6,	1.1,\bf 58.6,\bf    1.1,	56.4,	    1.2,	55.5,	    1.1,	56.3,	    1.1,	55.7,	    1.1,	    51.8,	    1.1
Omniglot        ,88.4,	0.7,    90.7,	0.6,	91.7,	    0.6,	88.5,	    0.8,	90.2,	    0.6,	93.1,	    0.5,\bf 94.4,   \bf 0.4,	    93.2,	    0.5
Aircraft        ,80.5,	0.6,	83.8,	0.6,	82.4,	    0.7,	79.5,	    0.8,	79.8,	    0.7,	85.4,	    0.7,	85.8,	    0.6,   \bf 87.2,    \bf 0.5
Birds           ,72.2,	0.9,	74.6,	0.8,	74.9,	    0.8,	76.4,	    0.9,	77.5,	    0.8,	71.4,	    1.0,	76.3,	    0.8,   \bf 79.2,    \bf 0.8
Textures        ,58.3,	0.7,	62.1,	0.7,	67.8,	    0.8,\bf 73.1,   \bf 0.7,\bf 73.5,   \bf 0.7,	71.5,	    0.8,	71.8,	    0.7,	    68.8,	    0.8
Quickdraw       ,72.5,	0.8,	74.8,	0.7,	77.7,	    0.7,	75.7,	    0.7,	75.8,	    0.7,	81.3,	    0.6,\bf 82.5,   \bf 0.6,	    79.5,	    0.7
Fungi           ,47.4,	1.0,    48.7,	1.0,	46.9,	    1.0,	48.2,	    0.9,	48.1,	    0.9,\bf 63.1,   \bf 1.0,\bf 63.5,   \bf 1.0,	    58.1,	    1.1
Flower          ,86.0,	0.5,	89.6,	0.6,	90.7,	    0.5,	90.6,	    0.5,\bf 91.9,   \bf 0.5,	82.8,	    0.7,	88.2,	    0.6,   \bf 91.6,    \bf 0.6
Traffic Signs   ,56.5,	1.1,	-,	    -,	\bf 59.2,	\bf 1.0,	52.2,	    0.8,	52.0,	    1.4,	53.4,	    1.0,	51.1,	    1.1,	\bf 58.4,   \bf 1.1
MSCOCO          ,42.6,	1.1,	43.4,	1.0,	46.2,	    1.1,\bf 52.1,   \bf 1.0,\bf 52.1,   \bf 1.0,\bf 52.4,   \bf 1.1,\bf 52.2,   \bf 1.1,	    50.0,	    1.0
MNIST           ,92.7,	0.4,	92.3,	0.4,	93.9,	    0.4,	93.2,	    0.4,	93.9,	    0.4,	94.3,	    0.4,	94.8,	    0.4,   \bf 95.6,    \bf 0.5
CIFAR-10        ,61.5,	0.7,	69.3,	0.8,	74.3,	    0.7,	66.4,	    0.8,	66.1,	    0.8,	66.8,	    0.9,	67.3,	    0.8,   \bf 78.6,    \bf 0.7
CIFAR-100       ,50.1,	1.0,	54.6,	1.1,	60.5,	    1.0,	57.1,	    1.0,	57.3,	    1.0,	56.6,	    1.0,	56.9,	    1.0,   \bf 67.1,    \bf 1.0
Average WG      ,69.7,   ,       71.9,   ,       73.8,       ,       73.6,       ,       74.0,       ,       75.6,       ,   \bf 77.3,       ,       76.2
Average SG      ,60.7,   ,       -,     ,       66.8,        ,       64.2,       ,       64.3,       ,       64.7,       ,       64.5,       ,   \bf 69.9
Average all     ,66.2,   ,       -,     ,       71.1,        ,       70.0,       ,       70.3,       ,       71.4,       ,       72.3,       ,   \bf 73.8
\end{filecontents*}

\begin{filecontents*}{data/ablations_correct_TS.csv}
TASK                   ,avglt,          cifl,       avgsc,     cisc,        avgim,   ciim,   avgimsc,ciimsc, avgsimim,ciimim,avgsurpf,cisurpf
ImageNet        	   ,51.8,           1.1,        47.7,	    1.1,	\bf 53.9,	1.1,	34.2,	1.0,	46.9,	1.1,    46.9,	1.1
Omniglot        	   ,\bf 93.2,       0.5,	    91.2,	    0.6,	    75.6,	1.1,	57.2,	1.4,	61.6,	1.4,    77.9,	1.1
Aircraft        	   ,\bf 87.2,   \bf 0.5,	    80.6,	    0.8,	    66.0,	0.9,	35.3,	0.8,	48.5,	1.0,    67.3,	0.8
Birds        	       ,\bf 79.2,   \bf 0.8,	    72.0,	    0.9,	    73.2,	0.9,	28.8,	0.8,	47.9,	1.0,    60.0,	0.9
Textures        	   ,68.8,	        0.8,   	    69.8,  	    0.7,	\bf 71.9,	0.7,	55.5,	0.7,	63.8,	0.8,    61.2,	0.7
Quickdraw        	   ,\bf 79.5,	    0.7,   	    78.1,  	    0.7,	    69.3,	0.8,	50.6,	1.0,	57.5,	1.0,    60.9,	0.9
Fungi        	       ,\bf 58.1,	    1.1,   	    51.6,  	    1.1,	    46.0,	1.1,	23.0,	0.9,	31.8,	1.0,    34.3,	1.0
Flower        	       ,\bf 91.6,   \bf 0.6,	\bf 91.0,  	    0.6,	    89.2,	0.6,	65.9,	1.0,	80.1,	0.9,    73.8,	0.8
Traffic Signs          ,\bf 58.4,   \bf 1.1,	\bf 56.2,	    1.1,	    54.8,	1.1,	36.4,	1.0,	46.5,	1.1,    43.0,	1.1
MSCOCO        	       ,\bf 50.0,   	1.0,   	    42.5,	    1.0,	\bf 50.6,	1.0,	29.3,	0.9,	41.4,	1.0,    41.3,	1.0
MNIST        	       ,\bf 95.6,   \bf 0.4,   	\bf 95.5,	    0.5,	    83.4,	0.7,	78.9,	0.7,	80.8,	0.8,    86.9,	0.6
CIFAR-10        	   ,\bf 78.6,   \bf 0.7,	    69.6,	    0.9,	    76.9,	0.7,	47.1,	0.8,	65.4,	0.8,    65.4,	0.8
CIFAR-100        	   ,\bf 67.1,   \bf 1.0,	    58.0,  	    1.1,	\bf 67.5,	0.9,	34.5,	1.0,	52.7,	1.1,    52.2,	1.1
Average WG       	   ,\bf 76.2,           ,       72.8,	    ,	        68.1,	,	    43.8,	,	    54.8,   ,       60.3,
Average SG       	   ,\bf 69.9,           ,       64.4,	    ,	        66.7,	,	    45.2,	,	    57.4 ,  ,       57.8,
Average all       	   ,\bf 73.8,           ,       69.5,	    ,	        67.6,	,	    44.4,	,	    55.8 ,  ,       59.3,
\end{filecontents*}

\begin{filecontents*}{data/more_runs_inits.csv}
TASK                   , avg,     ci,   avg,     ci,    avg,     ci,    avg,     ci,    avg,     ci
ImageNet        	   , 53.4,	1.1,	53.9,	1.1,	53.9,	1.1,	53.4,	1.1,	53.8,	1.1
Omniglot        	   , 92.8,	0.5,	92.8,	0.5,	92.8,	0.5,	92.8,	0.5,	92.8,	0.5
Aircraft        	   , 87.1,	0.5,	87.1,	0.5,	87.1,	0.5,	87.1,	0.5,	87.1,	0.5
Birds           	   , 78.6,	0.8,	78.6,	0.8,	78.6,	0.8,	78.5,	0.8,	78.6,	0.8
Textures        	   , 67.7,	0.8,	67.7,	0.8,	67.7,	0.8,	67.8,	0.8,	67.7,	0.8
Quickdraw       	   , 79.6,	0.7,	79.6,	0.7,	79.6,	0.7,	79.6,	0.7,	79.6,	0.7
Fungi           	   , 58.1,	1.1,	58.2,	1.1,	58.2,	1.1,	58.0,	1.1,	58.1,	1.1
Flower          	   , 91.5,	0.6,	91.5,	0.6,	91.5,	0.6,	91.5,	0.6,	91.5,	0.6
Traffic Signs   	   , 72.7,	0.7,	73.1,	0.7,	72.4,	0.7,	72.5,	0.7,	73.3,	0.7
MSCOCO          	   , 48.7,	1.0,	50.2,	1.0,	50.3,	1.0,	49.3,	1.0,	50.1,	1.0
MNIST           	   , 95.9,	0.4,	94.3,	0.5,	94.3,	0.5,	94.3,	0.5,	94.3,	0.5
CIFAR-10        	   , 75.8,	0.7,	75.1,	0.8,	75.7,	0.8,	73.8,	0.8,	76.4,	0.7
CIFAR-100       	   , 66.3,	1.0,	63.4,	1.0,	65.1,	1.0,	62.9,	1.0,	66.4,	1.0
Average WG             , 75.4,	,	    75.5,	,	    75.5,	,	    75.4,	,       75.5,
Average SG       	   , 71.9,   ,		71.2,	,	    71.6,	,	    70.6,   ,       72.1,
Average all       	   , 74.1,   ,		73.9,	,	    74.0,	,	    73.6,   ,       74.2,	
\end{filecontents*}

\begin{filecontents*}{data/more_runs_inits_correct_TS.csv}
TASK                   , avg,     ci,   avg,     ci,    avg,     ci,    avg,     ci,    avg,     ci
ImageNet        	   , 53.4,	1.1,	53.9,	1.1,	53.9,	1.1,	53.4,	1.1,	53.8,	1.1
Omniglot        	   , 92.8,	0.5,	92.8,	0.5,	92.8,	0.5,	92.8,	0.5,	92.8,	0.5
Aircraft        	   , 87.1,	0.5,	87.1,	0.5,	87.1,	0.5,	87.1,	0.5,	87.1,	0.5
Birds           	   , 78.6,	0.8,	78.6,	0.8,	78.6,	0.8,	78.5,	0.8,	78.6,	0.8
Textures        	   , 67.7,	0.8,	67.7,	0.8,	67.7,	0.8,	67.8,	0.8,	67.7,	0.8
Quickdraw       	   , 79.6,	0.7,	79.6,	0.7,	79.6,	0.7,	79.6,	0.7,	79.6,	0.7
Fungi           	   , 58.1,	1.1,	58.2,	1.1,	58.2,	1.1,	58.0,	1.1,	58.1,	1.1
Flower          	   , 91.5,	0.6,	91.5,	0.6,	91.5,	0.6,	91.5,	0.6,	91.5,	0.6
Traffic Signs   	   , 51.8,	1.1,	52.6,	1.1,	51.8,	1.1,	52.0,	1.1,	53.8,	1.1
MSCOCO          	   , 48.7,	1.0,	50.2,	1.0,	50.3,	1.0,	49.3,	1.0,	50.1,	1.0
MNIST           	   , 96.0,	0.4,	94.3,	0.5,	94.3,	0.5,	94.3,	0.5,	94.3,	0.5
CIFAR-10        	   , 75.8,	0.7,	75.1,	0.8,	75.7,	0.8,	73.8,	0.8,	76.4,	0.7
CIFAR-100       	   , 66.3,	1.0,	63.4,	1.0,	65.1,	1.0,	62.9,	1.0,	66.4,	1.0
Average WG             , 76.1,	,	    76.2,	,	    76.2,	,	    76.1,	,       76.2,
Average SG       	   , 67.7,   ,		67.1,	,	    67.4,	,	    66.5,   ,       68.2,
Average all       	   , 72.9,   ,		72.7,	,	    72.8,	,	    72.4,   ,       73.1,	
\end{filecontents*}

\begin{filecontents*}{data/more_runs_ftune.csv}
TASK                   , avg,     ci,   avg,     ci,    avg,     ci,    avg,     ci,    avg,     ci
ImageNet        	   , 53.4,	1.1,	53.8,	1.1,	53.6,	1.1,	53.4,	1.1,	51.8,	1.1
Omniglot        	   , 92.9,	0.5,	92.9,	0.5,	93.0,	0.5,	92.9,	0.5,	93.2,	0.5
Aircraft        	   , 87.2,	0.5,	87.2,	0.5,	87.3,	0.5,	87.2,	0.5,	87.2,	0.5
Birds           	   , 78.8,	0.8,	78.8,	0.8,	79.0,	0.8,	78.8,	0.8,	79.2,	0.8
Textures        	   , 68.0,	0.8,	68.0,	0.8,	68.3,	0.8,	68.1,	0.8,	68.8,	0.8
Quickdraw       	   , 79.6,	0.7,	79.6,	0.7,	79.6,	0.7,	79.6,	0.7,	79.5,	0.7
Fungi           	   , 58.2,	1.1,	58.3,	1.1,	58.5,	1.1,	58.3,	1.1,	58.1,	1.1
Flower          	   , 91.6,	0.6,	91.6,	0.6,	91.6,	0.6,	91.6,	0.6,	91.6,	0.6
Traffic Signs   	   , 73.1,	0.7,	73.3,	0.7,	73.0,	0.7,	72.9,	0.7,	74.8,	0.7
MSCOCO          	   , 49.0,	1.0,	50.4,	1.0,	50.8,	1.0,	49.7,	1.0,	50.0,	1.0
MNIST           	   , 96.0,	0.4,	94.5,	0.5,	94.9,	0.5,	94.7,	0.5,	95.6,	0.5
CIFAR-10        	   , 76.6,	0.7,	75.7,	0.8,	77.4,	0.8,	74.8,	0.8,	78.6,	0.7
CIFAR-100       	   , 66.8,	1.0,	63.8,	1.0,	66.2,	1.0,	63.5,	1.0,	67.1,	0.9
Average WG             , 75.6,	,	    75.6,   ,		75.8,   ,		75.6,   ,		75.6,
Average SG       	   , 72.3,	,	    71.5,	,	    72.5,	,	    71.1,	,	    73.2,
Average all       	   , 74.3,	,	    74.1,	,   	74.5,	,   	73.9,	,   	74.7,
















\end{filecontents*}

\begin{filecontents*}{data/more_runs_ftune_correct_TS.csv}
TASK                   , avg,     ci,   avg,     ci,    avg,     ci,    avg,     ci,    avg,     ci
ImageNet        	   , 53.4,	1.1,	53.8,	1.1,	53.6,	1.1,	53.4,	1.1,	51.8,	1.1
Omniglot        	   , 92.9,	0.5,	92.9,	0.5,	93.0,	0.5,	92.9,	0.5,	93.2,	0.5
Aircraft        	   , 87.2,	0.5,	87.2,	0.5,	87.3,	0.5,	87.2,	0.5,	87.2,	0.5
Birds           	   , 78.8,	0.8,	78.8,	0.8,	79.0,	0.8,	78.8,	0.8,	79.2,	0.8
Textures        	   , 68.0,	0.8,	68.0,	0.8,	68.3,	0.8,	68.1,	0.8,	68.8,	0.8
Quickdraw       	   , 79.6,	0.7,	79.6,	0.7,	79.6,	0.7,	79.6,	0.7,	79.5,	0.7
Fungi           	   , 58.2,	1.1,	58.3,	1.1,	58.5,	1.1,	58.3,	1.1,	58.1,	1.1
Flower          	   , 91.6,	0.6,	91.6,	0.6,	91.6,	0.6,	91.6,	0.6,	91.6,	0.6
Traffic Signs   	   , 52.5,	1.1,	53.2,	1.1,	54.5,	1.1,	53.4,	1.1,	58.4,	1.1
MSCOCO          	   , 49.0,	1.0,	50.4,	1.0,	50.8,	1.0,	49.7,	1.0,	50.0,	1.0
MNIST           	   , 96.0,	0.4,	94.5,	0.5,	94.9,	0.5,	94.7,	0.5,	95.6,	0.5
CIFAR-10        	   , 76.6,	0.7,	75.7,	0.8,	77.4,	0.8,	74.8,	0.8,	78.6,	0.7
CIFAR-100       	   , 66.8,	1.0,	63.8,	1.0,	66.2,	1.0,	63.5,	1.0,	67.1,	0.9
Average WG             , 76.2,	,	    76.3,   ,		76.4,   ,		76.2,   ,		76.2,
Average SG       	   , 68.2,	,	    67.5,	,	    68.8,	,	    67.2,	,	    69.9,
Average all       	   , 73.1,	,	    72.9,	,   	73.4,	,   	72.8,	,   	73.8,
















\end{filecontents*}

\begin{filecontents*}{data/traffic_signs_shuffled.csv}
TASK                , avg,      ci,     avg,     ci,    avg,     ci,    avg,     ci,    avg,     ci
Traffic Signs       , 52.5,	    1.1,	53.2,	1.1,	54.5,	1.1,	53.4,	1.1,    58.4,	1.1
\end{filecontents*}

\begin{filecontents*}{data/traffic_signs_unshuffled.csv}
TASK                , avg,      ci,     avg,     ci,    avg,     ci,    avg,     ci,    avg,     ci
Unshuffled Traffic Signs       ,73.1,	    0.7,	73.3,	0.7,	73.0,	0.7,	72.9,	0.7,    74.8,	0.7
\end{filecontents*}

\begin{filecontents*}{data/compare_with_sur.csv}
TASK                   ,avgfl,          cifl,       avgsurpf,     cisurpf,        avgsurpfall,   cisurpfall
ImageNet        	   ,51.8,           1.1,    \bf 56.4,	    1.2,    54.4,	1.1
Omniglot        	   ,\bf 93.2,       0.5,	    88.5,	    0.8,    92.0,	0.6
Aircraft        	   ,\bf 87.2,   \bf 0.5,	    79.5,	    0.8,    87.0,	0.6
Birds        	       ,\bf 79.2,   \bf 0.8,	    76.4,	    0.9,    79.4,	0.8
Textures        	   ,68.8,	        0.8,   	\bf 73.1,   \bf 0.7,    72.3,	0.7
Quickdraw        	   ,\bf 79.5,	    0.7,   	    75.7,	    0.7,    79.1,	0.7
Fungi        	       ,\bf 58.1,	    1.1,   	    48.2,	    0.9,    54.4,	1.1
Flower        	       ,\bf 91.6,   \bf 0.6,        90.6,	    0.5,    91.9,	0.6
Traffic Signs          ,\bf 74.8,   \bf 0.7,        65.1,	    0.8,    69.8,	0.76
MSCOCO        	       ,\bf 50.0,   	1.0,    \bf 52.1,   \bf 1.0,    47.6,	1.0
MNIST        	       ,\bf 95.6,   \bf 0.4,        93.2,	    0.4,    95.8,	0.4
CIFAR-10        	   ,\bf 78.6,   \bf 0.7,        66.4,	    0.8,    66.2,	0.8
CIFAR-100        	   ,\bf 67.1,   \bf 1.0,        57.1,	    1.0,    56.4,	1.0
Average WG       	   ,\bf 76.2,           ,       73.6,       ,       76.32,  ,
Average SG       	   ,\bf 73.2,           ,       66.8,       ,       67.20,  ,
Average all       	   ,\bf 75.0,           ,       70.9,       ,       72.81,  ,
\end{filecontents*}

\begin{filecontents*}{data/compare_with_sur_correct_TS.csv}
TASK                   ,avgfl,          cifl,       avgsurpf,     cisurpf,        avgsurpfall,   cisurpfall
ImageNet        	   ,51.8,           1.1,    \bf 56.4,	    1.2,    54.4,	    1.1
Omniglot        	   ,\bf 93.2,       0.5,	    88.5,	    0.8,    92.0,	    0.6
Aircraft        	   ,\bf 87.2,   \bf 0.5,	    79.5,	    0.8,    \bf 87.0,	\bf 0.6
Birds        	       ,\bf 79.2,   \bf 0.8,	    76.4,	    0.9,    \bf 79.4,	\bf 0.8
Textures        	   ,68.8,	        0.8,   	\bf 73.1,   \bf 0.7,    \bf 72.3,	\bf 0.7
Quickdraw        	   ,\bf 79.5,	    0.7,   	    75.7,	    0.7,    \bf 79.1,	\bf 0.7
Fungi        	       ,\bf 58.1,	    1.1,   	    48.2,	    0.9,    54.4,	    1.1
Flower        	       ,\bf 91.6,   \bf 0.6,        90.6,	    0.5,    \bf 91.9,	\bf 0.6
Traffic Signs          ,\bf 58.4,   \bf 1.1,        52.2,	    0.8,    49.4,	    1.1
MSCOCO        	       ,\bf 50.0,   	1.0,    \bf 52.1,   \bf 1.0,    47.6,	    1.0
MNIST        	       ,\bf 95.6,   \bf 0.4,        93.2,	    0.4,    \bf 95.8,	\bf 0.4
CIFAR-10        	   ,\bf 78.6,   \bf 0.7,        66.4,	    0.8,    66.2,	0.8
CIFAR-100        	   ,\bf 67.1,   \bf 1.0,        57.1,	    1.0,    56.4,	1.0
Average WG       	   ,\bf 76.2,           ,       73.6,       ,       76.3,  ,
Average SG       	   ,\bf 69.9,           ,       64.2,       ,       63.1,  ,
Average all       	   ,\bf 73.8,           ,       70.0,       ,       71.2,  ,
\end{filecontents*}

\begin{filecontents*}{data/hard_blender.csv}
TASK                   ,avgbl,  cibl, avgblft, ciblft,  avghard, cihard, avghardft, cihardft
ImageNet        	   ,53.8,	1.1,	51.8,	1.1,	53.9,	1.1,	51.8,    1.1
Omniglot        	   ,92.8,	0.5,	93.2,	0.5,	92.8,	0.5,	93.2,    0.5
Aircraft        	   ,87.1,	0.5,	87.2,	0.5,	87.1,	0.5,	87.3,    0.5
Birds        	       ,78.6,	0.8,	79.2,	0.8,	78.6,	0.8,	79.2,    0.8
Textures        	   ,67.7,	0.8,	68.8,	0.8,	67.7,	0.8,	68.8,    0.8
Quickdraw        	   ,79.6,	0.7,	79.5,	0.7,	79.6,	0.7,	79.5,	 0.7
Fungi        	       ,58.1,	1.1,	58.1,	1.1,	58.2,	1.1,	58.1,    1.1
Flower        	       ,91.5,	0.6,	91.6,	0.6,	91.5,	0.6,	91.6,    0.6
Traffic Signs          ,73.3,	0.7,	74.8,	1.0,	73.0,	0.7,	74.1,    0.7
MSCOCO        	       ,50.1,	1.0,	50.0,	0.7,	50.4,	1.0,	50.1,    1.0
MNIST        	       ,94.3,	0.5,	95.6,	1.0,	94.3,	0.5,	95.6,    0.5
CIFAR-10        	   ,76.4,	0.7,	78.6,	0.5,	76.5,	0.7,	78.4,    0.7
CIFAR-100        	   ,66.4,	1.0,	67.1,	0.7,	67.0,	0.9,	66.7,    1.0
Average WG       	   ,76.2,	,	76.2,	,	76.2,	,	76.2,
Average SG       	   ,72.1,	,	73.2,	,	72.2,	,	73.0,
Average all       	   ,74.6,	,	75.0,	,	74.7,	,	75.0,
\end{filecontents*}

\begin{filecontents*}{data/hard_blender_correct_TS.csv}
TASK                   ,avgbl,  cibl, avgblft, ciblft,  avghard, cihard, avghardft, cihardft
ImageNet        	   ,53.8,	1.1,	51.8,	1.1,	53.9,	1.1,	51.8,    1.1
Omniglot        	   ,92.8,	0.5,	93.2,	0.5,	92.8,	0.5,	93.2,    0.5
Aircraft        	   ,87.1,	0.5,	87.2,	0.5,	87.1,	0.5,	87.3,    0.5
Birds        	       ,78.6,	0.8,	79.2,	0.8,	78.6,	0.8,	79.2,    0.8
Textures        	   ,67.7,	0.8,	68.8,	0.8,	67.7,	0.8,	68.8,    0.8
Quickdraw        	   ,79.6,	0.7,	79.5,	0.7,	79.6,	0.7,	79.5,	 0.7
Fungi        	       ,58.1,	1.1,	58.1,	1.1,	58.2,	1.1,	58.1,    1.1
Flower        	       ,91.5,	0.6,	91.6,	0.6,	91.5,	0.6,	91.6,    0.6
Traffic Signs          ,53.8,	1.1,	58.4,	1.1,	52.4,	1.1,	57.2,    1.1
MSCOCO        	       ,50.1,	1.0,	50.0,	0.7,	50.4,	1.0,	50.1,    1.0
MNIST        	       ,94.3,	0.5,	95.6,	1.0,	94.3,	0.5,	95.6,    0.5
CIFAR-10        	   ,76.4,	0.7,	78.6,	0.5,	76.5,	0.7,	78.4,    0.7
CIFAR-100        	   ,66.4,	1.0,	67.1,	0.7,	67.0,	0.9,	66.7,    1.0
Average WG       	   ,76.2,	,	76.2,	,	76.2,	,	76.2,
Average SG       	   ,68.2,	,	69.9,	,	68.1,	,	69.6,
Average all       	   ,73.1,	,	73.8,	,	73.1,	,	73.7,
\end{filecontents*}

\newcommand{\rz}[1]{{\color{magenta}RZ: #1}}
\newcommand{\et}[1]{{\color{blue}ET: #1}}

\twocolumn[
  \icmltitle{\titleprefix Learning a Universal Template for Few-shot Dataset Generalization}
  \icmlsetsymbol{equal}{*}
  
  \begin{icmlauthorlist}
  \icmlauthor{Eleni Triantafillou}{vector,google,atGoogle}
  \icmlauthor{Hugo Larochelle}{google}
  \icmlauthor{Richard Zemel}{vector}
\icmlauthor{Vincent Dumoulin}{google}
  \end{icmlauthorlist}
  
  \icmlaffiliation{google}{Google Research, Brain Team}
  \icmlaffiliation{vector}{University of Toronto, Vector Institute}
  \icmlaffiliation{atGoogle}{Work done at Google.}
  
  \icmlcorrespondingauthor{Eleni Triantafillou}{eleni@cs.toronto.edu}
  
  \icmlkeywords{Machine Learning, ICML, Few-Shot Classification}
  
  \vskip 0.3in
]

\printAffiliationsAndNotice{}

\begin{abstract}
Few-shot dataset generalization is a challenging variant of the well-studied few-shot classification problem where a diverse training set of several datasets is given, for the purpose of training an adaptable model that can then learn classes from \emph{new datasets} using only a few examples. To this end, we propose to utilize the diverse training set to construct a \emph{universal template}: a partial model that can define a wide array of dataset-specialized models, by plugging in appropriate components.
For each new few-shot classification problem, our approach therefore only requires inferring a small number of parameters to insert into the universal template. We design a separate network that produces an initialization of those parameters for each given task, and we then fine-tune its proposed initialization via a few steps of gradient descent. Our approach is more parameter-efficient, scalable and adaptable compared to previous methods, and achieves the state-of-the-art on the challenging Meta-Dataset benchmark.
\end{abstract}
\section{Introduction}
While deep learning approaches have recently driven remarkable progress on many important applications, their ability to rapidly learn new concepts from small datasets is significantly lacking. This observation has inspired research towards creating more flexible and adaptable methods. A well-studied problem in this direction is few-shot classification: the problem of utilizing a (possibly large) labeled training set to create an \textit{adaptable} model that is then capable of learning new classes from few examples. Concretely, a few-shot learning model is evaluated on test {\em tasks}, each of which poses a classification problem between previously-unseen classes, given only a few examples of each. 

Departing from early benchmarks for this problem, recent work is shifting towards a more challenging instance of few-shot classification \emph{across datasets} \cite{chen2019closer,tseng2020cross,triantafillou2020meta,requeima2019fast,bateni2019improved,bronskill2020tasknorm,saikia2020optimized,dvornik2020selecting,liu2020universal}. Similarly, we consider the challenging problem of \textit{few-shot dataset generalization}, where the training set is comprised of classes originating from multiple diverse datasets, like ImageNet, Omniglot, Aircraft, and so on, and the aim is to utilize this diverse data towards building a model that can solve new classification tasks between classes of previously-unseen datasets at test time, using only a few examples.

Few-shot dataset generalization exhibits the same challenge of data scarcity as its traditional (single-dataset) few-shot classification counterpart, but it also presents two additional difficulties. Firstly, the heterogeneous nature of the training dataset is reminiscent of \emph{multi-task learning}~\citep{caruana1997multitask}, 
where care must be taken to avoid interference when training jointly for different objectives, or different datasets in this case. 
Secondly, the fact that the test-time tasks are composed of classes from previously-unseen datasets breaks the i.i.d. assumption, resembling the problem of \emph{domain generalization} \cite{gulrajani2020search}. In a nutshell then, to succeed in few-shot dataset generalization, a method must 1) ingest diverse information without interference, and 2) define a mechanism to appropriately re-purpose that acquired knowledge in order to accommodate extensively different data at test time, using few examples. 

In this work, we propose to tackle this problem by learning a {\em universal template}: a partially-parameterized model trained on multiple datasets in parallel that can be used to define a wide array of dataset-specialized models by providing values for the remaining parameters. We posit that this design incorporates a useful inductive bias for few-shot dataset generalization: in addition to benefiting from knowledge acquired on diverse data, it also reflects the requirement that the template should be flexible enough to serve as the basis to parameterize a diverse set of task-solving networks. 

\begin{figure*}[t]%
    \centering%
    \includegraphics[width=0.48\linewidth]{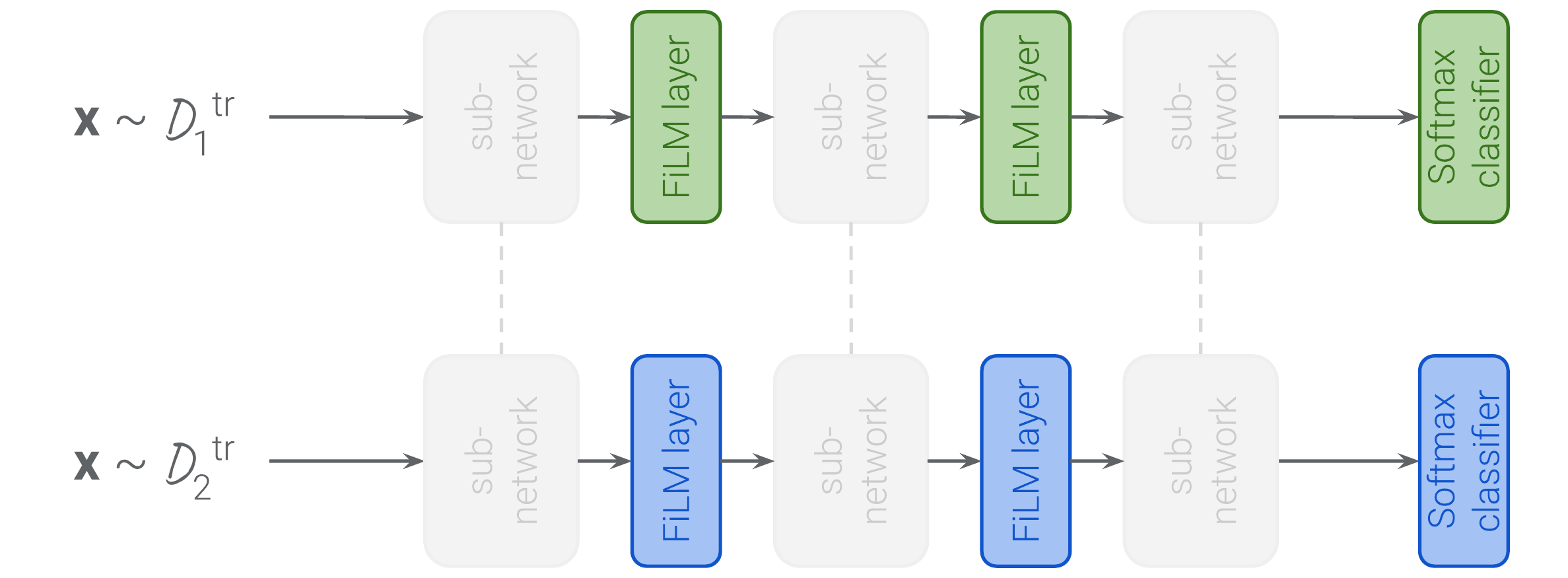}%
    \hfill%
    \includegraphics[width=0.48\linewidth]{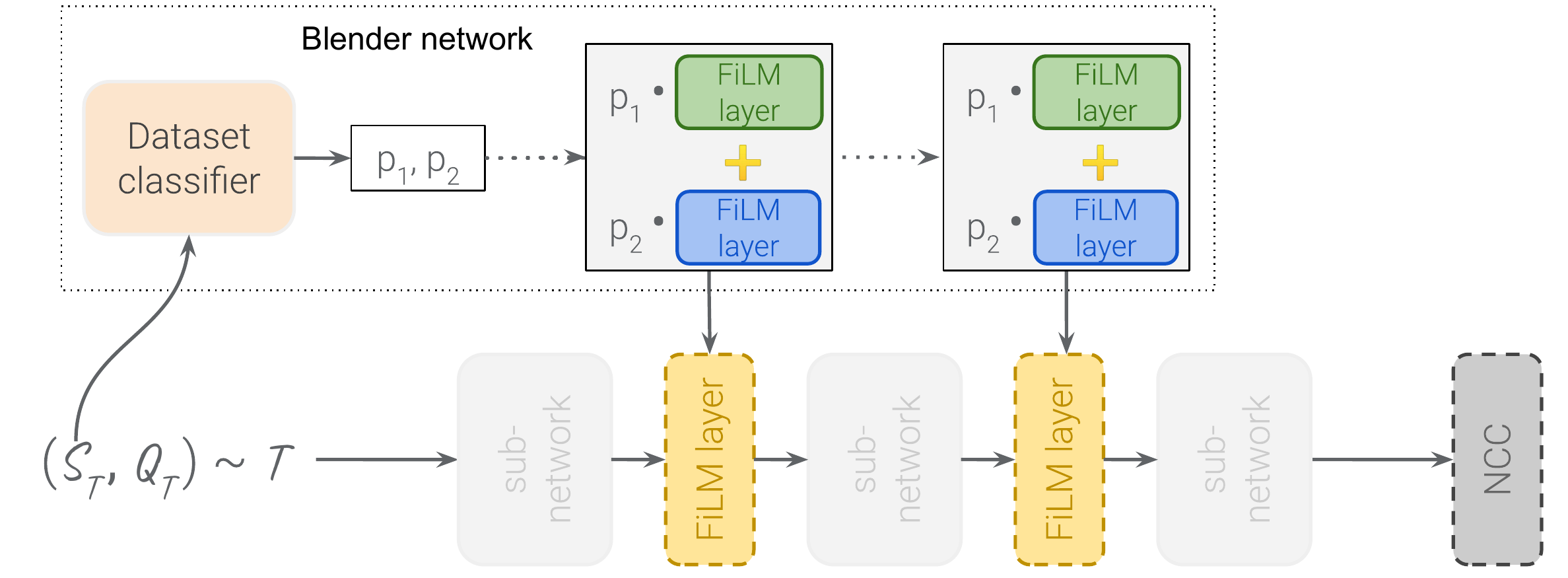}%
    \caption{\label{fig:flute_diagram}({\em Left}) FLUTE trains the shared convolutional weights comprising the universal template, and the $M$ dataset-specific sets of FiLM parameters ($M = 2$ in this illustration) using a multi-task classification loss. ({\em Right}) The FiLM parameter values for a test episode are learned through gradient descent using a nearest-centroid classifier (NCC) fitted on the support set as the output layer. The {\em Blender network} is used to initialize the new FiLM parameters as a convex combination of the trained sets of FiLM parameters. 
    }
\end{figure*}

To this end, we propose to train a feature extractor jointly across diverse datasets using FiLM~\citep{perez2018film} in the form of conditional batch normalization: we share the parameters of the convolutional layers across datasets, but allocate a separate set of batch normalization parameters for each.
This joint training regime is key for creating a universal template: by forcing the batch normalization parameters to be fully responsible for dataset specialization, our training objective ensures that the convolutional parameters are \emph{general}, and thus indeed act as a universal template that is able to support generalization to vastly different datasets. To then tackle test-time tasks from unseen datasets, we propose to use the few available labeled examples of each given task to \emph{learn} a new set of batch normalization parameters for that task, starting from a task-dependent initialization: a combination of the per-dataset trained sets of parameters whose co-efficients are proposed by a separate learned network. We refer to this as Few-shot Learning with a Universal TEmplate (FLUTE), illustrated in~\autoref{fig:flute_diagram}.


We experimentally evaluate FLUTE on few-shot dataset generalization using the recent Meta-Dataset benchmark \cite{triantafillou2020meta} that is comprised of 10 diverse datasets, 8 of which can be used for training, with the remaining 2 reserved for evaluation. To obtain a richer set of evaluation tasks, we incorporate 3 additional evaluation-only datasets, following \citet{requeima2019fast}. FLUTE significantly outperforms the state-of-the-art on few-shot dataset generalization on Meta-Dataset (by $>$ 5\%), despite having significantly fewer parameters (approx. 8 times fewer).

Aside from its strong performance, FLUTE has several advantages over previous approaches. Firstly, it is more scalable: it's more parameter efficient, produces more parsimonious representations, and has favorable test-time computational complexity, as summarized in Table~\ref{table:flute_vs_sur_urt}. It is also very expressive since, for each test task, it blends the per-dataset trained parameters in \emph{all} representation levels (in the batch normalization layers throughout the network), and it is easily adaptable due to its efficient training of a small subset of the feature extractor's parameters in each test task.

\section{Background}
\paragraph{Few-shot Classification} Let $\mathcal{D}^{tr} = \{ (x_i, y_i) \}_{i=1}^{|\mathcal{D}^{tr}|}$ denote a (possibly large) training set, comprised of examples $x_i$ and their corresponding class labels $y_i$, with each $y_i \in  \mathcal{C}^{tr}$. That is, all of the training examples originate from a dedicated training set of classes $\mathcal{C}^{tr}$, while a disjoint validation set of classes $\mathcal{C}^{val}$ is also available for model selection. The aim is to use $\mathcal{D}^{tr}$ to obtain an \emph{adaptable} model that supports learning new classes from few examples.

Then, evaluation is carried out over a series of classification \emph{tasks}, or \emph{episodes}, designed to measure the trained model's ability to learn new classes from few examples. Specifically, a $k$-shot $N$-way test task presents $N$ classes from a set $\mathcal{C}^{test}$ that is disjoint from $\mathcal{C}^{tr}$ and $\mathcal{C}^{val}$. We denote each task as a tuple $\mathcal{T} = (\mathcal{S}_\mathcal{T}, \mathcal{Q}_\mathcal{T})$ of a `support set' and a `query set', representing small `within-task' training and test sets, respectively. The support set contains $k$ labeled examples of each of the $N$ classes: $\mathcal{S}_\mathcal{T} = \{ (x_1, y_1), \dots (x_{kN}, y_{kN}) \}$ with each $y_i \in \{ 1 \dots N \}$, while the query set $\mathcal{Q}_\mathcal{T} = \{ (x_1^*, y_1^*), \dots \}$ contains different (unlabeled) examples of the same $N$ classes.

Assuming a model with parameters $\theta$, the predicted label $\hat y^*$ of a query example $x^*$ is:
\begin{equation*}
    \hat y^* = \argmax_{y^* \in \{ 1 \dots N \}} \log p_{\theta}(y^* | x^*, \mathcal{S}_\mathcal{T})
\end{equation*}
The evaluation metric is then the average query set accuracy over multiple few-shot classification tasks. 

\paragraph{Few-shot Dataset Generalization} 
In this challenging instance of few-shot classification, the training and validation sets are defined as the union over $M$ different datasets:
\begin{align*}
    \mathcal{D}^{tr} &= \mathcal{D}^{tr}_1 \cup \mathcal{D}^{tr}_2 \dots \cup \mathcal{D}^{tr}_M \\
    \mathcal{D}^{val} &= \mathcal{D}^{val}_1 \cup \mathcal{D}^{val}_2 \dots \cup \mathcal{D}^{val}_M 
\end{align*}
where $\mathcal{D}^{tr}_m$ and $\mathcal{D}^{val}_m$ contain data from the dedicated training and validation sets of classes of dataset $m$, respectively. More compactly, we can view the training set (and analogously the validation set too) as $\mathcal{D}^{tr} = \{ (x_i, y_i, d_i) \}_{i=1}^{|\mathcal{D}^{tr}|}$, where the $x_i$'s and $y_i$'s denote examples and their corresponding class labels, as before, and $d_i$ denotes the dataset from which the associated example originates from.

Then, in keeping with the spirit of few-shot classification, the aim at evaluation time is to learn previously-unseen classes from few examples. However, in this more challenging setup, the new classes originate from \emph{new datasets} $\mathcal{D}_{M+1} \dots $, thus inducing a large generalization gap between the training set $\mathcal{D}^{train}$ and the test tasks. The evaluation metric now is the query set accuracy averaged over multiple tasks sampled from multiple held-out datasets.

\paragraph{Feature-wise Linear Modulation (FiLM)} FiLM \cite{dumoulin2018feature-wise} is a general-purpose conditioning mechanism that has been used extensively across diverse applications, like question answering, drawing a painting in the style of another, and so on. It performs an affine feature-wise transformation of its input $x$ based on a condition $d$:
\begin{equation*}
    \mathrm{FiLM}(x) = \gamma(d) \odot x + \beta(d).
\end{equation*}
The dependency of $\gamma$ and $\beta$ on $d$ is handled by maintaining distinct values for each setting of $d$ 
and selecting the appropriate one for the forward pass of each example. 
\section{Introducing FLUTE}\label{sec:flute}

At a high-level, the core idea behind FLUTE is to use multiple diverse training datasets to construct a \emph{universal template} that can represent a wide array of dataset-specific feature extractors by filling in the shared template with appropriate parameters. 
During training, FLUTE learns $M$ such sets of dataset-specific parameters, one per dataset in $\mathcal{D}^{train}$. 

We propose to use FiLM on batch normalization layers (also known as conditional batch normalization~\citep{de2017modulating}) as a concrete instantiation of this framework. The shared template therefore consists of the convolutional kernels of a residual network~\citep{he2016deep}, and the dataset-specific parameters to be filled-in are the batch normalization layers' scaling and shifting coefficients, which we refer to as the \emph{FiLM parameters} from now on. 


Our reason for choosing FiLM for this purpose is two-fold. First, while simple, it is very expressive and flexible, as evidenced by its wide success in vastly different problem settings~\citep{dumoulin2018feature-wise}. This is very important for our goal of generalizing to extensively different data. Second, FiLM layers contain a small fraction of a network's overall parameters, which is crucial for our few-shot setting where limited data is available for learning how to specialize our template to test tasks of different datasets.

At evaluation time, for each test task, our goal is to produce a new set of FiLM parameters that yields a good feature extractor for that task when plugged into the universal template. We propose to \emph{learn} such parameters through gradient descent on the given task's support set. In doing so, we also need to specify how to initialize the new FiLM parameters. In practice, we find that initialization really matters, and that simply performing more gradient steps is insufficient to recover from a bad initialization. We propose a simple heuristic that we find works well in practice: initialize a new task's FiLM parameters from a convex combination of the training datasets' FiLM parameters where the combination weights are a function of how ``compatible'' each training dataset is with the new task. We define this ``compatibility'' to be the output of a dataset classifier that discriminates between the $M$ training datasets. We refer to this approach in what follows using the term \emph{Blender network}.

In what follows, we first describe how we train our universal template, FiLM parameters, and Blender network, and then present our algorithm for tackling few-shot classification tasks from previously-unseen datasets.

\subsection{Training the universal and per-dataset parameters} 
We define our universal template as the convolutional layer parameters of our ResNet-18 feature extractor, and the dataset-specific components take the form of FiLM parameters via different settings for the batch normalization layers.


We train our feature extractor \emph{jointly} over all $M$ training datasets, and we maintain $M$ sets of FiLM parameters for distinguishing the forward passes of examples of different datasets. Formally, let $f$ denote our feature extractor function, and let $\Phi$ and $\Psi$ be the shared universal template parameters (the convolutional layer parameters of $f$) and the dataset-specific  parameters, respectively, with $\Psi_m$ (row vector of $\Psi$) denoting the FiLM parameters for dataset $m$. Finally, let $\Omega$ denote the parameters of per-dataset classification readout heads $r_1, \dots r_M$, where the readout layer $r_m$ for dataset $m$ has $|\mathcal{C}^{tr}_m|$ units and is parameterized by $\Omega_m$.

Then, each update to the model is computed as follows: we sample a batch $\mathcal{B}$ of data from the training set $\mathcal{B} = \{ (x_i, y_i, d_i) \}_{i=1}^{|\mathcal{B}|}$, with each element $i$ of the batch being a tuple that contains an image $x_i$, its one-hot-encoded class label $y_i$, and its one-hot encoded dataset label $d_i$. We compute the network's prediction of the class label $\hat y_i$ as follows:
\begin{equation*}
    \hat y_i = r_{d_i}(f(x_i; \Phi, \Psi_{d_i}); \Omega_{d_i})
\end{equation*}
We then train $\Phi$, $\Psi$, and $\Omega$ to minimize the cross-entropy loss:
\begin{equation*}
    - \sum_{i=1}^{|\mathcal{B}|} y_i \log \hat y_i 
\end{equation*}

Notice that the loss of an example $x_i$ influences only the FiLM parameters corresponding to its originating dataset $d_i$, but \emph{all} examples influence the shared parameters $\Phi$. We posit that this design is essential for rendering $\Phi$ a \emph{universal} template that is general enough to enable dataset specialization to a wide range of diverse datasets by merely changing the values of the parameter-light batch normalization layers.

After this phase, we no longer require the per-dataset  readout heads $r_1, \dots r_M$, so we discard them at this point.

\subsection{Training the Blender network}
The purpose of this network is to create (an initialization for) the FiLM parameters for an unseen dataset $d^*$ via appropriately blending the $M$ sets of FiLM parameters $\{ \Psi_m \}_{m=1}^M$. The aim, in particular, is to yield a good feature extractor for $d^*$ by in-filling the universal template with these new FiLM parameters. We define the blending operation as a convex combination whose co-efficient for a particular dataset $m$ is given by an estimate of compatibility between $m$ and $d^*$.


Towards estimating that compatibility, we propose to train a \emph{dataset classifier}: a network that reads a batch of data and predicts which of the training datasets it was sampled from. The idea is then to interpret this dataset classifier's probablities as a proxy for the compatibility between the data it ingested and each training dataset.

Formally, let $g$ denote a network that takes as input a \emph{set} of examples $\mathcal{B}$ and creates a vector representation of that set. We use a permutation-invariant `set encoder' \cite{zaheer2017deep} for $g$ that we describe in the Appendix. Also let $l$ denote a $M$-dimensional linear readout layer. 

Each update to the parameters of $g$ and $l$ is performed by first sampling a training dataset $m$ (an integer between $1$ and $M$), and then sampling $\mathcal{B}$ from $\mathcal{D}^{tr}_m$. We compute the dataset classifier's logits as $\hat d = l(g(\mathcal{B}))$, and we train the parameters of $g$ and $l$ to minimize the cross-entropy loss for the dataset classification problem. 

Once the dataset classifier is trained, we define the blender function as
\begin{equation*}
    \blend(\mathcal{B}) = \softmax(l(g(\mathcal{B})))^T\Psi
\end{equation*}
In words, this is a convex combination of the trained sets of FiLM parameters (the rows of $\Psi$), where the combination co-efficients are the probabilities of the given batch of data $\mathcal{B}$ of unknown origin belonging to each of the training datasets.

\subsection{Tackling few-shot test tasks}
Let $\mathcal{T} = \{ \mathcal{S}_{\mathcal{T}}, \mathcal{Q}_{\mathcal{T}} \}$ denote a few-shot test task coming from an unseen dataset $d^*$. Our approach to solving $\mathcal{T}$ can be broken down into two steps. First, we define a feature extractor for $\mathcal{T}$ via an appropriate in-filling of the universal template, and second, we predict a class label for each query example via a simple algorithm that operates on top of the constructed feature extractor.

\paragraph{Step 1: Defining the feature extractor} In this step, we seek a set of FiLM parameters $\Psi_{d^*}$ for the given task of unseen dataset $d^*$ that yield a good feature extractor $f(\cdot ; \Phi, \Psi_{d^*})$ when plugged into the universal template.

We initialize $\Psi_{d^*}$ to the convex combination proposed by the blender network based on the task's support set:
\begin{equation*}
    \Psi_{d^*}^{init} = \blend(\mathcal{S}_{\mathcal{T}})
\end{equation*}

Then, we use gradient descent to train $\Psi_{d^*}$ from the initialization $\Psi_{d^*}^{init}$. We next describe the training objective that we use for that purpose, using a Nearest-Centroid Classifier.

\paragraph{Nearest-Centroid Classifier (NCC)}
The NCC \cite{mensink2013distance,snell2017prototypical} uses the support set to define a centroid $c_j$ for each class $j$ by averaging the features of the support examples belonging to class $j$:
\begin{equation*}
    c^{\Psi_{d^*}}_j = \frac{1}{|\mathcal{S}_{\mathcal{T}}^j|} \sum_{x \in \mathcal{S}_{\mathcal{T}}^j} f(x; \Phi, \Psi_{d^*})
\end{equation*}
where $\mathcal{S}_{\mathcal{T}}^j = \{ (x, y) \in \mathcal{S}_{\mathcal{T}}: y = j \}$ is the subset of the support set for class $j$. The $\Psi_{d^*}$ superscript in $c^{\Psi_{d^*}}_j$ serves to emphasize that the centroid was created from features  obtained by plugging in $\Psi_{d^*}$ into the shared template.

Under this classifier, the probability of an example $x$ belonging to class $j$ is proportional to the exponential of the cosine similarity between its features and that class centroid:
\begin{equation*}
    p^{\Psi_{d^*}}(j | x, \{ c^{\Psi_{d^*}}_j \}_{j=1}^N) \propto \exp  \frac{f(x; \Phi, \Psi_{d^*}) \cdot c^{\Psi_{d^*}}_j}{||f(x; \Phi, \Psi_{d^*})|| \cdot ||c^{\Psi_{d^*}}_j||}
\end{equation*}


\paragraph{Learning the per-task parameters} We learn $\Psi_{d^*}$ via gradient descent to minimize the NCC loss on the support set:
\begin{equation*}
    \Psi_{d^*} = \Psi_{d^*}^{init} - \epsilon \frac{\partial }{\partial \Psi_{d^*}}  \left(\sum_{(x, y) \in \mathcal{S}_\mathcal{T}} \mathcal{L}^{\Psi_{d^*}}(x, y, \{ c^{\Psi_{d^*}}_j \}_{j=1}^N)\right)
\end{equation*}
where $\epsilon$ is the learning rate and $\mathcal{L}^{\Psi_{d^*}}(x, y, \{ c^{\Psi_{d^*}}_j \}_{j=1}^N)$ denotes the cross-entropy loss for example $(x, y)$ under the NCC.
In practice, we might take more than a one step of gradient descent; we treat that as a hyperparameter. Note that all parameters aside from $\Psi_{d^*}$ are frozen at this time.

We also emphasize that we learn $\Psi_{d^*}$ on a per-task basis at test time, using the task's support set. That is, $\Psi_{d^*}$ is not re-used across the test tasks that are sampled from $d^*$.

\paragraph{Step 2: Predict a class label for each query example} Finally, having defined a feature extractor $f(\cdot ; \Phi, \Psi_{d^*})$ for the given test task, we can utilize the NCC to classify each query example. Concretely, we classify a query $x^*$ as:
\begin{equation*}
    \hat y^* = \argmax_{j} p^{\Psi_{d^*}}(j | x^*, \{ c^{\Psi_{d^*}}_j \}_{j=1}^N) 
\end{equation*}
\section{Related Work}
Few-shot dataset generalization is related to various well-studied areas. The aim in multi-task learning \cite{crawshaw2020multi}, for example, is to learn multiple tasks simultaneously with a shared model, similar to FLUTE's approach for training on multiple diverse datasets jointly. However, contray to that problem, our goal is to generalize to \emph{new datasets}. Domain generalization \cite{gulrajani2020search} also relates to our problem setting due to the shared goal of learning a model that generalizes to vastly different data distributions, but it typically assumes that the label space is shared across datasets, contrary to our setting.

Another related area is transfer learning \cite{zhuang2020comprehensive}, aiming to improve a learner's performance on target domains via leveraging knowledge obtained from related source domains. A representative approach that is related to FLUTE is that of \citet{puigcerver2021scalable}: they first train a set of diverse `experts' on different data and then, for each downstream task, they select the most appropriate expert and fine-tune it on that task. FLUTE differs in that, thanks to its universal template that can be re-used across tasks of different datasets, it only needs to fine-tune a small number of per-task parameters at test time (the FiLM parameters); a design that allows to effectively leverage previous knowledge and avoid overfitting in our data-scarce setting.

Moreover, FLUTE is related to several few-shot learning methods. Many popular ones are metric-based \cite{vinyals2016matching,snell2017prototypical,sung2018learning}, or learning-to-learn approaches \cite{ravi2016optimization,finn2017model}. A particular variant of the influential Model Agnostic Meta Learning (MAML) model that is especially related to FLUTE is CAVIA \cite{zintgraf2019fast} that also divides the feature extractor's parameters into task-general and task-specific ones. At test time, only the task-specific ones are adapted for each new episode. Moreover, \citep{flennerhag2019meta} propose to interleave meta-learned `warp-layers' between the layers of a task learner, in the context of gradient-based meta-learning. Reminiscent of FLTUE's template, their warp-layers are shared across tasks, with the aim in that case being to efficiently parameterize a preconditioning matrix that facilitates within-task learning. Albeit related, these previous approaches do not target our across-dataset problem setting of interest, and differ algorithmically as they utilize a meta-learning approach, whereas FLUTE is trained across datasets, with the simple training objective of joint dataset classification. Keeping with the spirit of separating shared from specific weights, \citep{wang2019meta} design a model for few-shot object detection based on the insight of disentangling category-specific and category-agnostic parameters, in a similar spirit to FLUTE's dataset-specific and dataset-agnostic parameters.  However, their approach differs significantly on an architectural level, as well as in terms of their training objective and overall goal and application area of the work.

FiLM has also been used in several few-shot learning approaches, typically for the purpose of conditioning on each episode's support set to construct a set of features that are tailored to the task at hand \cite{oreshkin2018tadam,requeima2019fast,bateni2019improved}. FLUTE differs from these works in that, consistent with its goal of constructing a dataset-general template, it conditions on the dataset label, instead of conditioning on a continuous representation of support-set features as those previous works. This idea relates to \cite{rebuffi2017learning}'s work that introduce `residual adapters', a generalization of FiLM, to condition on different datasets. However, that work does not consider the problem that is our focus: generalization to new datasets at test time.

The idea of using a dataset classifier in the Blender network to guide the initializaiton of new sets of FiLM parameters is reminiscent of \cite{torralba2011unbiased}'s earlier work, where they noticed that it is possible to accurately classify datasets, suggesting that each has a unique “signature”.

Finally, perhaps the most closely-related family of work is that of building a \emph{universal representation}; a term coined in \citet{bilen2017universal} to describe a set of rich features that enable strong performance on several different datasets in a multi-task learning setup. Building on this idea, the approach of Selecting from a Universal Representation (SUR \citep{dvornik2020selecting}) achieved strong performance by training a separate feature extractor on each training dataset (for a total of $M$ extractors). To tackle each test task, they obtain the $M$ sets of extracted features for the task's data, and concatenate them to form a universal representation of size $MD$, where $D$ is the representation size of each feature extractor. They then utilize a selection mechanism that weights these $MD$ features based on relevance to the task at hand. \citet{liu2020universal} later improved upon SUR by replacing its feature weighting mechanism by a meta-learned attention layer referred to as the Universal Representation Transformer layer (URT). To alleviate the cost of training $M$ separate feature extractors, SUR and URT offer `paramteric family' variants, SUR-pf and URT-pf, respectively, where their $M$ feature extractors share all but their batch normalization weights; a design that is architecturally similar to FLUTE's. Unfortunately, though, the compactness of SUR-pf/URT-pf results in 
a significant drop in accuracy.

Compared to those works, FLUTE embodies a different inductive bias: we don't seek a set of universal features from which we select dataset-relevant ones. Instead, we seek a universal template that defines several dataset-specialized feature extractors, each of which only produces dataset-relevant features. An additional difference is that SUR-pf/URT-pf train the shared parameters only on ImageNet, whereas we employ a joint training phase that permits all datasets to influence those parameters; a choice that 
is crucial for turning the shared parameters into a general-purpose universal template. Finally, FLUTE scales better with the number of training datasets, as summarized in Table~\ref{table:flute_vs_sur_urt}.

\begin{table}[ht]
	\ra{1.2}
	\caption{\label{table:flute_vs_sur_urt} Comparing FLUTE to universal representation methods in terms of scalability with the number $M$ of training datasets. $P$ is the number of feature extractor parameters, $D$ the size of representation that the feature extractor outputs (e.g. 512 for a ResNet-18), $H$ the number of URT's attention heads and $T$ the number of FLUTE's within-task training steps to learn $\Psi_h$ during evaluation. `Computation' measures the number of passes through the feature extractor(s) for each task. SUR(-pf) and URT(-pf) forward-pass through their $M$ extractors (some of which share parameters in the pf variants) whereas we only maintain a single feature extractor. We perform a forward and a backward pass on the support set for each of our $T$ steps, and a final forward pass on the query set ($2T+1$ in total); thus FLUTE's amount of test-time computation 
		does \emph{not} grow with the number of training datasets as for SUR(-pf) and URT(-pf), and $T$ can be small in practice.}
	\begin{center}
		\begin{tabular}{lccc}
			\toprule
			& Num params & Size of repr. & Computation \\
			SUR & $MP$ & $MD$ & $M$\\
			URT & $MP$ & $HD$ & $M$\\
			SUR-pf & $P$ & $MD$ & $M$\\
			URT-pf & $P$ & $HD$ & $M$\\
			FLUTE & $P$ & $D$ & $2T+1$\\
			\bottomrule
		\end{tabular}
	\end{center}
\end{table}


\section{Experimental Setup}

    
    
    
We conduct our experimental evaluation on the recent Meta-Dataset benchmark, which lends itself well to studying few-shot dataset generalization, as its designated training set is indeed comprised of diverse datasets, and its test set includes classes from previously-unseen datasets too.

\paragraph{Meta-Dataset} In more detail, the training set contains classes from ImageNet, Omniglot, Aircraft, Birds, Flowers, Quickdraw, Fungi, and Textures, some of which contain natural images (e.g. ImageNet, Birds, and Flowers), while others differ significantly in appearance. For example, Omniglot and Quickdraw depict hand drawings on a plain black background and Textures's images present perceptual characteristics of varying structure and don't feature a single object in the center of the image. The test set introduces two new datasets: Traffic Signs and MSCOCO, with the former differing significantly thematically from the training set (it depicts images of different traffic signs), and the latter, albeit thematically similar to ImageNet, has lower resolution and exhibits inclusion. Since the aim of this work is to study few-shot dataset generalization, we incorporate the 3 additional test datasets introduced in \citet{requeima2019fast}, to obtain a larger and more varied test set for this problem. These datasets are MNIST, CIFAR-10, and CIFAR-100.

\paragraph{Strong versus weak generalization} The problem of few-shot dataset generalization requires \emph{strong generalization} as there is a large generalization gap between the training and test distributions. Another popular, albeit easier, setting creates the test tasks from held-out classes of the datasets that were encountered at training time. This still falls within the umbrella of few-shot classification, as the test classes were not used for training, but is an easier problem since the generalization gap is inherently smaller. We thus refer to this setting as \emph{weak generalization}. 
Although our focus is few-shot dataset generalization in this work, we also examine FLUTE's performance on the weak generalization tasks for completeness and compatibility with previous work.

We apply FLUTE in exactly the same way for this setting as described in Section~\ref{sec:flute}. In particular, despite the test classes originating from seen datasets in this setting, we are not told which of the $M$ datasets each task is sampled from. 
Therefore, we still require our Blender network to propose a blending of the $M$ sets of FiLM parameters as the initialization of $\Psi_{d^*}$, although we expect that it should almost exclusively select the correct dataset from which the task originates, assuming the dataset classifier is accurate.

\paragraph{Implementation details} Following the prior work, we use a ResNet-18 as the feature extractor. During training of our universal template, we treat each dataset-specific readout head $r_m$ as a cosine classifier (as in \citet{chen2019closer, chen2020new}): a linear layer without a bias, where both the layer inputs and the rows of the weight matrix are l2-normalized. We use stochastic gradient descent with momentum as the optimizer for this phase, with a cosine decay with restarts schedule for the learning rate. On the other hand, the second phase of training for our dataset classifier network converges quite quickly (around 14K steps). The design of the dataset classifier involves a deep set encoder, comprised by 5 convolutional layers, each followed by batch normalization and ReLU, with a standard linear readout on top. We use Adam for this phase. We report all details in the Appendix. Our code is publicly available and has been incorporated into the Meta-Dataset codebase\footnote{https://github.com/google-research/meta-dataset}.
\section{Results}

\begin{table*}
\ra{1.2}
\caption{\label{table:main} Comparison of FLUTE to the previous state-of-the-art approaches on Meta-Dataset. The few-shot dataset generalization performance is shown in the second group of rows (Traffic Signs - CIFAR-100), representing unseen datasets that require Strong Generalization (SG). For completeness, we also present results on the easier problem of Weak Generalization (WG) in the first 8 rows (ImageNet - Flower). We compare against CNAPs \cite{requeima2019fast}, TaskNorm \cite{bronskill2020tasknorm}\footnotemark, SimpleCNAPS \cite{bateni2019improved}, SUR(-pf) \cite{dvornik2020selecting} and URT(-pf) \cite{liu2020universal}. We emphasize that SUR and URT have 8 times more parameters compared to all other competitors (denoted by x8). FLUTE outperforms these previous methods by a significant margin for the problem of few-shot dataset generalization ($>5\%$ on Average SG) while still outperforming the state-of-the-art overall. Each number represents the average query set accuracy over 600 test tasks, and its 95\% confidence interval.}
\begin{center}
\scriptsize
\begin{tabular}{@{}lcccccccc@{}}
  \toprule
  Dataset & CNAPs & TaskNorm & SimpleCNAPs & SUR-pf & URT-pf & SUR (x8) & URT (x8) & FLUTE \\
  \midrule
  \DTLforeach{all_correct_TS}{%
    \task=Column1,%
    \avgI=Column2,%
    \ciI=Column3,%
    \avgII=Column4,%
    \ciII=Column5,%
    \avgIII=Column6,%
    \ciIII=Column7,%
    \avgIX=Column8,%
    \ciIX=Column9,%
    \avgX=Column10,%
    \ciX=Column11,%
    \avgXI=Column12,%
    \ciXI=Column13,%
    \avgXII=Column14,%
    \ciXII=Column15,%
    \avgXIII=Column16,%
    \ciXIII=Column17%
  }{%
    \ifthenelse{\value{DTLrowi}=1}{}{%
      \ifthenelse{\value{DTLrowi}=9 \OR \value{DTLrowi}=14}{\\\midrule}{\\}%
    }%
    \dtlformat{\task}&%
    \dtlformat{\avgI} \ifthenelse{\value{DTLrowi}>13}{}{$\pm$ \dtlformat{\ciI}}&%
    \dtlformat{\avgII} \ifthenelse{\value{DTLrowi}>13 \OR \value{DTLrowi}=9}{}{$\pm$ \dtlformat{\ciII}}&%
    \dtlformat{\avgIII} \ifthenelse{\value{DTLrowi}>13}{}{$\pm$ \dtlformat{\ciIII}}&%
    \dtlformat{\avgIX} \ifthenelse{\value{DTLrowi}>13}{}{$\pm$ \dtlformat{\ciIX}}&%
    \dtlformat{\avgX} \ifthenelse{\value{DTLrowi}>13}{}{$\pm$ \dtlformat{\ciX}}&%
    \dtlformat{\avgXI} \ifthenelse{\value{DTLrowi}>13}{}{$\pm$ \dtlformat{\ciXI}}&%
    \dtlformat{\avgXII} \ifthenelse{\value{DTLrowi}>13}{}{$\pm$ \dtlformat{\ciXII}}&%
    \dtlformat{\avgXIII} \ifthenelse{\value{DTLrowi}>13}{}{$\pm$ \dtlformat{\ciXIII}}%
  }
  \\\bottomrule
\end{tabular}
\end{center}
\end{table*}

\paragraph{Performance on Meta-Dataset} As our first experiment, we evaluate FLUTE on Meta-Dataset, in order to assess its performance on the challenging problem of few-shot dataset generalization, as well as on the weaker generalization setup. We compare against previous approaches that have the same number of parameters as FLUTE, as well as the state-of-the-art models on Meta-Dataset that use roughly 8 times more parameters than FLUTE. We present these results in Table~\ref{table:main}. We observe that FLUTE significantly outperforms all approaches on few-shot dataset generalization. Overall, it sets the new state-of-the-art on Meta-Dataset, despite FLUTE's compactness and parameter efficiency.

Having established FLUTE's effectiveness, we next take a closer look at the effect of its various design choices. \footnotetext{We don't report Traffic Signs results for this method due to an inconsistency in how evaluation on this dataset was performed (described in the Appendix)}

\begin{table}[h]
\ra{1.2}
\caption{\label{table:ablations_wo_ci} The effect of training on different data (`All', as in FLUTE, or `ImageNet-only'), and alternative initialization schemes for $\Psi_{d^*}$: from scratch (`Scratch'), from ImageNet's FiLM parameters (`$\Psi_{IN}$') and from Blender. The column corresponding to the setting ``All, Blender'' is our proposed FLUTE model. Each number represents the average query set accuracy over 600 test tasks, and its 95\% confidence interval. We include confidence intervals for this table in the Appendix.}
\begin{center}
\scriptsize
\begin{tabular}{@{}l|ccc|ccc@{}}
\toprule
\multicolumn{1}{l}{Training data} & \multicolumn{3}{c}{All} & \multicolumn{3}{c}{ImageNet only} \\
\multicolumn{1}{l}{Init scheme} & Scratch & $\Psi_{IN}$ & Blender & Scratch & $\Psi_{IN}$ & Blender \\
\midrule
\DTLforeach{ablations_correct_TS}{%
  \task=Column1,%
  \avgI=Column2,%
  \ciI=Column3,%
  \avgII=Column4,%
  \ciII=Column5,%
  \avgIII=Column6,%
  \ciIII=Column7,%
  \avgIX=Column8,%
  \ciIX=Column9,%
  \avgX=Column10,%
  \ciX=Column11,%
  \avgXI=Column12,%
  \ciXI=Column13%
}{%
  \ifthenelse{\value{DTLrowi}=1}{}{%
    \ifthenelse{\value{DTLrowi}=9 \OR \value{DTLrowi}=14}{\\\midrule}{\\}%
  }%
  \dtlformat{\task}&%
  \dtlformat{\avgII} \ifthenelse{\value{DTLrowi}>13}{}{}&%
  \dtlformat{\avgIII} \ifthenelse{\value{DTLrowi}>13}{}{}&%
  \dtlformat{\avgI} \ifthenelse{\value{DTLrowi}>13}{}{}&%
  \dtlformat{\avgIX} \ifthenelse{\value{DTLrowi}>13}{}{}&%
  \dtlformat{\avgX} \ifthenelse{\value{DTLrowi}>13}{}{}&%
  \dtlformat{\avgXI} \ifthenelse{\value{DTLrowi}>13}{}{}%
}
\\\bottomrule
\end{tabular}
\end{center}
\end{table}

\paragraph{The importance of the Blender network} Recall that the role of this network is to propose an initialization for the FiLM parameters for each test task. We now explore two alternative ways of initializing those parameters: `from scratch' and `from ImageNet' ($\Psi_{IN}$). For each layer $i$, the former sets the additive terms $\beta_i$ to vectors of 0s, and the multiplicative terms $\gamma_i$ to vectors of 1s, while the latter sets the FiLM parameters throughout the network to those learned specifically for the ImageNet dataset, which we denote by $\Psi_{IN}$. We singled out ImageNet for this since it is often treated as a good source for transfer.

The results for this experiment are shown in the first three columns of Table~\ref{table:ablations_wo_ci}. We find that initializing from $\Psi_{IN}$ yields impressive results on several datasets (e.g. ImageNet (unsurprisingly), but also Textures, CIFAR-10, CIFAR-100) but very poor results on datasets that are vastly different from ImageNet (e.g. Omiglot, Quickdraw, and Aircraft). On the other hand, initializing from scratch does not recover that strong performance on the ImageNet-like datasets, but evidently constitutes a much better solution for various other datasets (e.g. Omniglot, Quickraw and Aircraft). Therefore, the initialization of $\Psi_{d^*}$ really matters, with different techniques working best for different evaluation datasets. Our Blender network thus provides a good solution by tailoring the initialization to each task at hand, via the predicted compatibility with each training dataset.

\begin{figure*}[t]%
    \centering%
    \includegraphics[height=2.1cm]{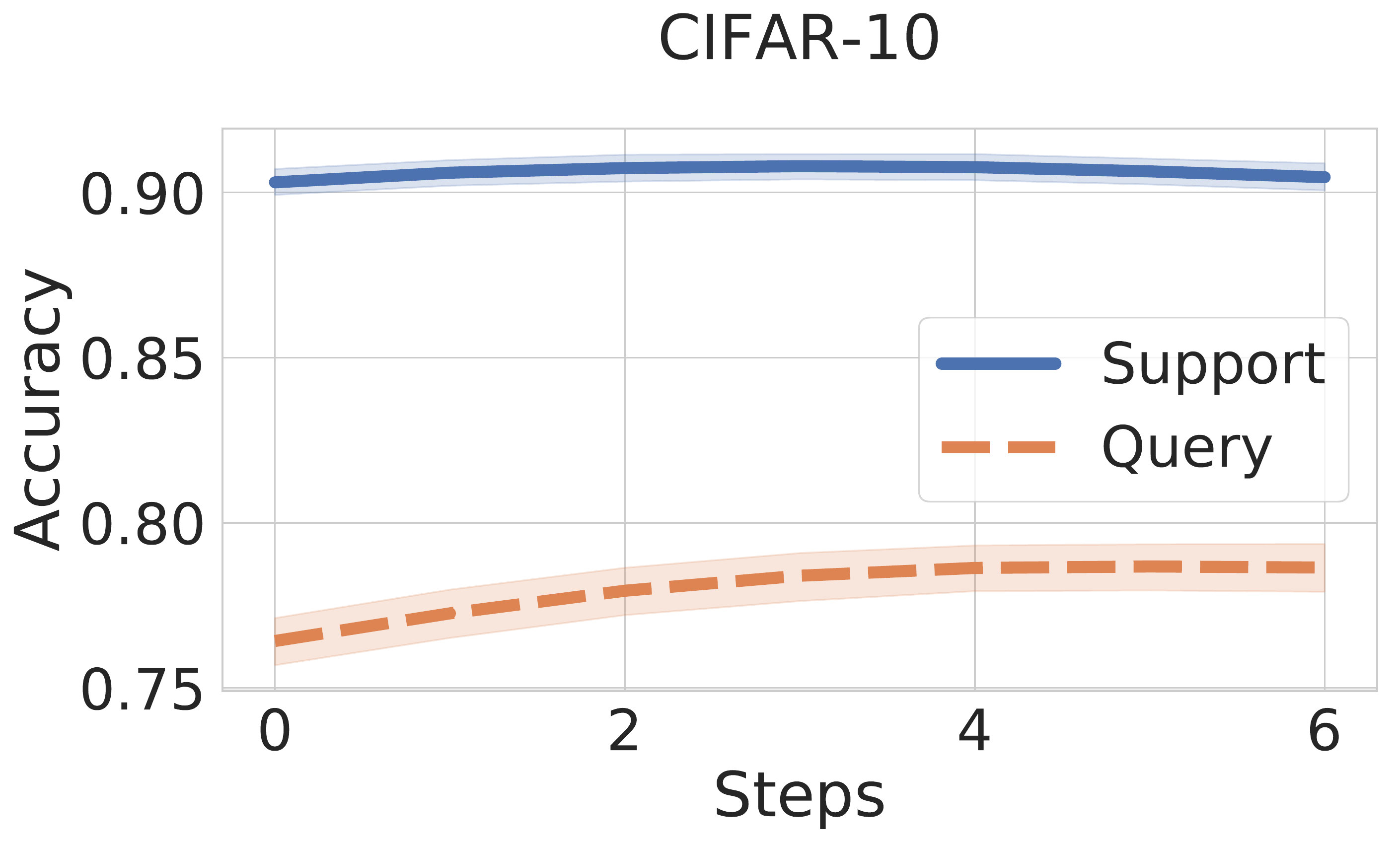}%
    \includegraphics[height=2.1cm]{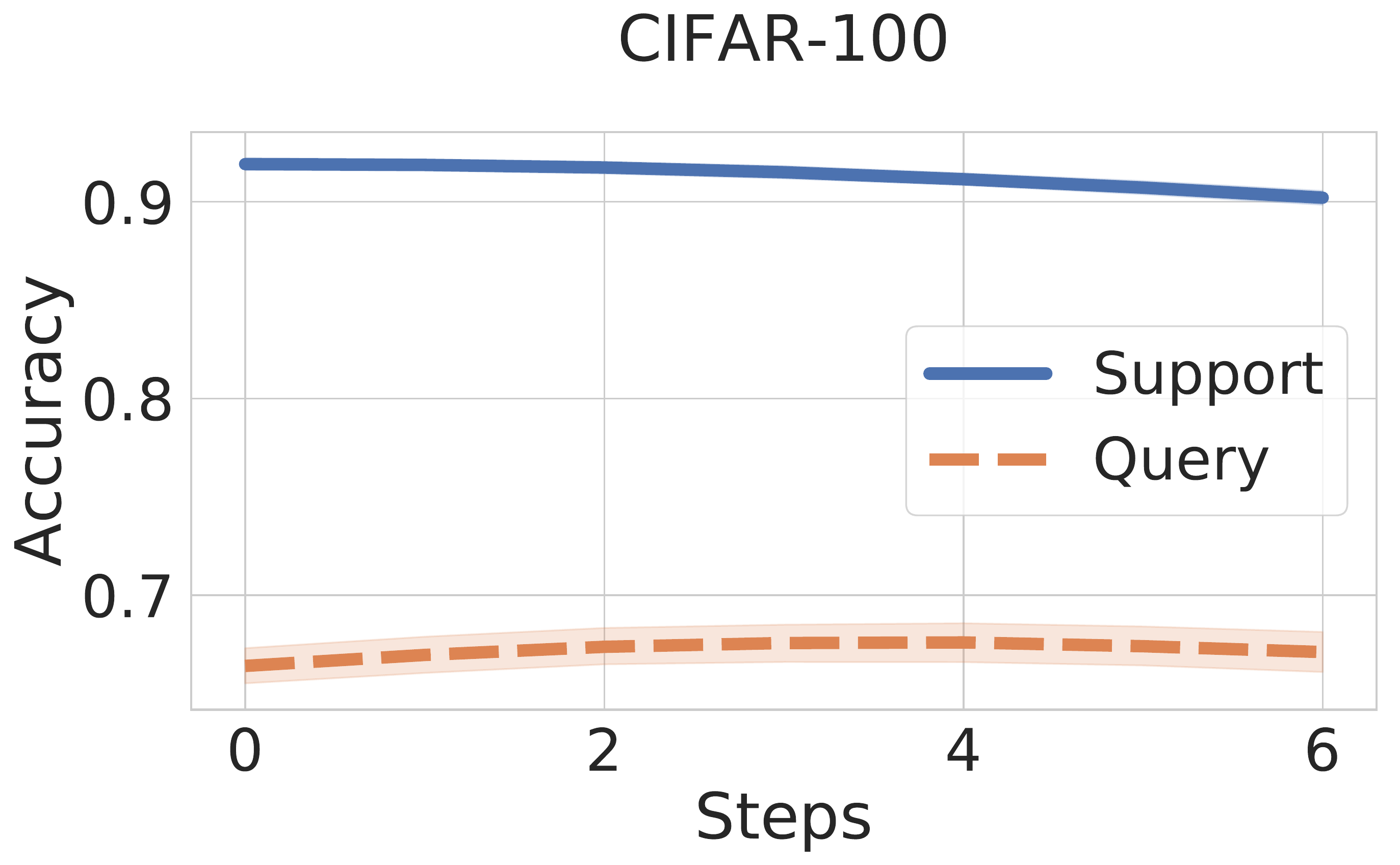}%
    \includegraphics[height=2.1cm]{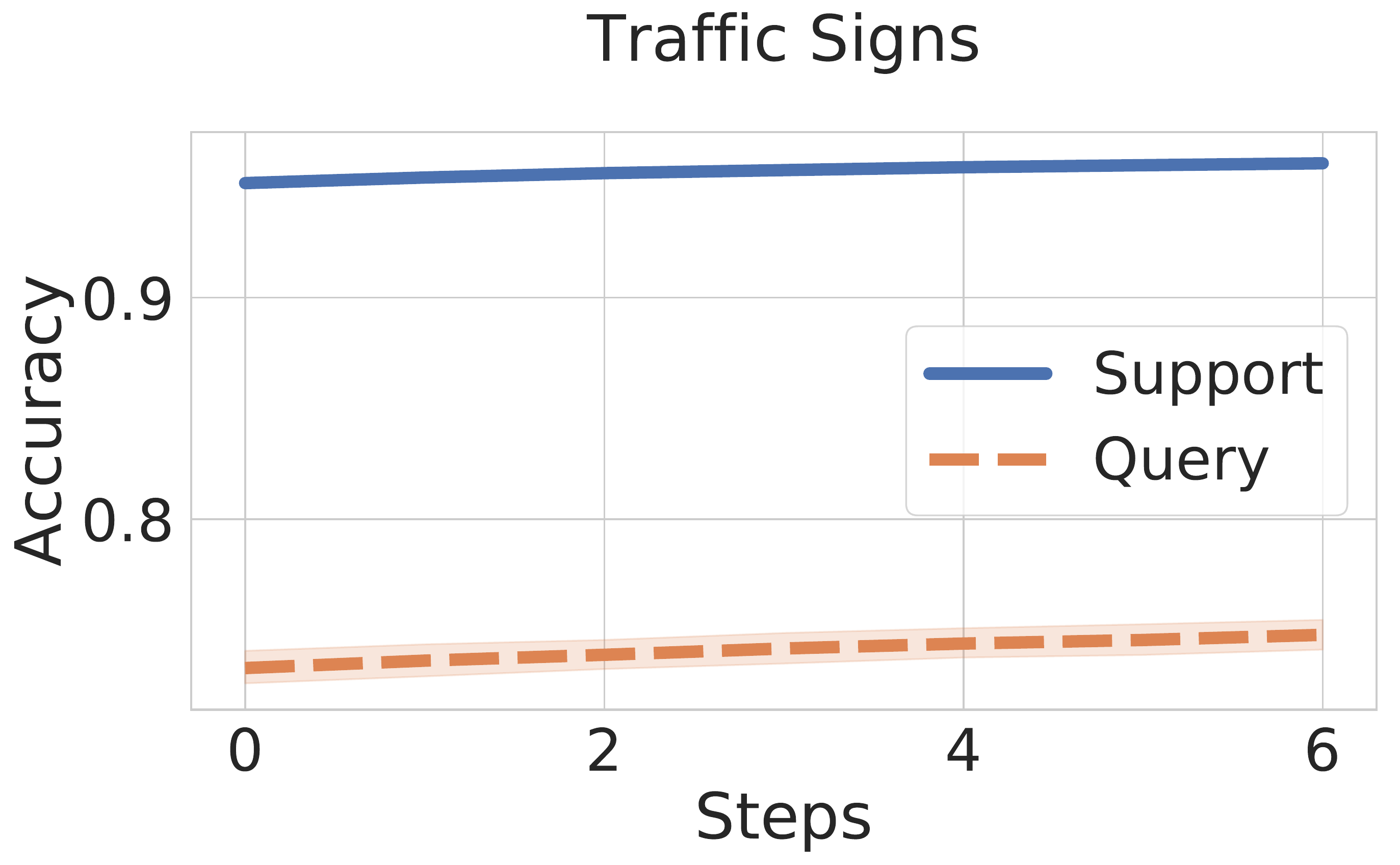}%
    \includegraphics[height=2.1cm]{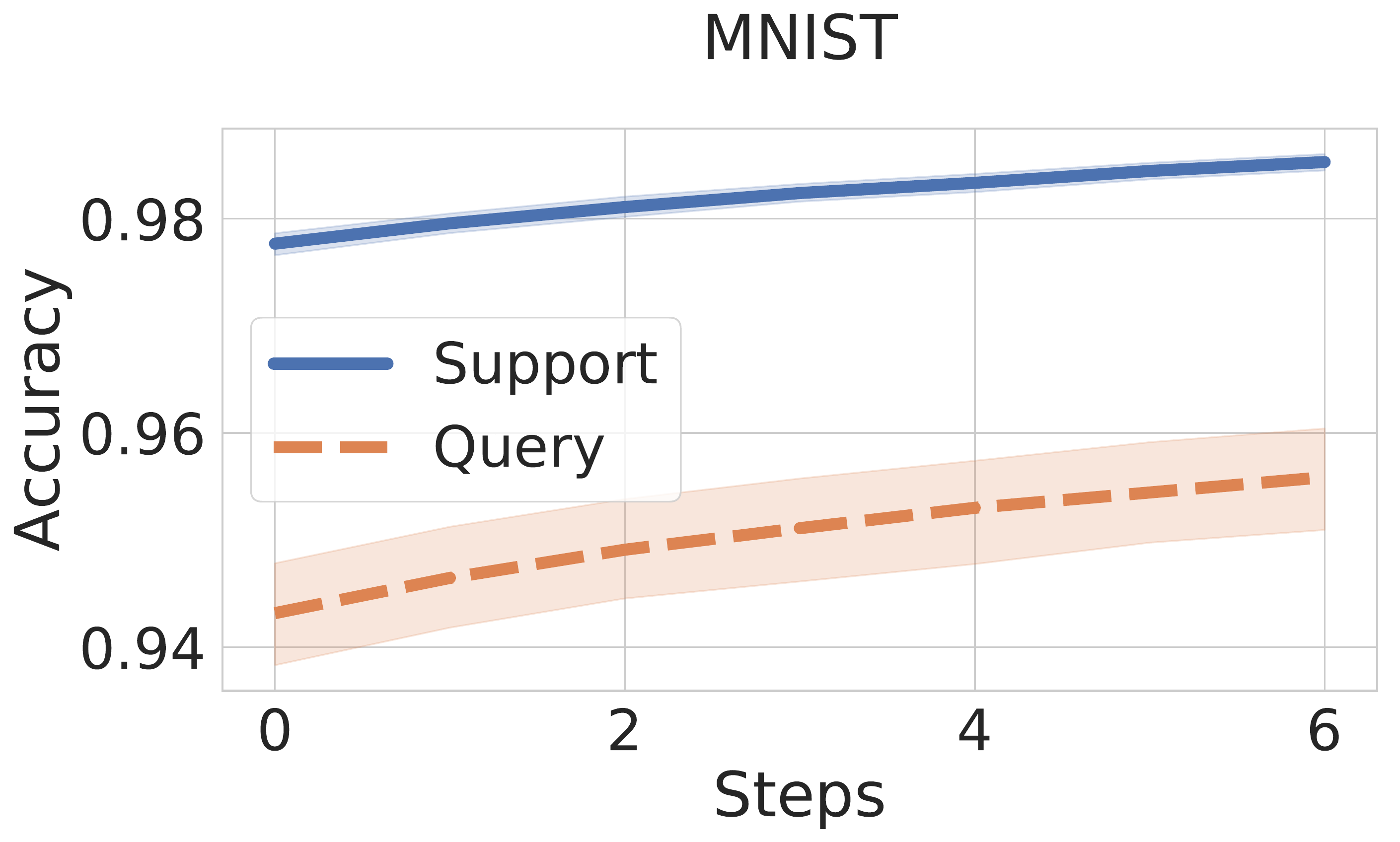}%
    \includegraphics[height=2.1cm]{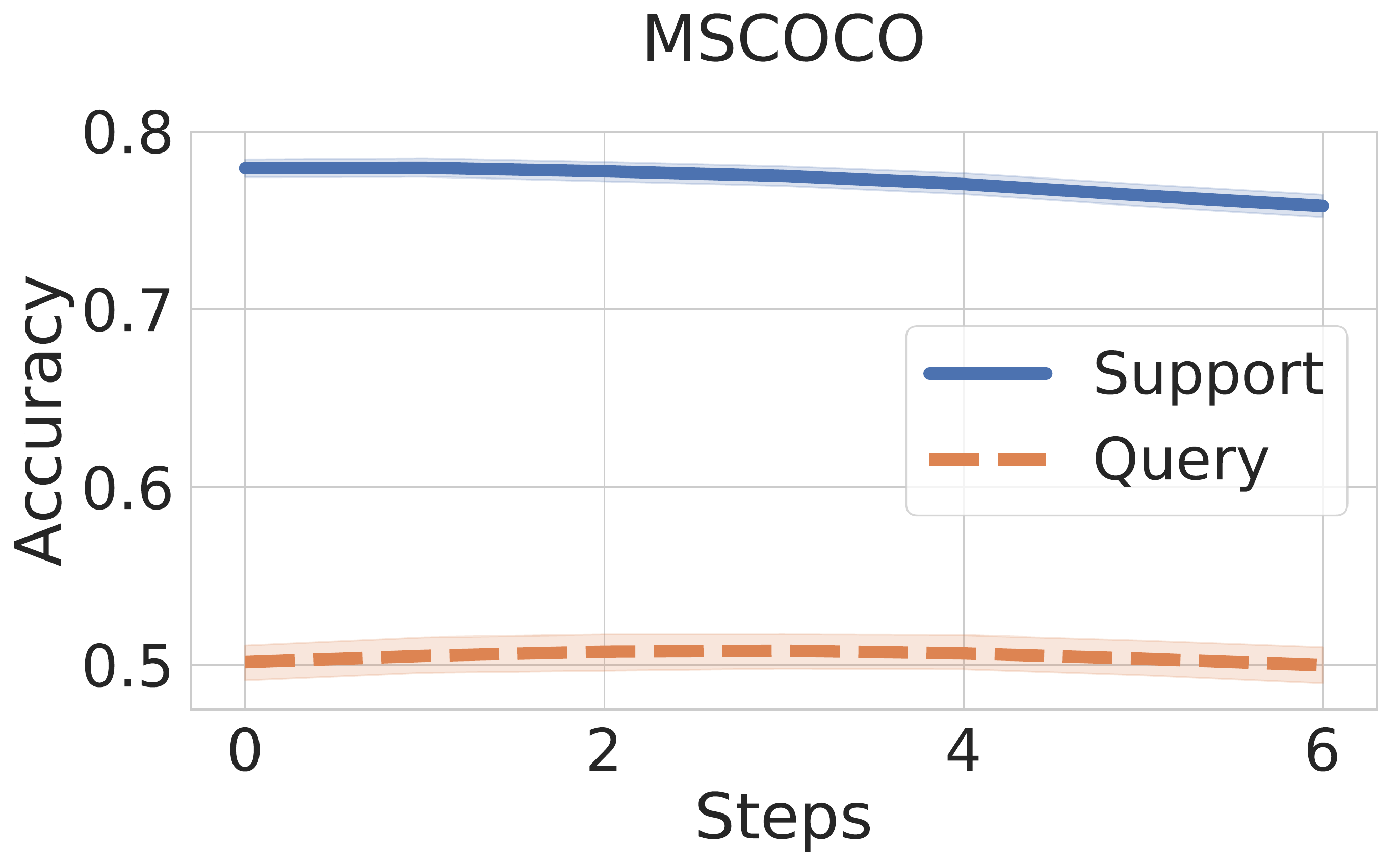}%
    \caption{The support and query accuracy over 600 test episodes of unseen datasets as a function of the fine-tuning steps for $\Psi_{d^*}$.}%
    \label{fig:fine_tuning}%
\end{figure*}

\paragraph{The importance of diverse training data}  Next, we empirically examine our hypothesis that training our feature extractor on \emph{all} training datasets is an important choice for encouraging (the convolutional layers of) our feature extractor to act as a universal template. In other words, had we not trained it on diverse data, we hypothesize it would be less general and thus less capable of supporting generalization to new datasets via merely changing the values of the batch normalization parameters. To investigate this, we trained a feature extractor (of identical architecture to FLUTE's) on ImageNet only. In this case, we only learn one set of FiLM parameters during training, since there is only one training dataset. We then attempt to tackle the test tasks similarly to FLUTE, by learning a new set of FiLM parameters for each, via one of the three initialization schemes proposed in the previous round of experiments.

In order to be able to use the Blender heuristic with this feature extractor too, we require a set of FiLM parameters for each of the training datasets. For the ImageNet dataset, we already have this: the FiLM parameters correspond to the batch normalization parameters obtained when training this feature extractor on ImageNet. For each of the remaining training datasets, we launch a training phase to learn a new set of batch normalization parameters, keeping the convolutional parameters frozen to those learned by this ImageNet-only feature extractor. This is reminiscent of SUR-pf's procedure for learning batch normalization parameters for the different training datasets, for use with an ImageNet-trained feature extractor. Once we have obtained the FiLM parameters for each of the training datasets for this ImageNet feature extractor, we use our dataset classifier within Blender to obtain the initialization of $\Psi_{d^*}$ as usual.

The performance of this ImageNet-only feature extractor, with different initialization heuristics for $\Psi_{d^*}$, are shown in the last three columns of Table~\ref{table:ablations_wo_ci}. We find that, for any given initialization heuristic, the feature extractor trained on all datasets performs significantly better across the board compared to that trained on ImageNet only, validating our hypothesis. 
Notably, the `Scratch' heuristic no longer performs well for datasets like Omniglot and Quickdraw, as it did when using FLUTE's feature extractor. This speaks to the inferior generalizability of this feature extractor that was trained on less diverse data. 
Finally, we observe that even when evaluating on test classes from ImageNet (first row), FLUTE's feature extractor evidently constitutes a better template compared to the ImageNet-trained one, indicating positive transfer during FLUTE's joint training phase.

\paragraph{The effect of the number of steps to train $\Psi_{d^*}$} Figure~\ref{fig:fine_tuning} plots the support and query accuracy during the fine-tuning of the per-task parameters $\Psi_{d^*}$ on the support set, averaged over 600 test tasks of each unseen domain (strong generalization setting). We also show the corresponding plots for weak generalization in the Appendix. We used 6 steps for this, with a learning rate of 0.005, which are the values we used for FLUTE's results in Table~\ref{table:main}, chosen based on the validation set. We observe that some datasets benefit more than others from this fine-tuning. In fact, in some cases (CIFAR-100, MSCOCO) the support accuracy even decreases. This indicates that perhaps those hyperparameters aren't ideal for all test tasks. Moreover, the validation set may not be a good proxy for the test set, since it is comprised of different datasets, making hyperparameter selection hard. Tailoring the learning rate or number of steps to each task would be an interesting avenue for future work.  

\begin{figure}%
    \centering%
    \includegraphics[width=\linewidth]{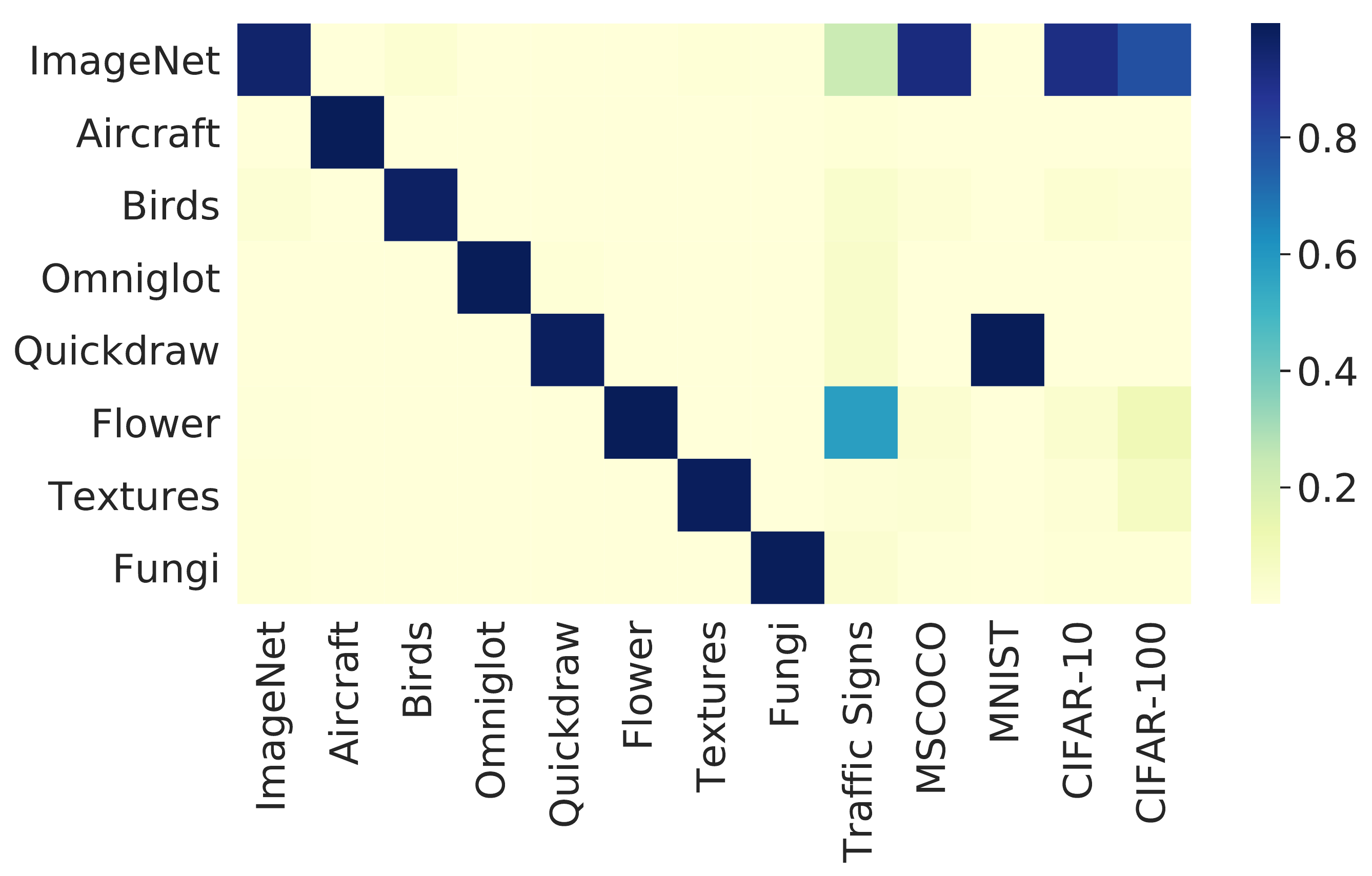}%
    \caption{\label{fig:heatmap} Visualization of the coefficients of the convex combination produced by the Blender network for various test datasets. Each column represents the average coefficients produced for 600 test tasks of a particular dataset.}%
\end{figure}

\paragraph{Inspecting the Blender's proposals} We visualize the combination coefficients that the Blender produces for test tasks in Figure~\ref{fig:heatmap}. We notice that for test episodes of datasets seen during training (weak generalization setting), the Blender very strongly selects the dataset from which the task originated. For new datasets that did not appear in the training set (last 5 columns), the Blender seems to make reasonable choices (e.g. Quickdraw for MNIST, ImageNet for MSCOCO) but the probability distribution may be less peaky (e.g. for Traffic Signs and CIFAR-100). In the Appendix we also investigate the variance of these predictions across different test tasks within each dataset.

\paragraph{The effect of taking a convex combination} Having inspected the combination weights that Blender produces, we noticed that it often relies almost exclusively on a single dataset's FiLM parameters to form the initialization of the FiLM parameters of a new task. A natural question to ask, therefore, is whether the convex combination is required. An alternative design choice is a `Hard Blender' that selects only the FiLM parameters of the most compatible training dataset (as assessed by the dataset classifier), instead of combining them all in a weighted manner. We provide a detailed comparison with this variant in Table~\ref{table:hard_blender_comparison} in the Appendix. Indeed, the two initialization schemes perform similarly, especially on weak generalization tasks, where the Blender's co-efficients already resemble a hard selection. We observe, though, that the combination approach may yield small gains on strong generalization tasks. Overall, since Blender is a more general approach and it may improve the performance on new datasets, we adopt this initialization scheme as our default one for FLUTE.


\paragraph{Discussion} As alluded to earlier, an open challenge is defining an appropriate model selection mechanism for the difficult scenario when the provided validation set is not a good proxy for the test set. The problem that we study in this work, similarly to domain generalization \cite{gulrajani2020search},
suffers from underspecification: there may be several solutions that all perform well on the training and validation sets but behave differently on the test set. With this in mind, perhaps we should adjust our evaluation metrics to account for this potential source of variance. For instance, in the context of FLUTE, we noticed in Table~\ref{table:ablations_wo_ci} that the initialization of $\Psi_{d^*}$ really matters, therefore perhaps the results are sensitive to the chosen checkpoint of the dataset classifier. We report in the Appendix the performance across more runs of that component and observe some variance on held-out datasets, but our conclusions hold across runs.
\section{Conclusion}
To conclude, we proposed FLUTE, a method for few-shot dataset generalization that creates a universal template, i.e. a partially-parameterized model, that can serve to readily specify a good feature extractor for a held-out dataset, by simply plugging in an appropriate set of FiLM parameters. We propose an efficient approach for estimating those parameters, using a few steps of gradient descent starting from a task-dependent initialization. FLUTE sets the new state-of-the-art on Meta-Dataset, significantly outperforming previous methods on few-shot dataset generalization.

\bibliography{references}
\bibliographystyle{icml2021}

\clearpage
\section*{Implementation Details}

\paragraph{Architecture} We use a ResNet-18 as our feature extractor, to be consistent with the previous work we compare against. The dataset classifier that is used in our Blender network is comprised of a permutation-invariant set encoder $g$ \cite{zaheer2017deep} followed by a linear layer $l$, as explained in our main paper. We adopt a similar architecture for $g$ to the one used in \cite{requeima2019fast} for their `adaptation networks'. This consists of 5 convolutional blocks, each of which is comprised of a 3x3 convolution operation with 64 channels, followed by batch normalization, ReLU, and 2x2 max-pooling with stride 2. We then apply global average pooling to the output, followed by averaging over the first dimension (i.e. over the different examples of the batch), to obtain our set encoding of the given batch. This vector is then fed into $l$ to classify the given batch into one of the $M$ training datasets.

\paragraph{Training the shared and per-dataset parameters} We train FLUTE via a joint phase that utilizes data from all $M$ training datasets in order to learn a universal template $\Phi$ and M per-dataset sets of FiLM parameters $\Psi_1 \dots \Psi_M$. As detailed in the main paper, our training objective is a multi-task classification one, that requires $M$ per-dataset classification readout heads. Following recent work \cite{chen2019closer,chen2020new,dvornik2020selecting}, we treat each of those readout heads as a \textit{cosine classifier}, i.e. a layer without a bias, parameterized only by a weight matrix, where the activations that are the inputs to the layer, as well as the rows of that matrix are L2-normalized before the matrix multiplication is performed. Following those previous works, we also utilize a learnable softmax temperature for these cosine classifiers.

We use stochastic gradient descent with a momentum of 0.9, with a cosine decay schedule with restarts for the learning rate. We also applied weight decay to the parameters of the convolutional layers and to the FiLM parameters. We tuned these parameters on the validation set, and used a starting learning rate of 0.01. The first round decays over 10000 steps from that starting learning rate to ``alpha'' (we use the default value of 0 for ``alpha''). Then, a warm restart is performed, where the learning rate is now ``m mul'' times smaller than our original starting learning rate (we use the default value of 1 for ``m mul''), and the decay is done over ``t mul'' times more steps than the previous decay round (we use the default value of 2 for ``t mul''). We set the weight decay parameter for the convolutional layers to $7e-4$, and the weight decay for the FiLM parameters to $0.001$, which regularizes the network's $\beta$ offset parameters to 0 and the $\gamma$ scaling parameters to 1 (i.e. for the $\gamma$, we apply the weight decay to $(\gamma - 1)$). Following previous work \cite{chen2020new}, during our joint training phase, we sample examples from ImageNet half the time, with the other half being devoted to examples from all training datasets uniformly.

\paragraph{Training the dataset classifier} To train our dataset classifier, we use Adam with a cosine decay schedule for the learning rate, without restarts. The values that worked best for this (as per the validation set performance) were an initial learning rate of 0.001 that is decayed over 3000 steps. Note that this phase is significantly shorter compared to the previously-described phase that trains our feature extractor. We early-stopped the training of the dataset classifier based on the validation accuracy: specifically, this is the accuracy on the $M$-way dataset classification task computed on the validation set, which contains held-out classes of the $M$ training datasets, as explained in the main paper.

\paragraph{Fine-tuning $\Psi_{d^*}$} During evaluation, the fine-tuning phase within each test task also uses Adam as the optimizer, without any learning rate decay in this case. We tuned the learning rate and the number of fine-tuning steps based on episodes from the validation set. Our best variant used a learning rate of $0.005$ and 6 steps. The values we considered for these were $0.0005,0.001,0.005$ for the learning rate and $1,2,3,4,5,6,7,8,9,10,150,20,30$ for the number of steps.

\paragraph{Hypothesis testing}
We follow the same procedure as in \citep{triantafillou2020meta} to compute ranks for different methods that in turn determine which entries to bold in our tables. Specifically, we perform a 95\% confidence interval statistical test on the difference between the mean accuracies of pairs of entries of each row. If for two entries we are not able to reject the null hypothesis that the difference between their means is 0, they will receive the same rank. For example, if model A and model B are tied for the first place according to that test, they will each receive the rank 1.5 (the average of the ranks 1 and 2). If we are able to reject that hypothesis, however, the entry with the larger mean accuracy will receive a higher rank than the other. In each row, we bold the entries that are tied for the highest rank.

\section*{The effect of the number of steps to train $\Psi_{d^*}$} 
\begin{figure*}[t]
    \centering
    \subfigure{\includegraphics[width=.3\textwidth]{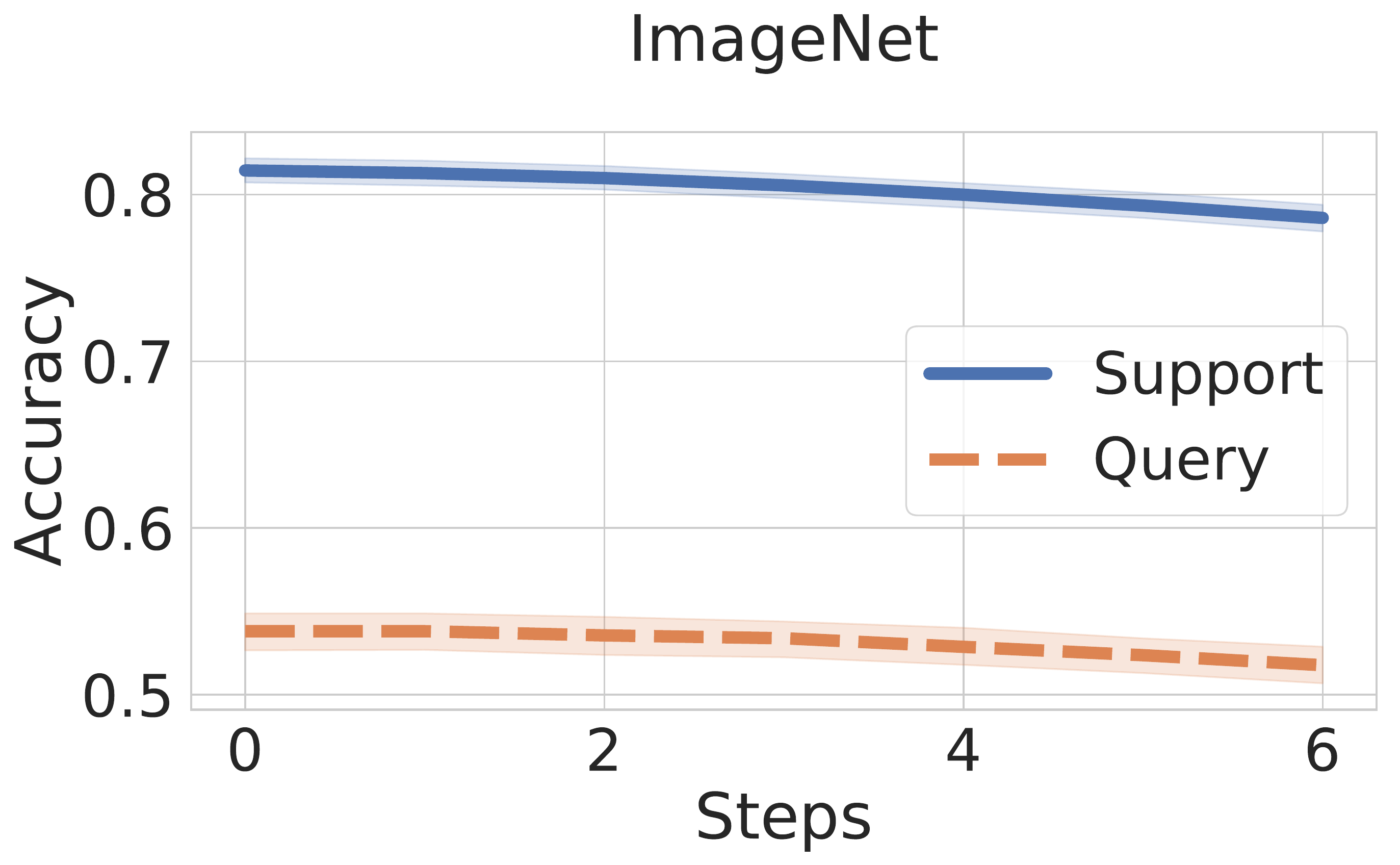}}
    \subfigure{\includegraphics[width=.3\textwidth]{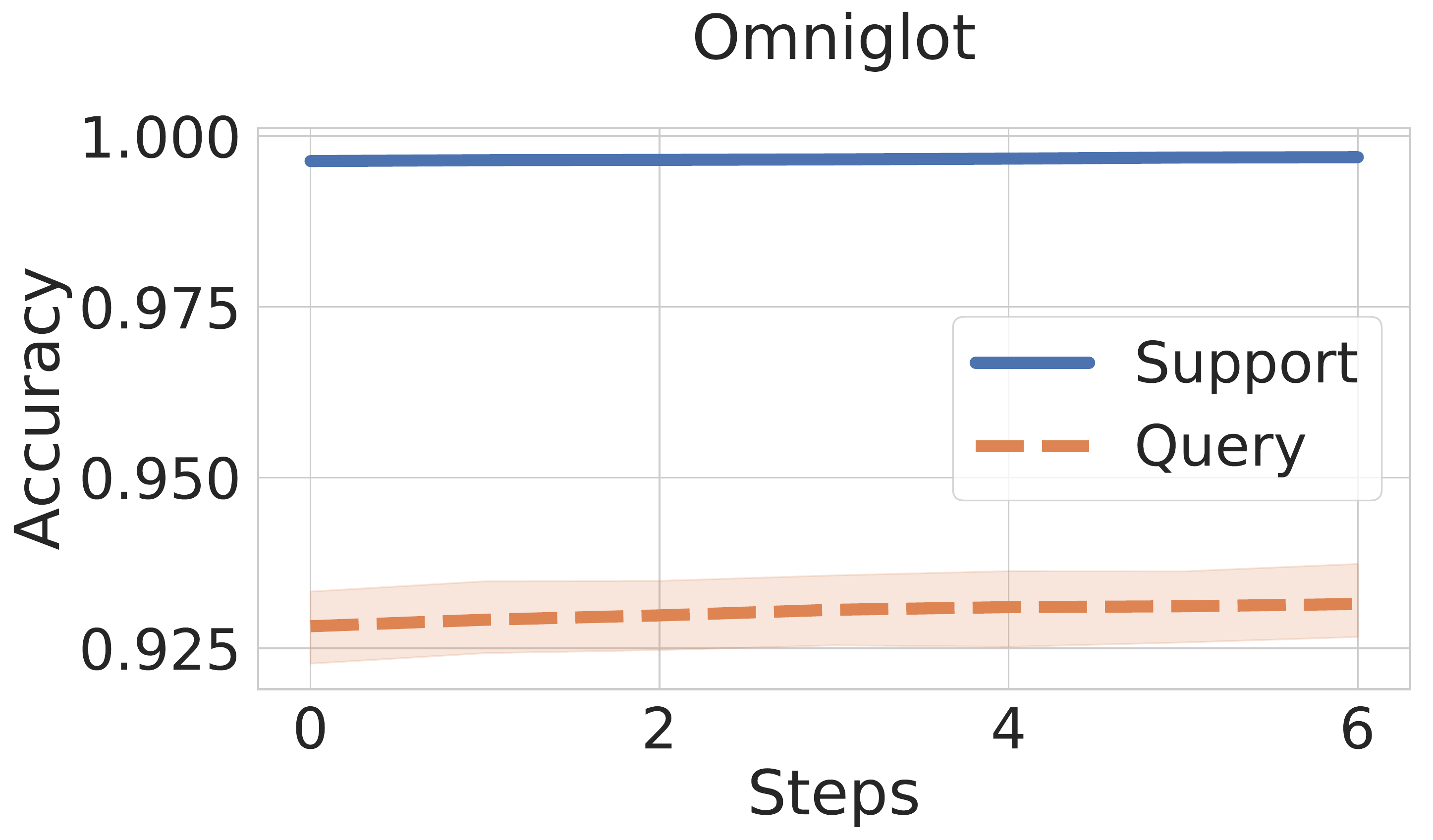}}
    \subfigure{\includegraphics[width=.3\textwidth]{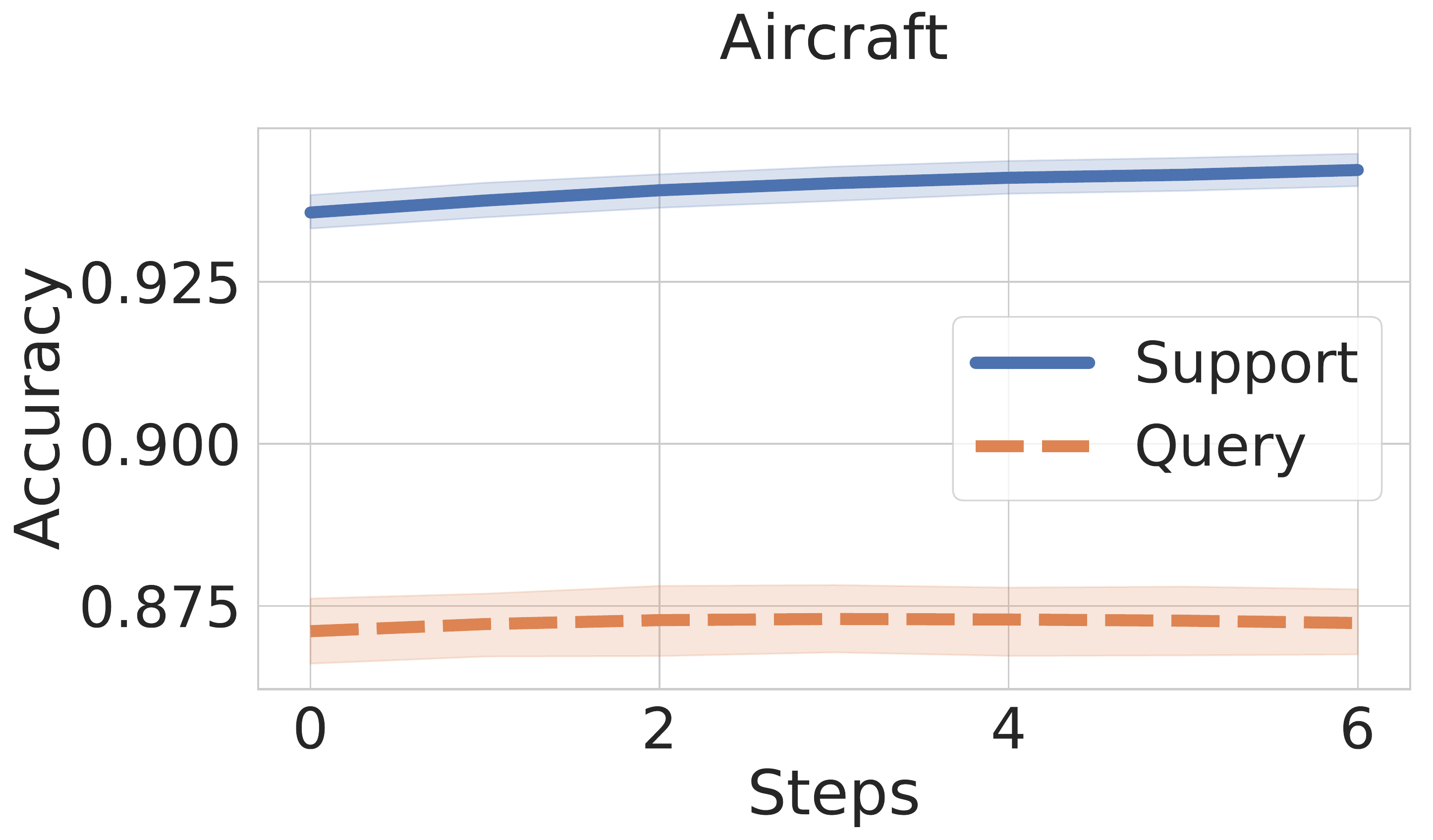}}
    \subfigure{\includegraphics[width=.3\textwidth]{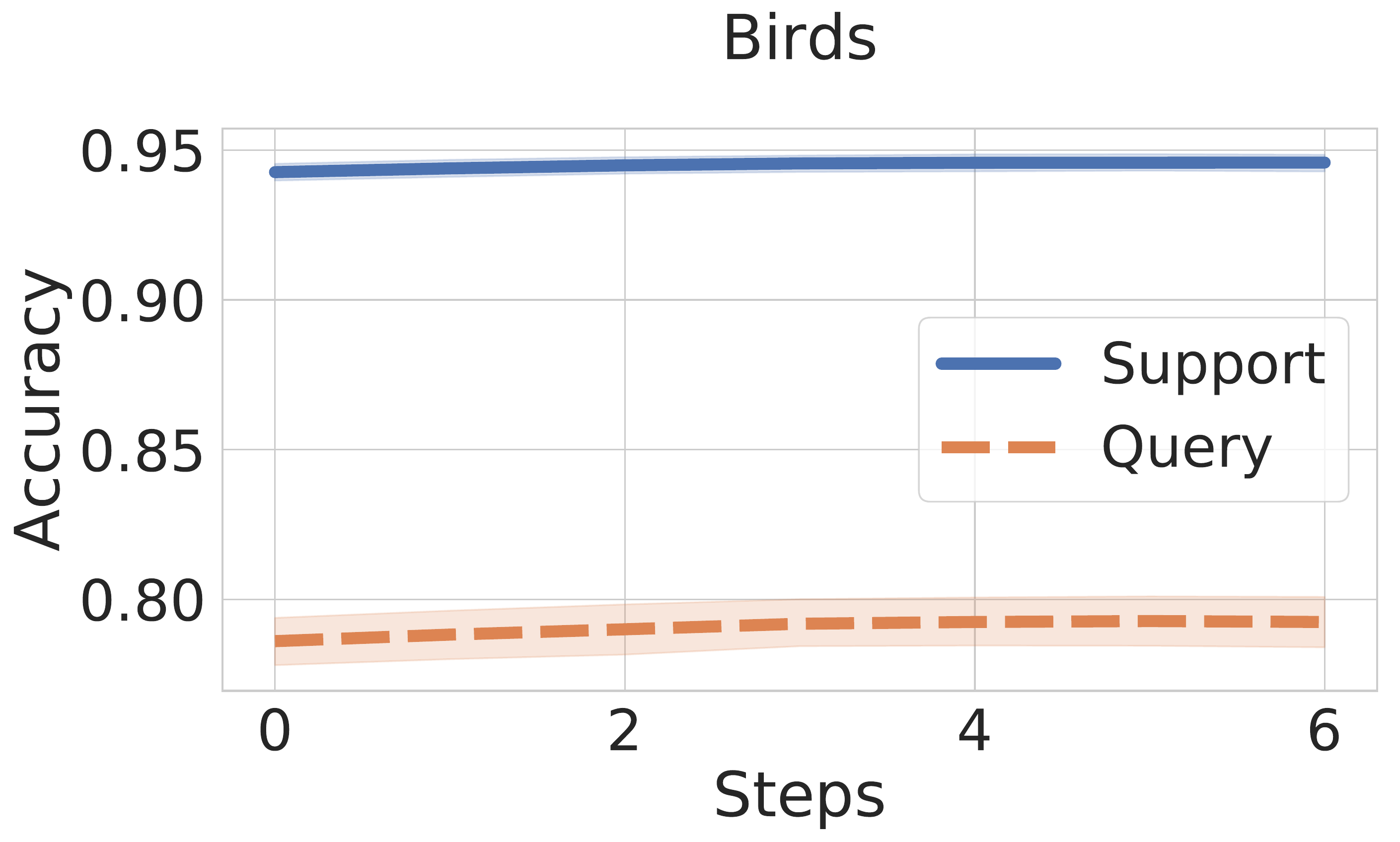}}
    \subfigure{\includegraphics[width=.3\textwidth]{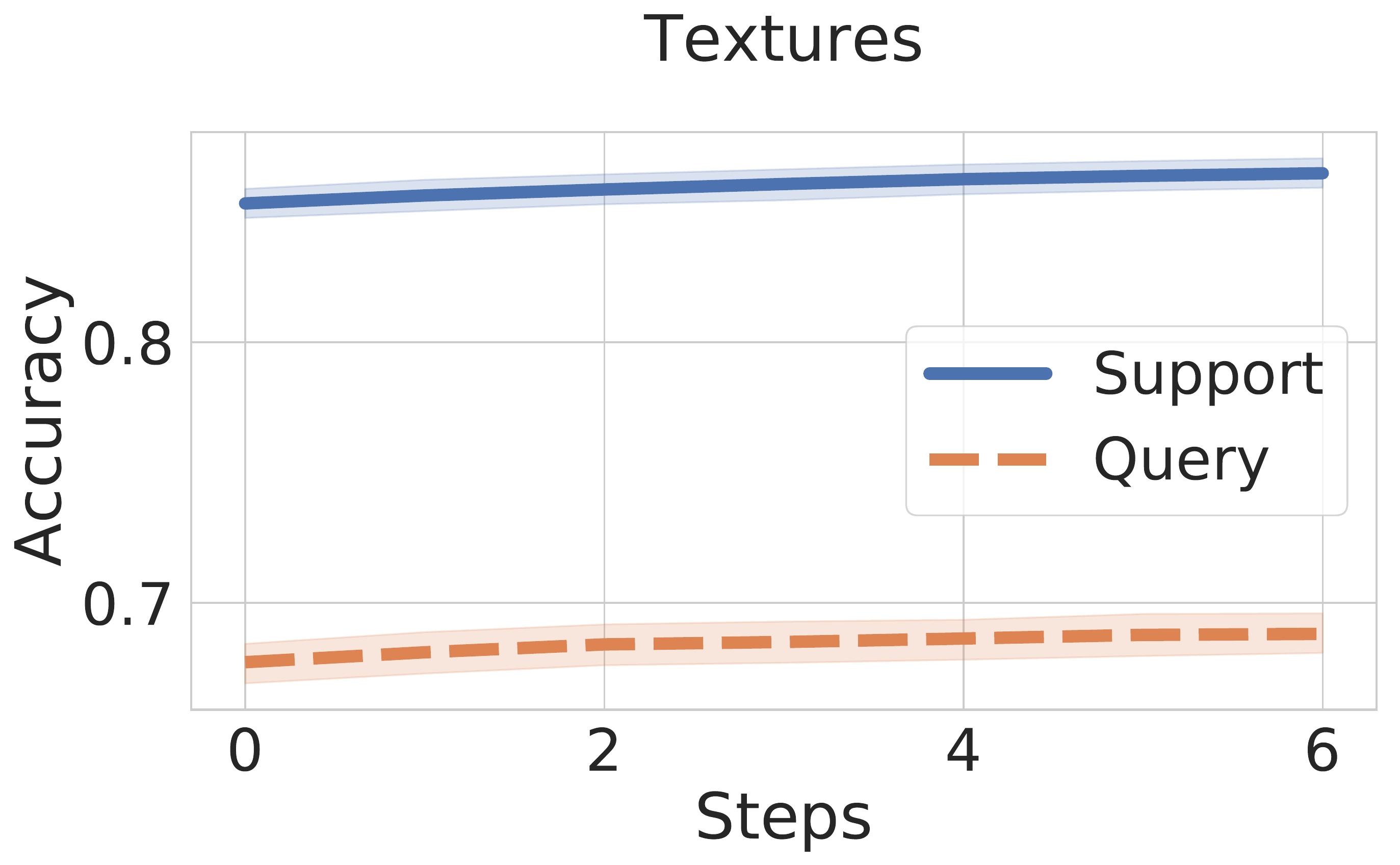}}
    \subfigure{\includegraphics[width=.3\textwidth]{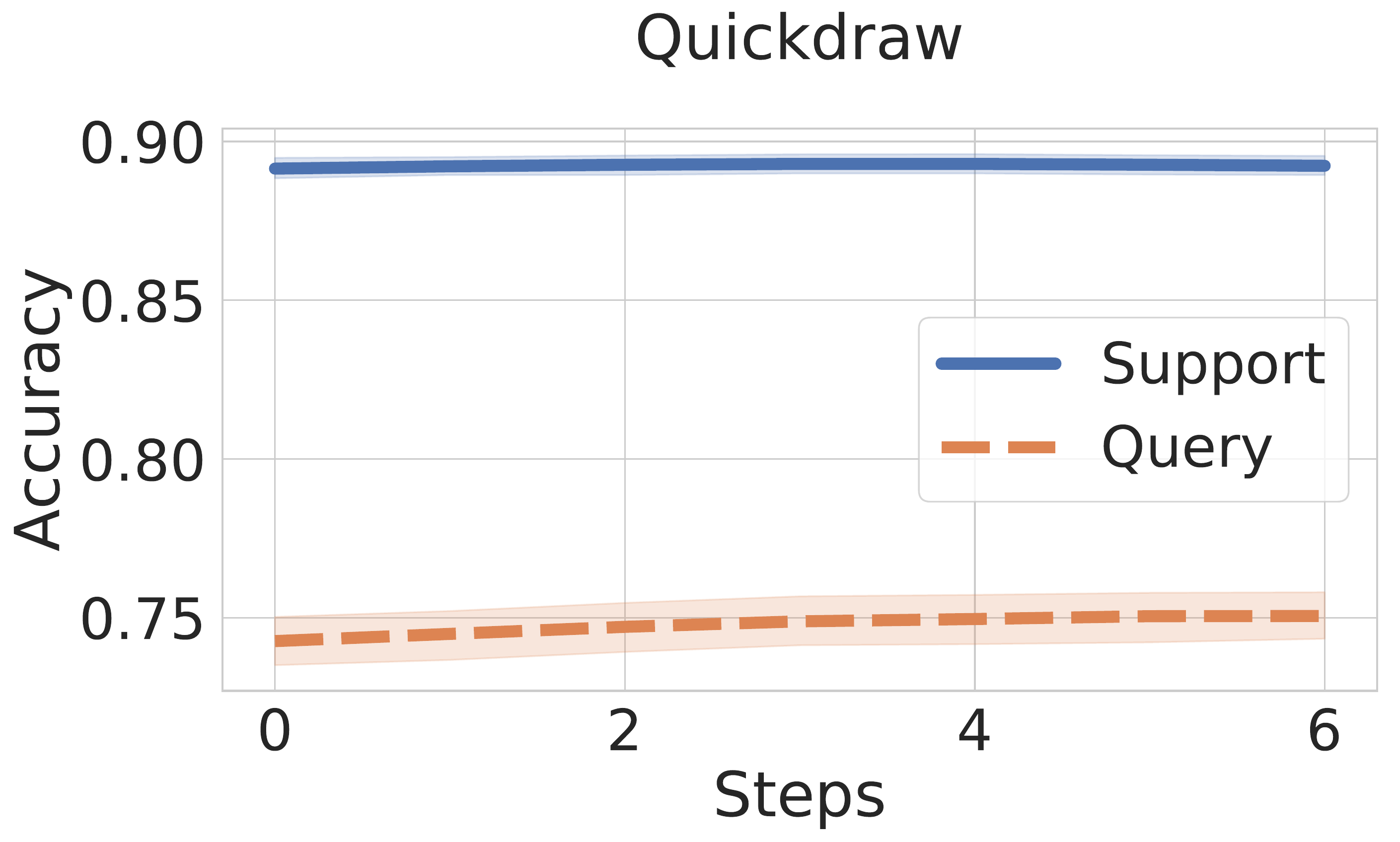}}
    \subfigure{\includegraphics[width=.3\textwidth]{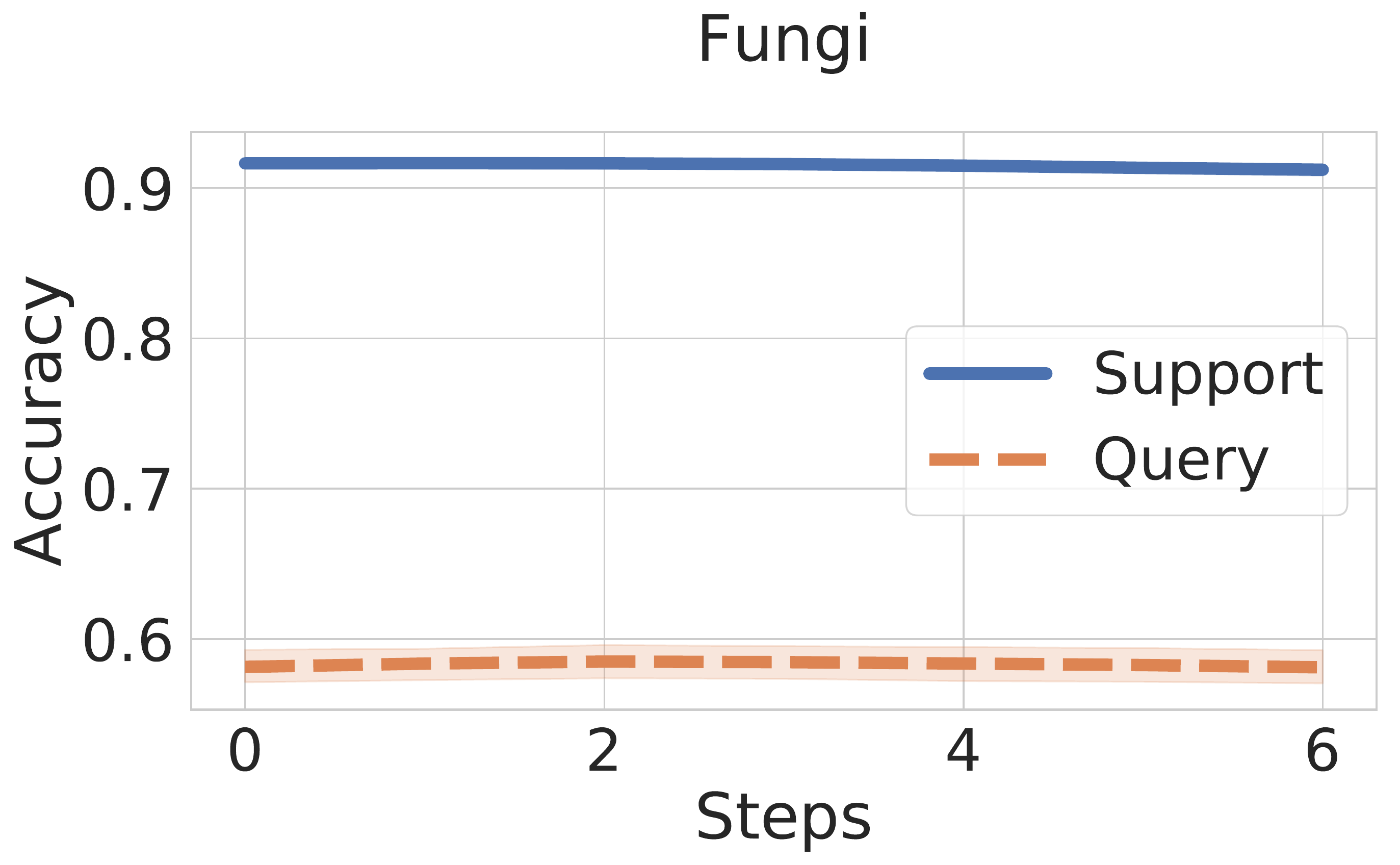}}
    \subfigure{\includegraphics[width=.3\textwidth]{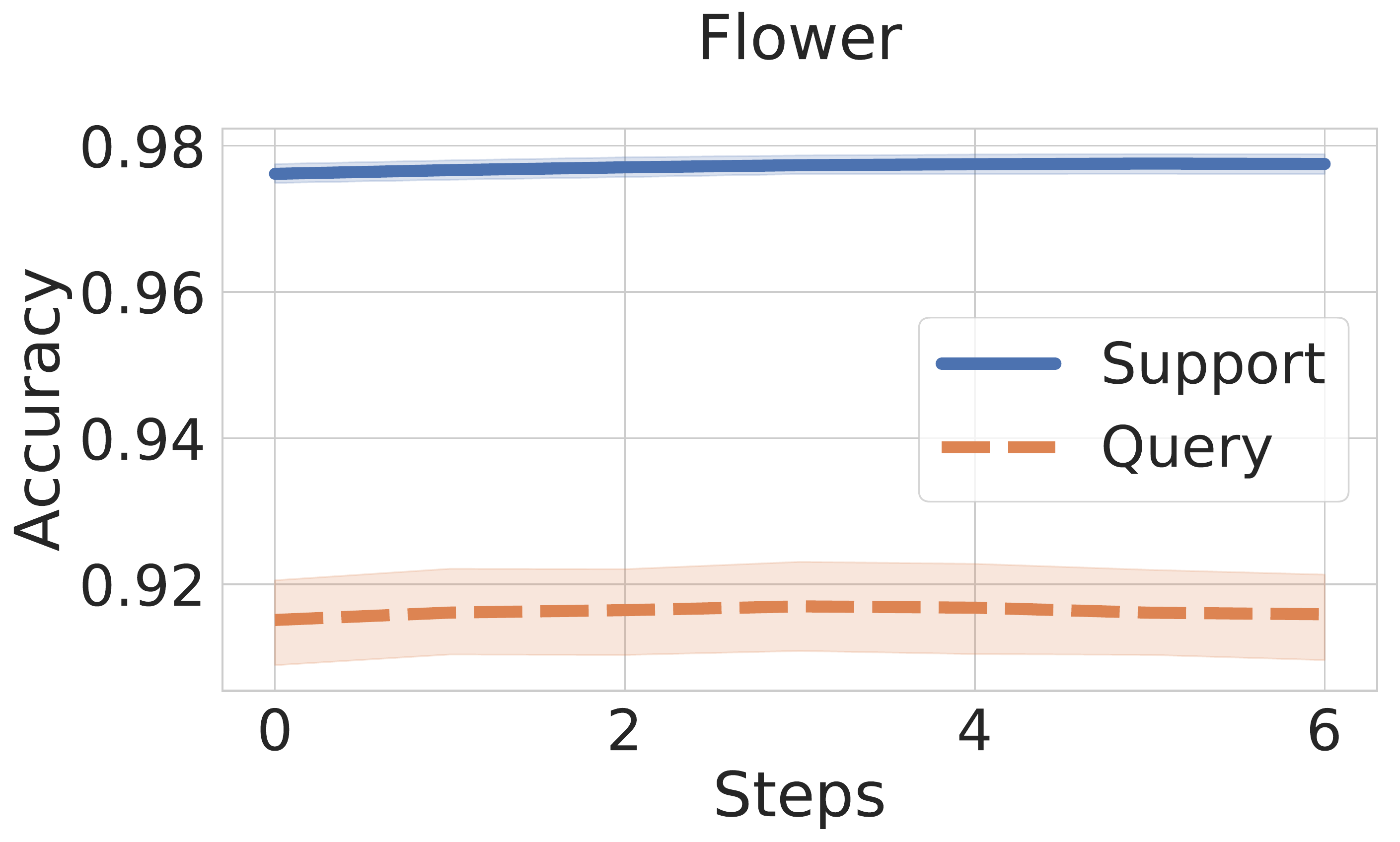}}
    \caption{The support and query accuracy over 600 test episodes of seen datasets (weak generalization setting) as a function of the fine-tuning steps for $\Psi_{d^*}$.}
    \label{fig:fine_tuning_weak}
\end{figure*}

In Figure~\ref{fig:fine_tuning} of the main paper we visualized the performance (on the support and query sets) for test episodes of held-out datasets throughout the fine-tuning of $\Psi_{d^*}$. For completeness, we also present in Figure~\ref{fig:fine_tuning_weak} the same result, but for test episodes of seen datasets (weak generalization setting). We observe that the increase in accuracy is less pronounced for these seen datasets. This is expected, since we know that there already exists a set of FiLM parameters that performs well for each test task sampled from a training dataset $m$ (namely the set $\Psi_m$ of FiLM parameters), and assuming the dataset classifier is accurate, the Blender would almost exclusively pick that set of FiLM parameters.

\section*{Inpsecting the Blender's proposals} In Figure~\ref{fig:heatmap} in the main paper, we visualized the average combination co-efficients that the Blender produces for test tasks of different datasets. Since that figure only shows the average, we now take a closer look at the distribution of the Blender's proposed combination co-efficients differs \textit{within} different test tasks of the same dataset. As a reminder, these co-efficients are computed based on the support set of each given test task, so they are not re-used across different tasks of the same dataset (in fact, the dataset identity is not known at test time).

\begin{figure*}[t]
    \centering
    \subfigure{\includegraphics[width=.3\textwidth]{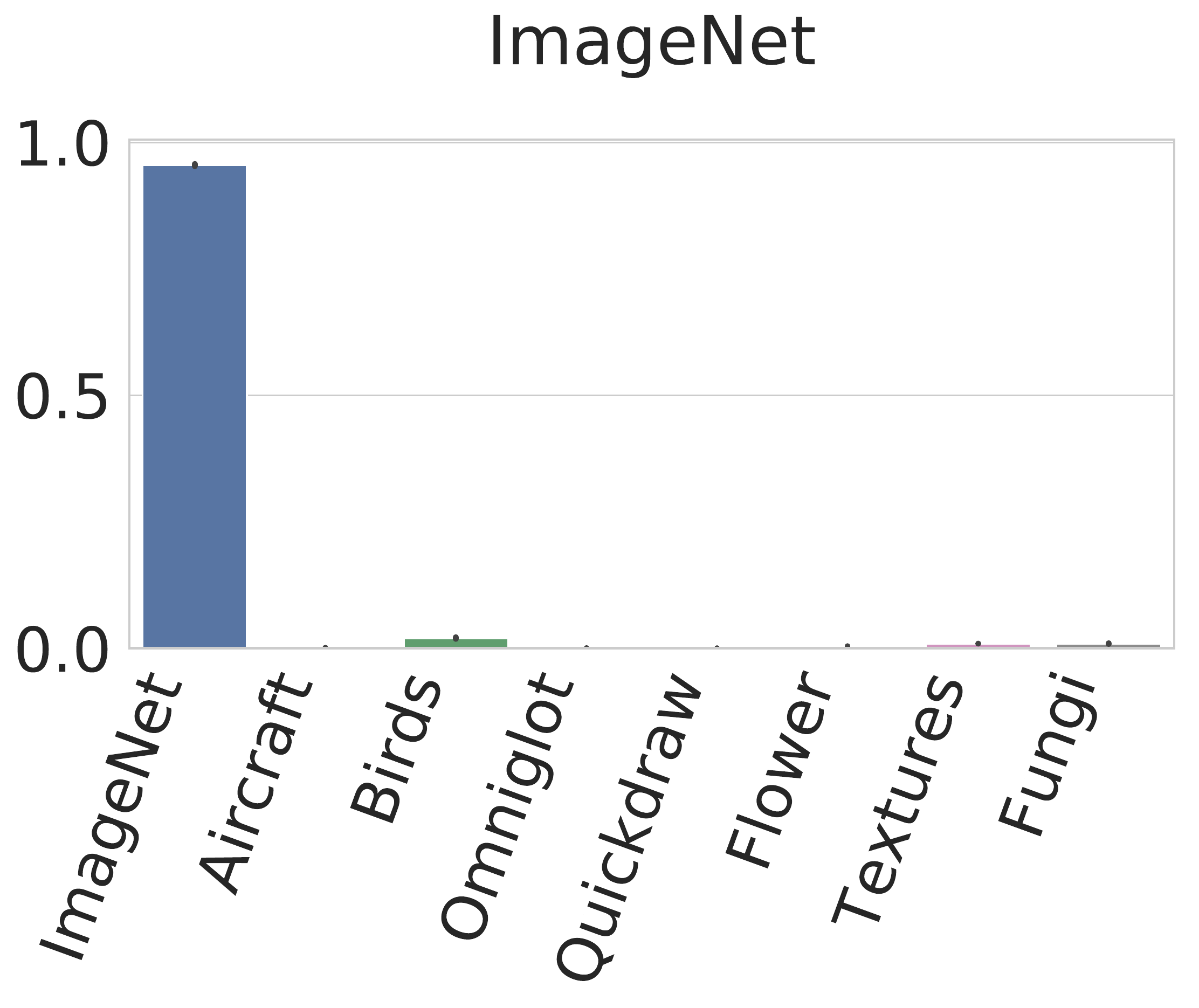}}
    \subfigure{\includegraphics[width=.3\textwidth]{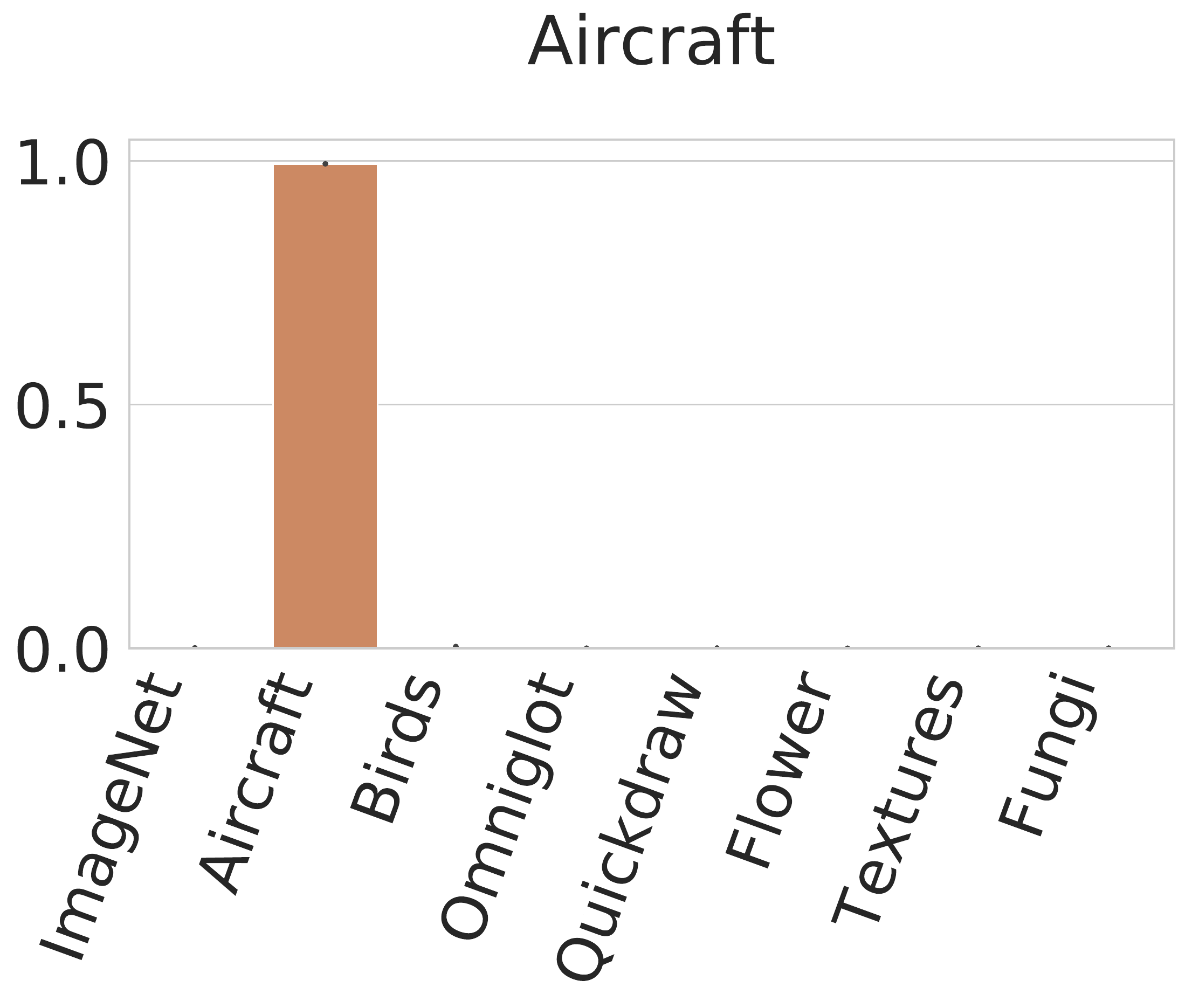}}
    \subfigure{\includegraphics[width=.3\textwidth]{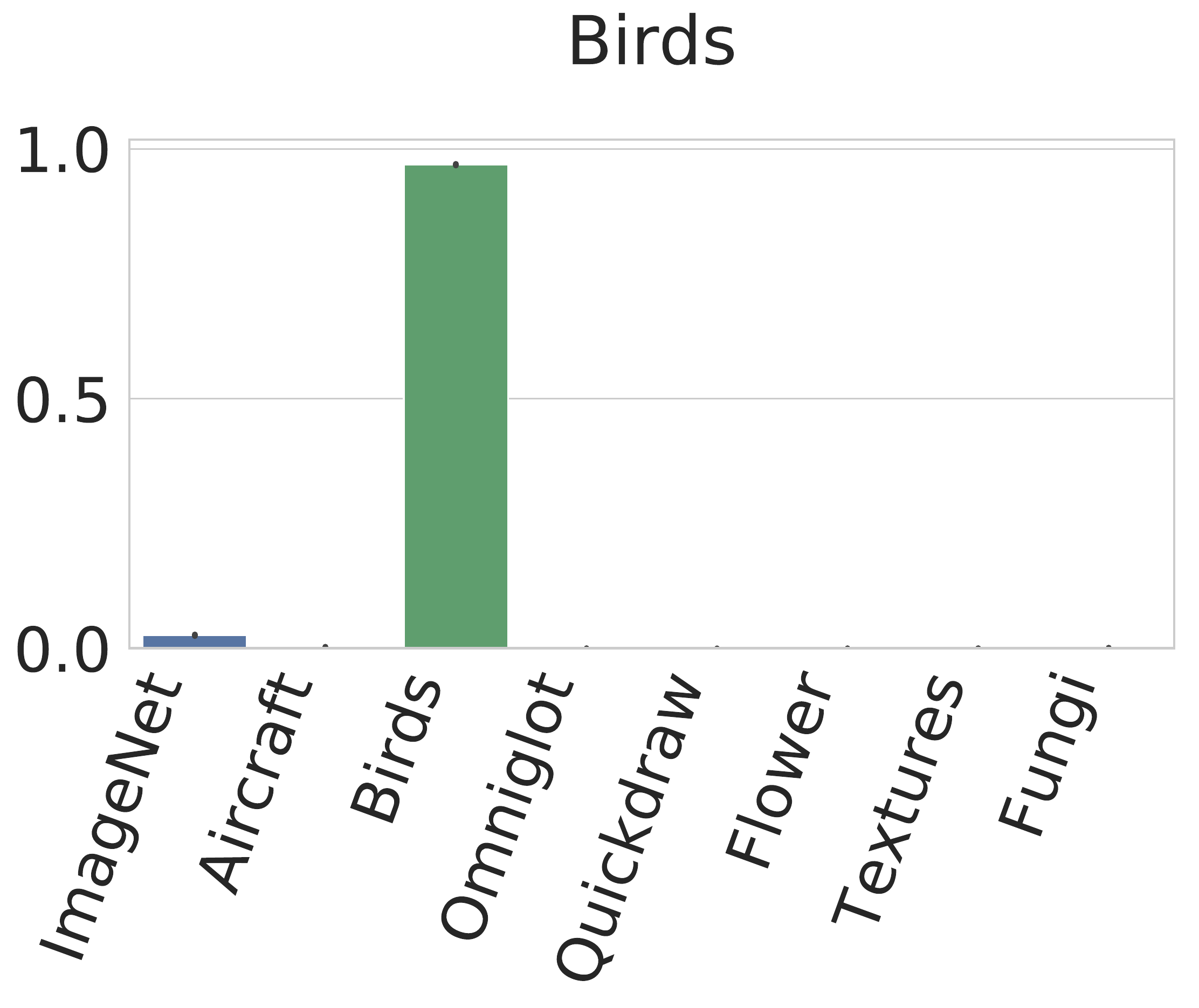}}
    \subfigure{\includegraphics[width=.3\textwidth]{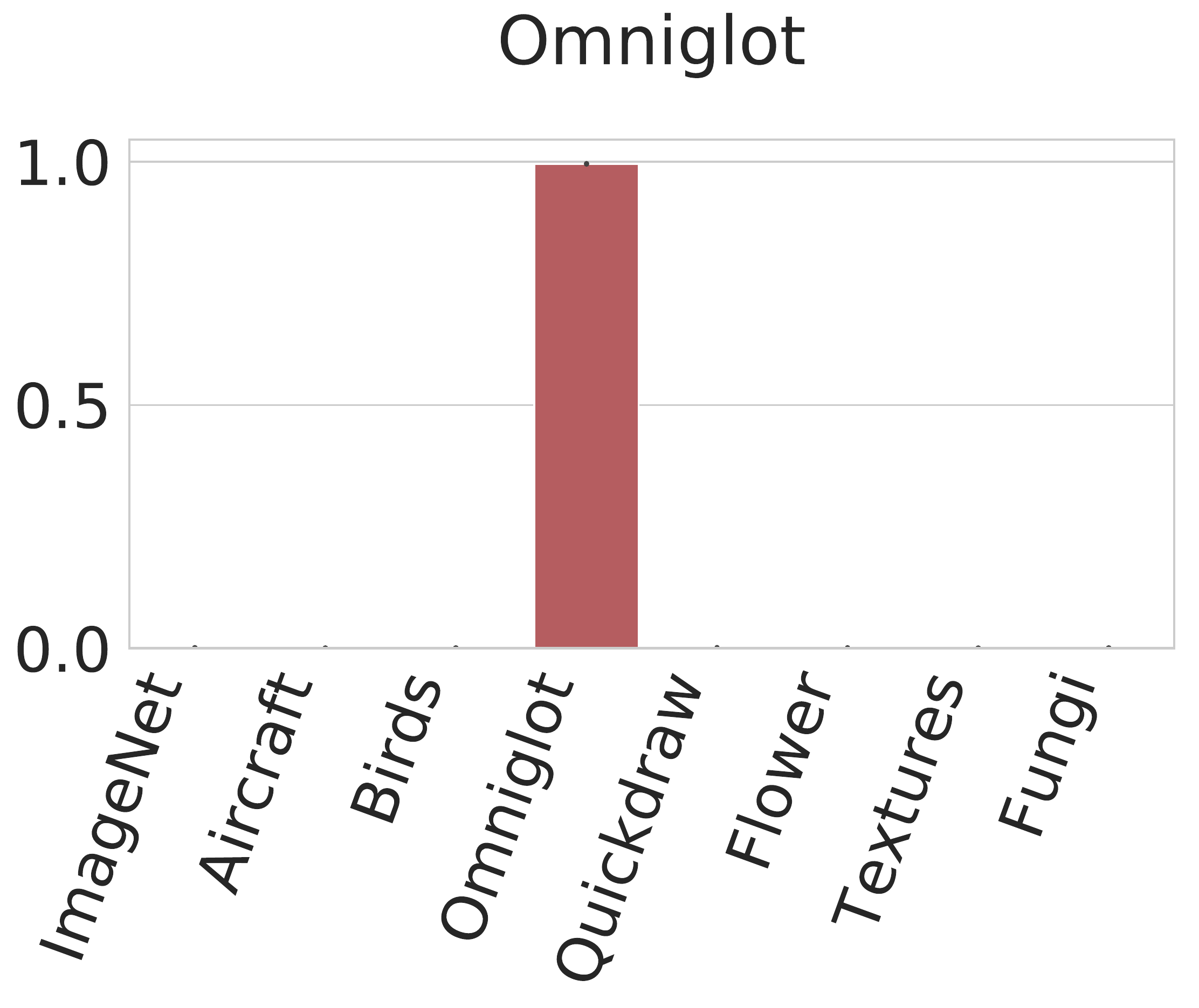}}
    \subfigure{\includegraphics[width=.3\textwidth]{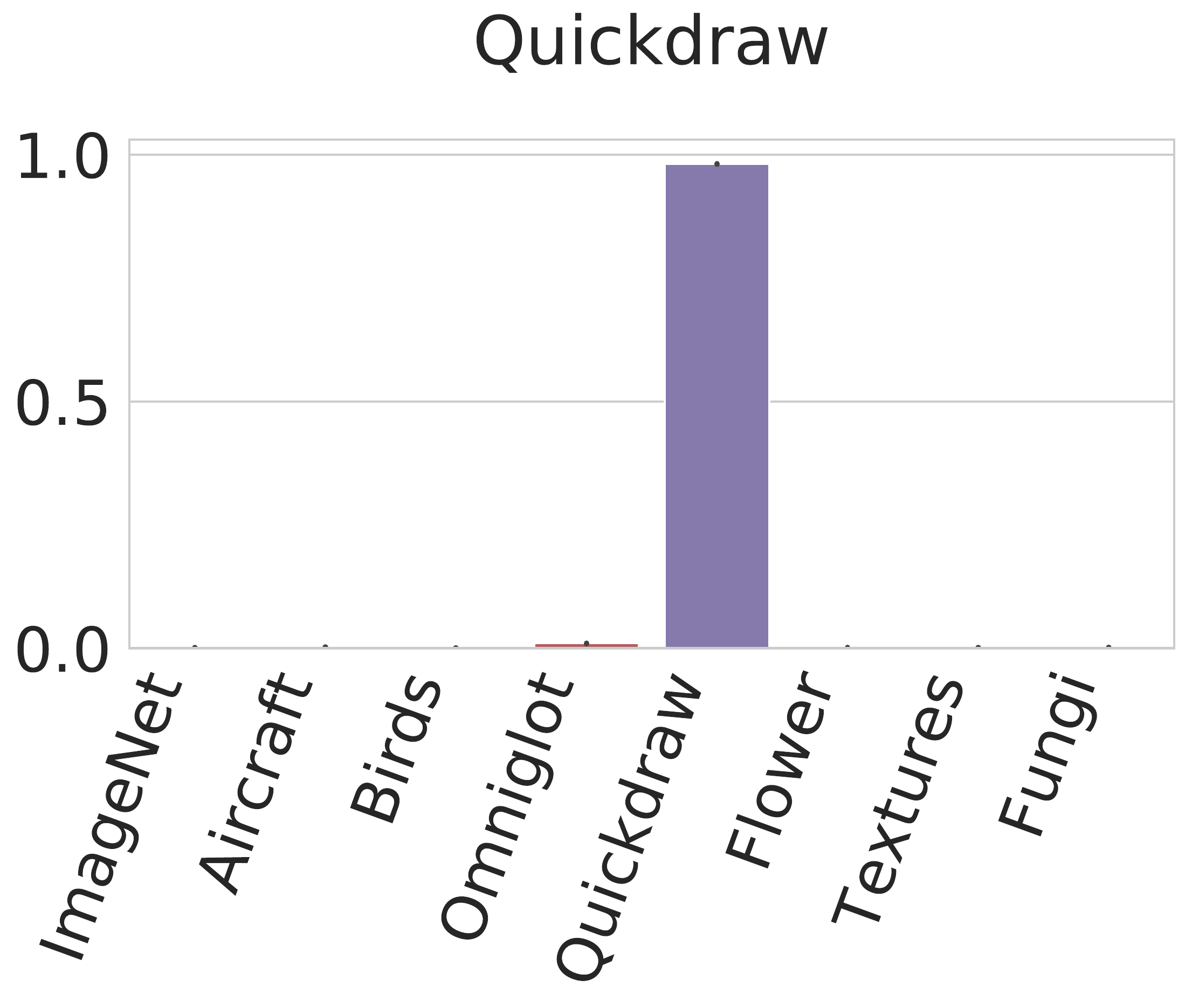}}
    \subfigure{\includegraphics[width=.3\textwidth]{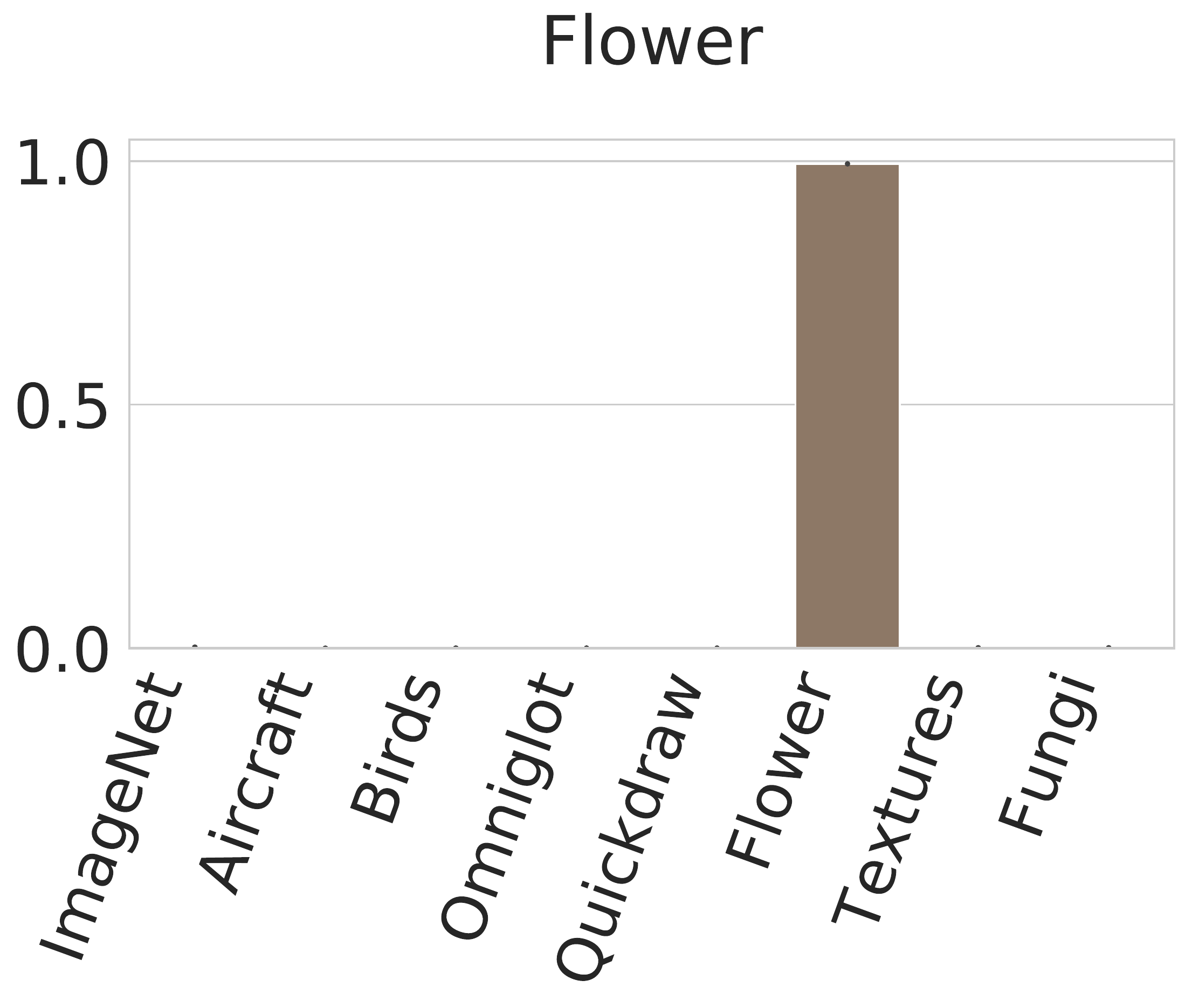}}
    \subfigure{\includegraphics[width=.3\textwidth]{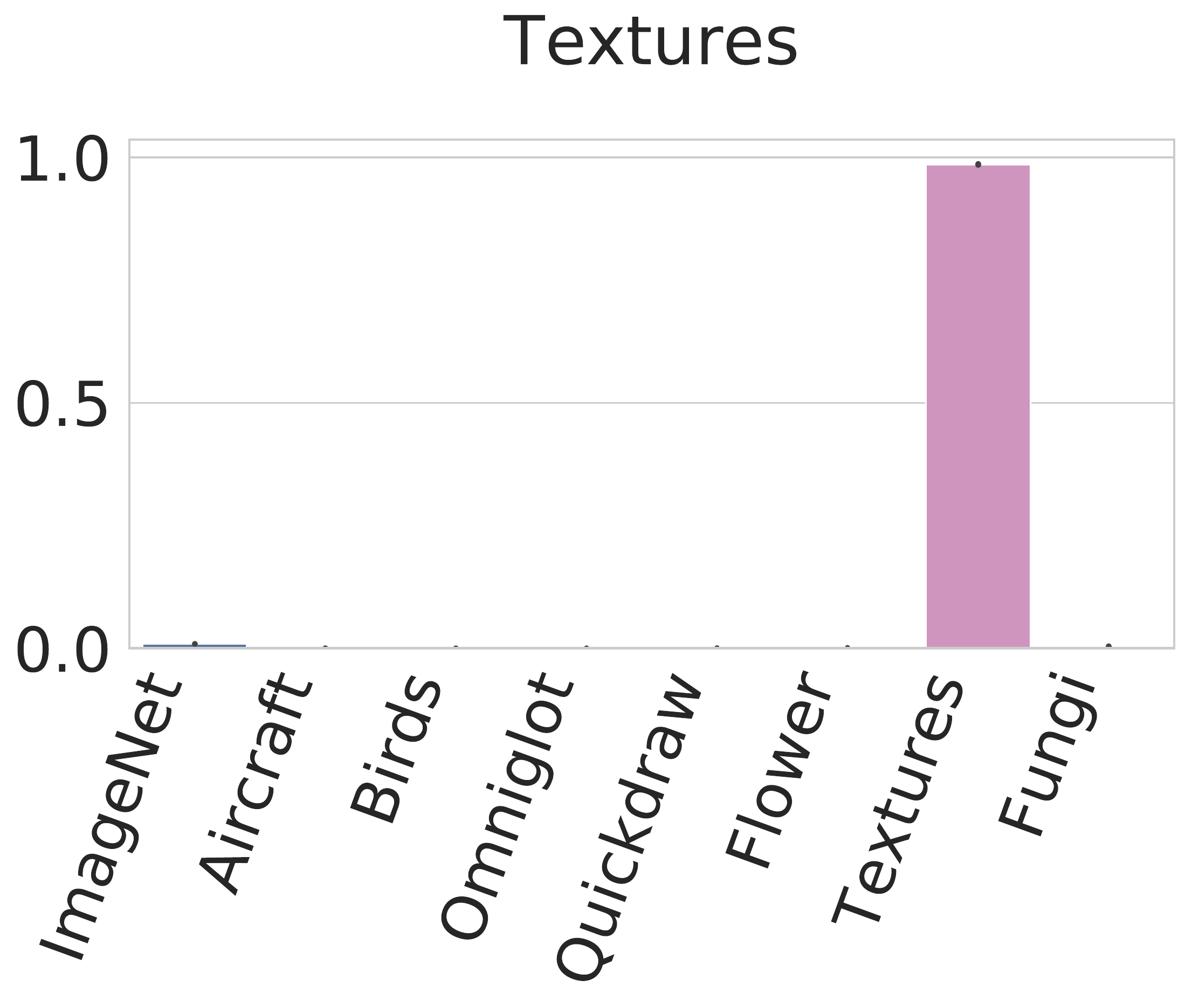}}
    \subfigure{\includegraphics[width=.3\textwidth]{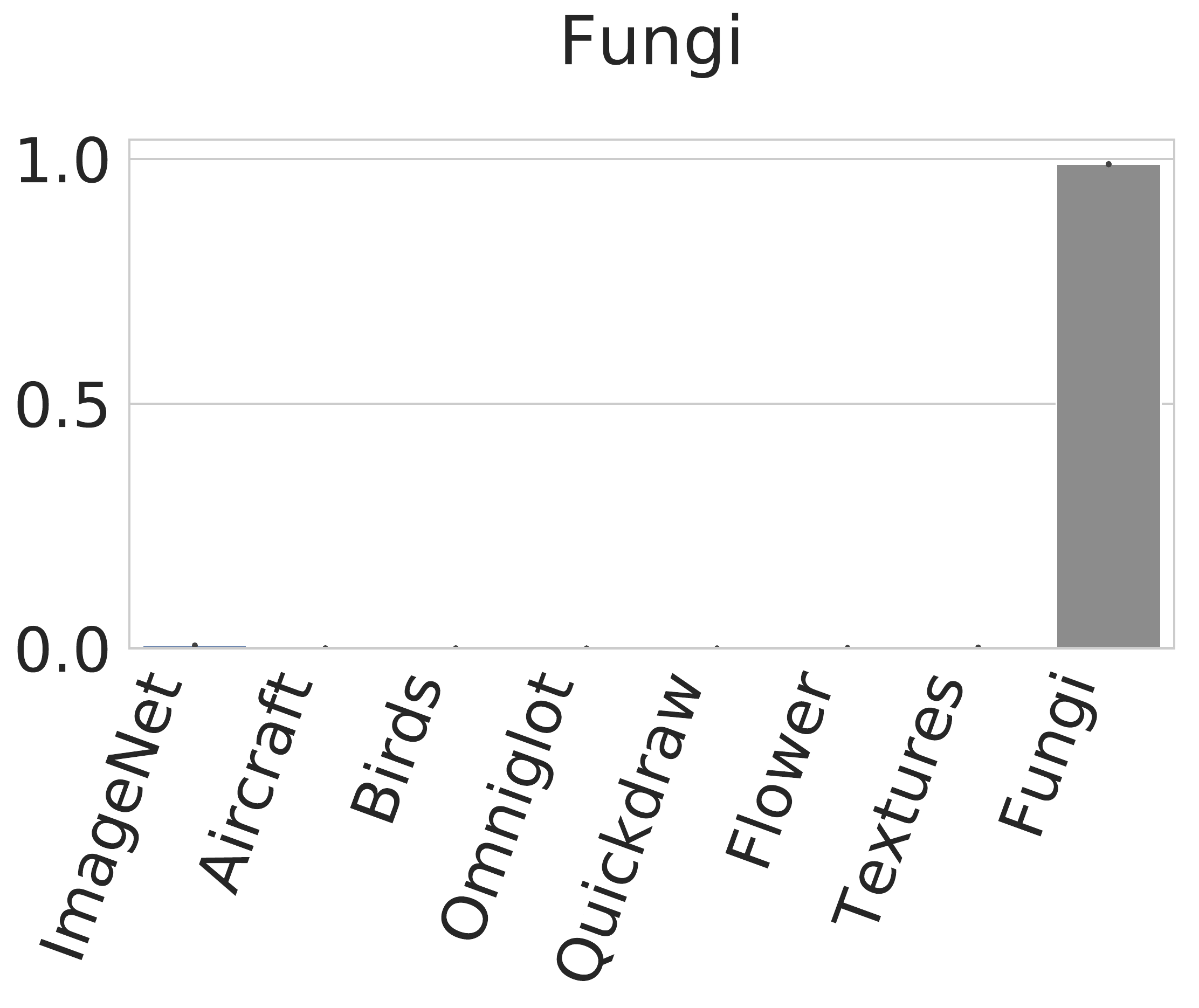}}
    \subfigure{\includegraphics[width=.3\textwidth]{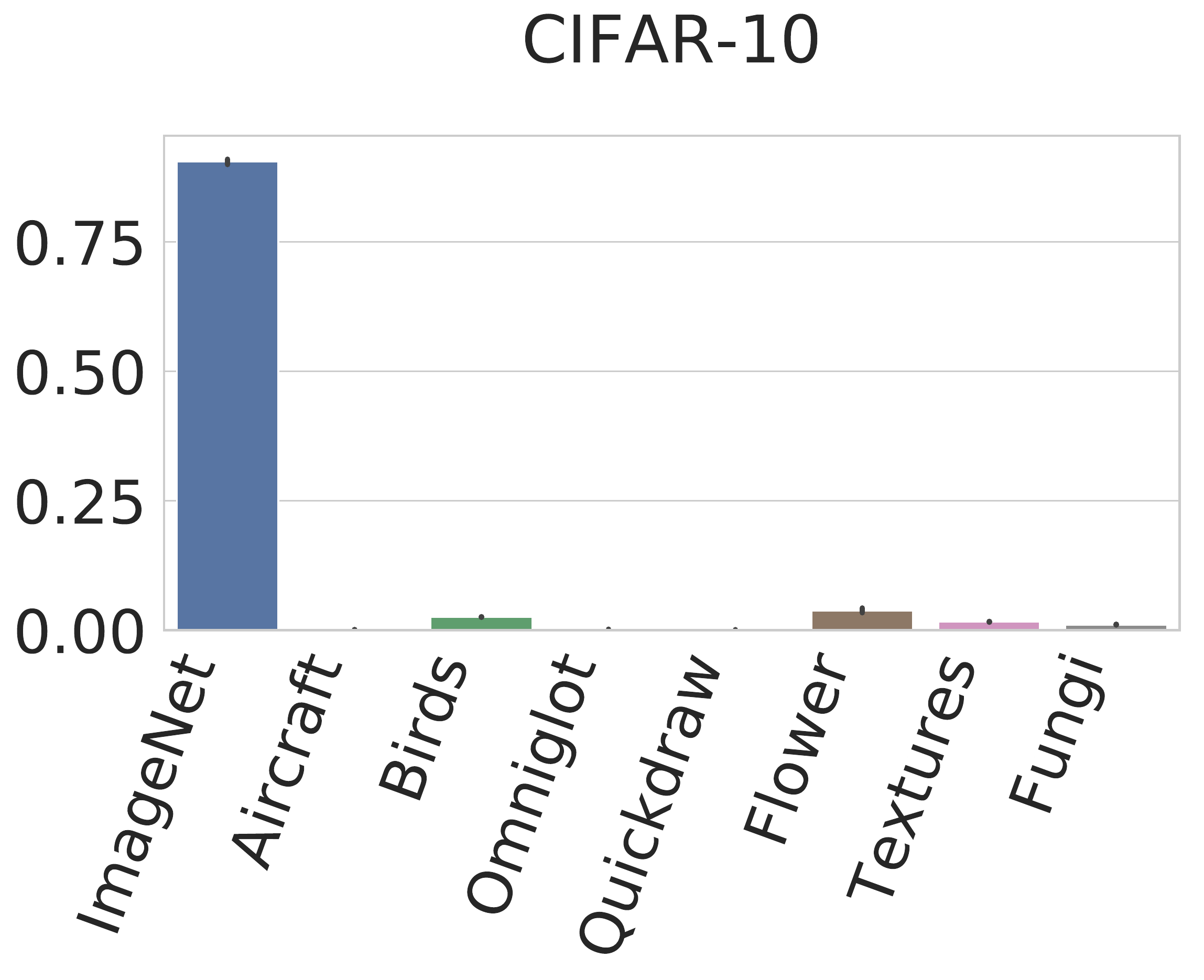}}
    \subfigure{\includegraphics[width=.3\textwidth]{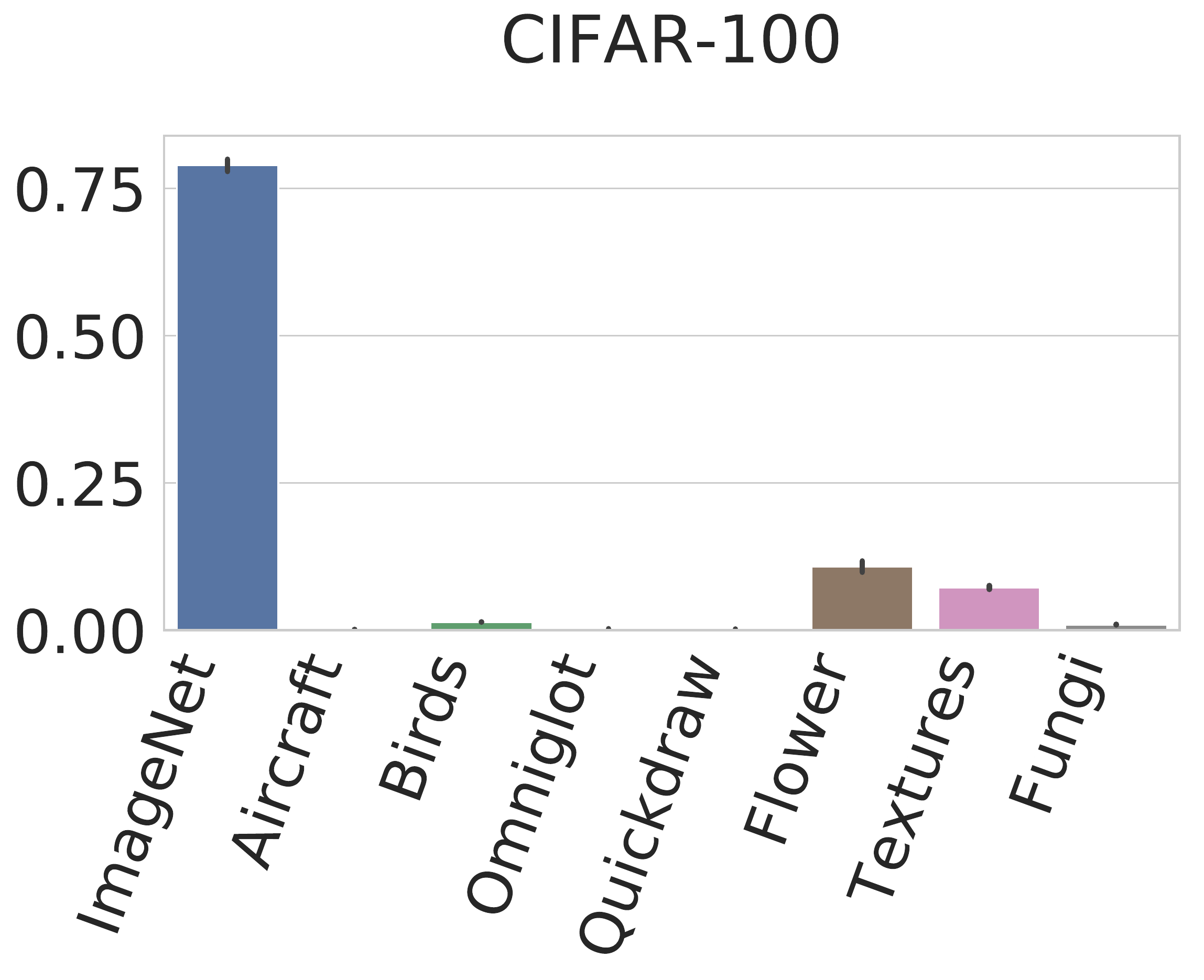}}
    \subfigure{\includegraphics[width=.3\textwidth]{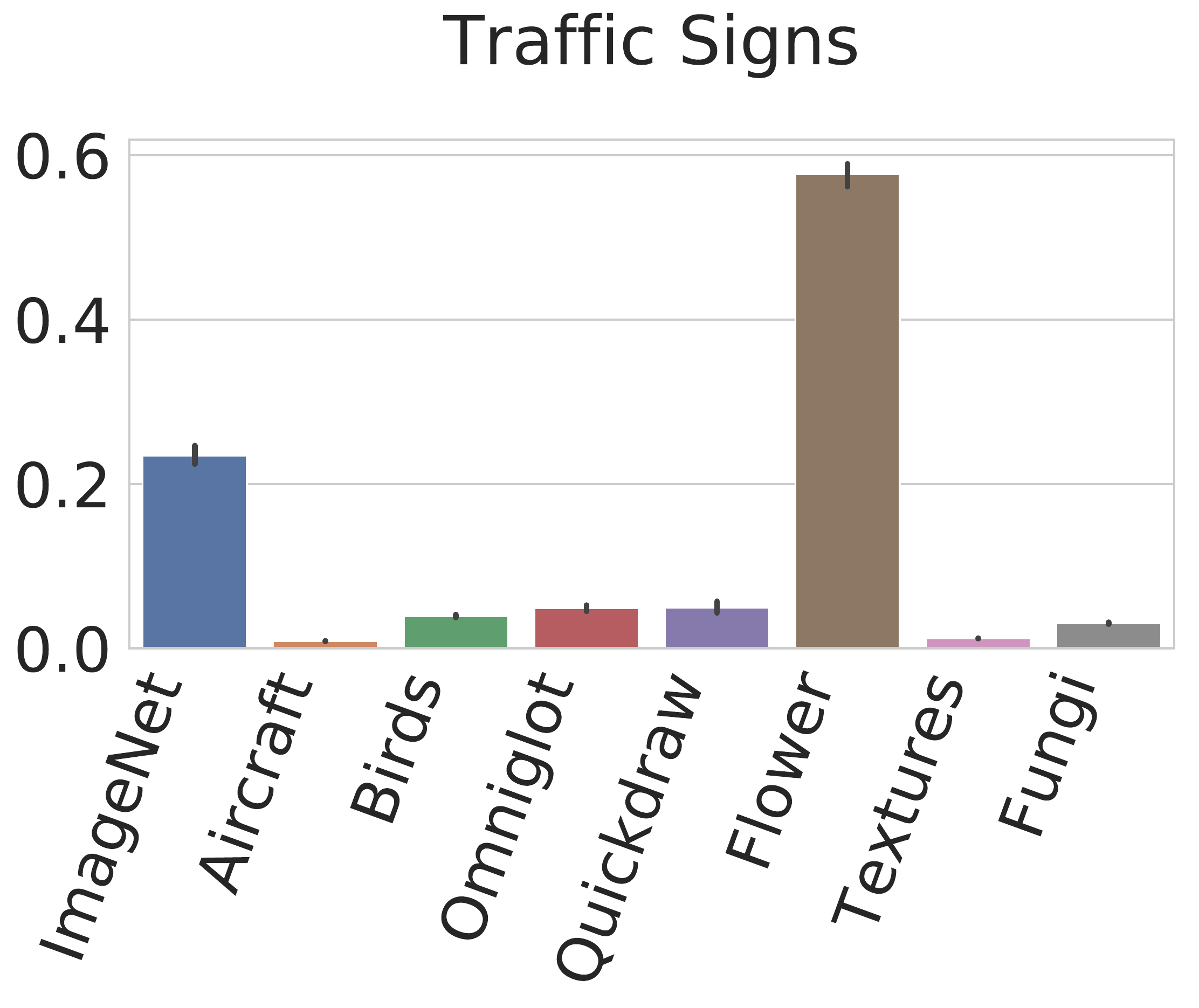}}
    \subfigure{\includegraphics[width=.3\textwidth]{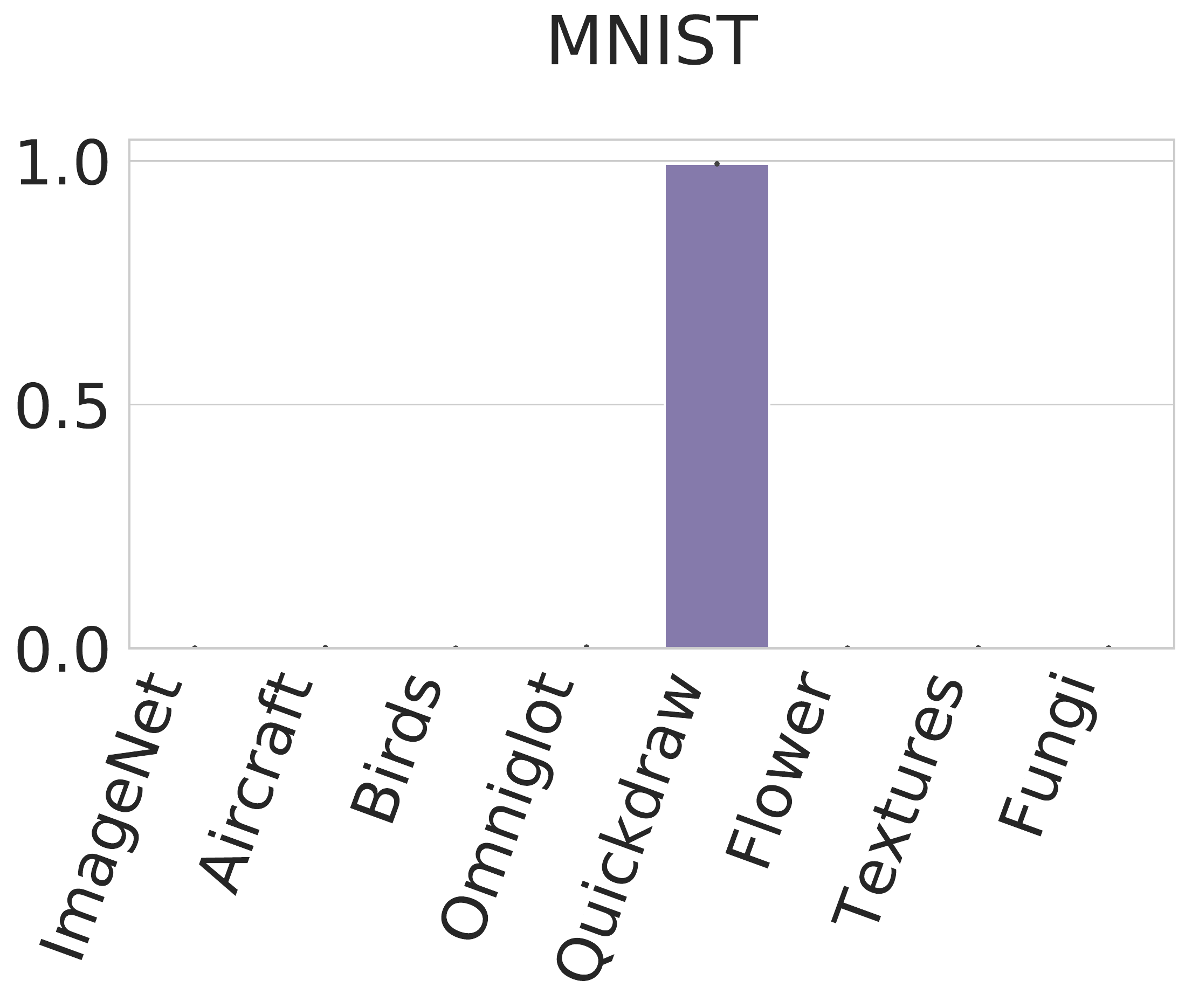}}
    \subfigure{\includegraphics[width=.3\textwidth]{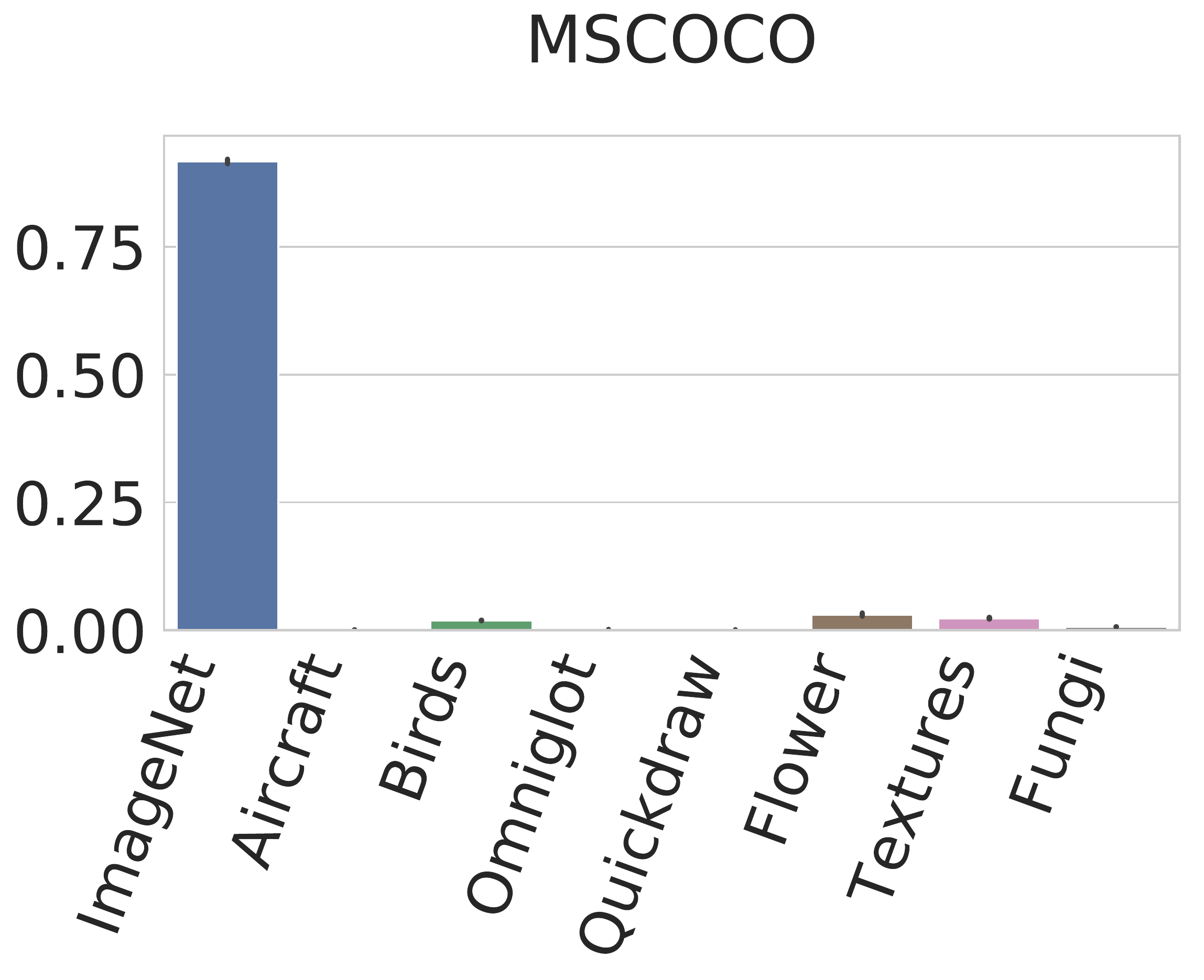}}
    \caption{The combination co-efficients that the Blender outputs for test episodes of each dataset. Each plot is creating using 600 test episodes of its corresponding dataset.}
    \label{fig:blender_within_dataset}
\end{figure*}

We visualize the Blender's proposals for 600 test tasks of each dataset in Figure~\ref{fig:blender_within_dataset}. We observe that, for each seen dataset (the first 8 sub-plots), the Blender almost exclusively picks the FiLM parameters dedicated to that specific dataset (although ImageNet and Birds sometimes pick each other to some small extent). This means that the dataset classifier is accurate across several held-out episodes of the seen datasets. For the unseen datasets, on the other hand, there is some more variability, as expected, consistent with Figure~\ref{fig:heatmap}.

\begin{figure}%
    \centering%
    \includegraphics[width=\linewidth]{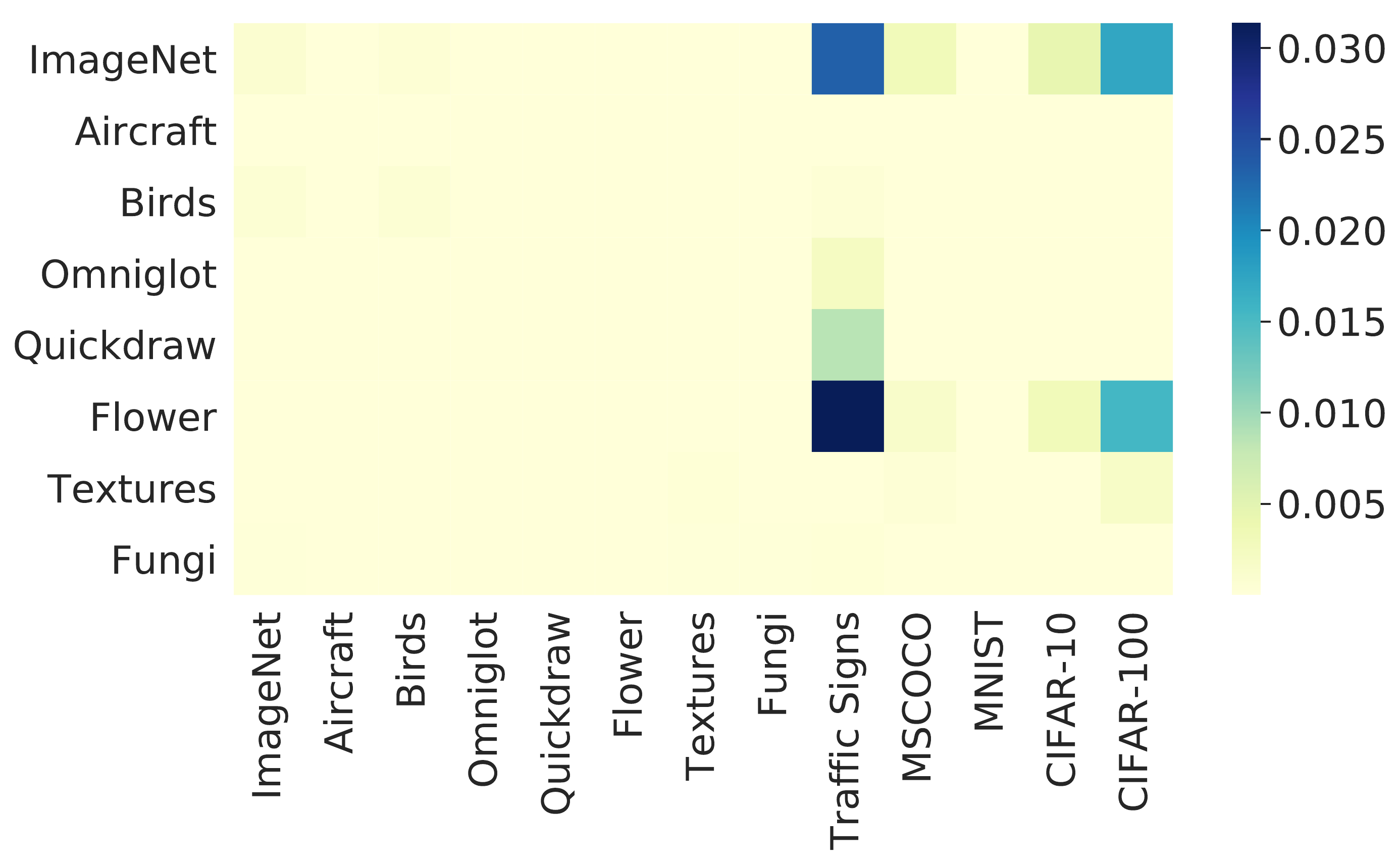}%
    \caption{\label{fig:var_heatmap} Visualization of the variance of the co-efficients that the Blender produces for each dataset over 600 test episodes of that dataset. This aids us to understand how much the dataset classifier's predictions change based on the specific support set that it ingests. As a reminder, the dataset classifier makes predictions based on the support set of each given test task, so the resulting combination co-efficients are not re-used across different tasks of the same dataset (in fact, the dataset identity is not known at test time).}%
\end{figure}

We also plot the variance of the distribution of the dataset classifier's predictions across several test tasks of each dataset, in Figure~\ref{fig:var_heatmap}. Specifically, each column corresponds to a test dataset, and the different rows show the variance of the dataset classifier's predictions over the 8 dimensions of its output vector (one for each of the $M$ training datasets). We observe that there is no variance for the first 8 columns (seen datasets) since, as expected, the dataset classifier is accurate on the seen datasets and successfully predicts the training dataset from which each support set originates from. Out of the held-out ones, we observe that MNIST also has very low variance (it always picks Quickdraw as we can see from Figure~\ref{fig:blender_within_dataset}), but the remaining held-out datasets exhibit larger variance, especially Traffic Signs where the dataset classifier's estimate of whether support sets from test episodes of Traffic Signs belong to the Flower dataset really vary from task to task.

\section*{Additional confidence intervals} We omitted the confidence intervals of Table~\ref{table:ablations_wo_ci} from the main paper due to space constraints, so we report a copy of that table along with the 95\% confidence intervals here in Table~\ref{table:ablations_with_ci}.
\begin{table*}[h]
\ra{1.2}
\caption{\label{table:ablations_with_ci} The effect of training on different data (`All', as in FLUTE, or `ImageNet-only'), and alternative initialization schemes for $\Psi_{d^*}$: from scratch (`scratch'), from ImageNet's FiLM parameters (`$\Psi_{IN}$') and from Blender. The column corresponding to the setting ``All, Blender'' is our proposed FLUTE model. This is the same Table as \ref{table:ablations_wo_ci} from the main paper, but additionally annotated with confidence intervals.}
\begin{center}
\scriptsize
\begin{tabular}{@{}l|ccc|ccc@{}}
\toprule
\multicolumn{1}{l}{Training data} & \multicolumn{3}{c}{All} & \multicolumn{3}{c}{ImageNet only} \\
\multicolumn{1}{l}{Init scheme} & Scratch & $\Psi_{IN}$ & Blender & Scratch & $\Psi_{IN}$ & Blender \\
\midrule
\DTLforeach{ablations_correct_TS}{%
  \task=Column1,%
  \avgI=Column2,%
  \ciI=Column3,%
  \avgII=Column4,%
  \ciII=Column5,%
  \avgIII=Column6,%
  \ciIII=Column7,%
  \avgIX=Column8,%
  \ciIX=Column9,%
  \avgX=Column10,%
  \ciX=Column11,%
  \avgXI=Column12,%
  \ciXI=Column13%
}{%
  \ifthenelse{\value{DTLrowi}=1}{}{%
    \ifthenelse{\value{DTLrowi}=9 \OR \value{DTLrowi}=14}{\\\midrule}{\\}%
  }%
  \dtlformat{\task}&%
  \dtlformat{\avgII} \ifthenelse{\value{DTLrowi}>13}{}{$\pm$ \dtlformat{\ciII}}&%
  \dtlformat{\avgIII} \ifthenelse{\value{DTLrowi}>13}{}{$\pm$ \dtlformat{\ciIII}}&%
  \dtlformat{\avgI} \ifthenelse{\value{DTLrowi}>13}{}{$\pm$ \dtlformat{\ciI}}&%
  \dtlformat{\avgIX} \ifthenelse{\value{DTLrowi}>13}{}{$\pm$ \dtlformat{\ciIX}}&%
  \dtlformat{\avgX} \ifthenelse{\value{DTLrowi}>13}{}{$\pm$ \dtlformat{\ciX}}&%
  \dtlformat{\avgXI} \ifthenelse{\value{DTLrowi}>13}{}{$\pm$ \dtlformat{\ciXI}}%
}
\\\bottomrule
\end{tabular}
\end{center}
\end{table*}

\section*{Additional runs of the dataset classifier network} Since we noticed in Table~\ref{table:ablations_wo_ci} of the main paper that the initialization of $\Psi_{d^*}$ has a large effect on performance, here we evaluate different checkpoints of our dataset classifier. Specifically, we performed different runs when training the dataset classifier, each with different hyperparameters, as outlined in the previous section. In what follows, we present the results not only of the top-performing one (in terms of validation accuracy), but the five top-performing ones. This allows us to understand the sensitivity of our results to the choice of the specific checkpoint of the dataset classifier that we use. We show these results in Tables~\ref{table:more_runs_init} and \ref{table:more_runs_ftune}, with and without fine-tuning of $\Psi_{d^*}$, respectively. That is, the former sets the Blender's proposal as $\Psi_{d^*}$ directly, instead of treating that as the initialization for fine-tuning via gradient descent, as we do in the latter. The last column of each table corresponds to the checkpoint of the dataset classifier that we used for our results in the main paper.

To generate the results in Table~\ref{table:more_runs_ftune}, for each of the 5 checkpoints of the dataset classifier, we performed a validation round where we used the performance on validation episodes to determine the learning rate and number of steps that will be used for fine-tuning. The hyperparameters that worked best for the 5 different checkpoints and their respective validation accuracies are shown in Table~\ref{table:ftune_chosen_hypers}. These validation accuracies are averaged over a large number of validation episodes (600 per dataset in the validation set), where as a reminder the validation set contains held-out classes of the $M$ training datasets.
\begin{table}[ht]
  \caption{\label{table:ftune_chosen_hypers} The learning rate and number of steps that were deemed best (as per the validation set accuracy) for each of the 5 checkpoints of the dataset classifier, as well as their associated validation set accuracy. These are the hyperparameters of the fine-tuning phase that were used to generate the results of Table~\ref{table:more_runs_ftune}.}
  \begin{center}
  \begin{tabular}{lccccc}
    \toprule
    & Run 1 & Run 2 & Run 3 & Run 4 & Run 5 \\
    learn rate & 1e-3  &5e-4  &5e-3 &5e-4 &5e-3\\
    num steps &     4   &  6     & 2    &   10  &6\\
    valid acc & 77.6    & 77.5  &77.5   &77.5   &78.1\\
    \bottomrule
  \end{tabular}
  \end{center}
  \end{table}
  
From Table~\ref{table:more_runs_init}, we observe that the results are reasonably consistent across the 5 checkpoints of the dataset classifier that we consider. This is especially true of the performance on (held-out classes of) the seen datasets, in rows ImageNet-Flower (weak generalization setting). On the unseen datasets in rows Traffic Signs - CIFAR-100 (strong generalization setting), there is some more variance, as expected. This is because of the fact that during training (and validation), the dataset classifier is not exposed to any data from these held-out datasets, so its behavior is underspecified in that regard, and it is plausible that different solutions perform equally well on the training and validation sets, but behave differently on the held-out datasets of the test set. Nevertheless, we find the observed variance reasonable (the difference between the best and worst performing runs is at most 0.1\% on average WG, at most 1.7\% on Average SG, and at most 0.7\% on the overall average).

Next, we look at Table~\ref{table:more_runs_ftune}, where there is an additional potential source of variance coming from the additional fine-tuning phase and the difference in the hyperparameters that were chosen for the different runs. However, we still find the observed variance reasonable (the difference between the best and worst performing runs is at most 0.2\% on average WG, at most 2.7\% on average SG, and at most 1\% on the overall average). We note that even our worst variant outperforms the previous state-of-the-art on average, and in fact with a large margin on the problem of few-shot dataset generalization that we study in this work (``Average SG''). These additional runs therefore further support FLUTE's effectiveness.

We encourage future work to also report the performance across several runs. We believe that there is an inherent underspecification in few-shot dataset generalization, due to the large gap between the training (and validation) data compared to the test data. \citep{gulrajani2020search} also offer an extensive discussion on the difficulty of model selection in the difficult regime of the domain generalization problem that they study, which is closely related to our setup. Given these difficulties, we believe it is important to report the variance of our approaches, instead of reporting only the accuracy of the top-performing run.

\begin{table*}
\ra{1.2}
\caption{\label{table:more_runs_init} The performance of FLUTE when using each of 5 different checkpoints of the dataset classifier. In this case, we omit the fine-tuning phase, and treat the Blender's proposal directly as the FiLM parameters of the new task (instead of treating that as the initialization for further fine-tuning). This allows us to more closely inspect the difference in performance induced by different dataset classifiers. As usual, we report the performance on the test set of each of the seen datasets (ImageNet - Flower) for the weak generalization setting (WG), and the performance on the held-out datasets (Traffic Signs - CIFAR-100) for the strong generalization setting (SG), corresponding to the problem of few-shot dataset generalization that we focus on in this work.}
\begin{center}
\scriptsize
\begin{tabular}{@{}lcccccccc@{}}
  \toprule
  Dataset & Run 1 & Run 2 & Run 3 & Run 4 & Run 5 \\
  \midrule
  \DTLforeach{more_runs_inits_correct_TS}{%
    \task=Column1,%
    \avgI=Column2,%
    \ciI=Column3,%
    \avgII=Column4,%
    \ciII=Column5,%
    \avgIII=Column6,%
    \ciIII=Column7,%
    \avgIX=Column8,%
    \ciIX=Column9,%
    \avgX=Column10,%
    \ciX=Column11%
  }{%
    \ifthenelse{\value{DTLrowi}=1}{}{%
      \ifthenelse{\value{DTLrowi}=9 \OR \value{DTLrowi}=14}{\\\midrule}{\\}%
    }%
    \dtlformat{\task}&%
    \dtlformat{\avgI} \ifthenelse{\value{DTLrowi}>13}{}{$\pm$ \dtlformat{\ciI}}&%
    \dtlformat{\avgII} \ifthenelse{\value{DTLrowi}>13}{}{$\pm$ \dtlformat{\ciII}}&%
    \dtlformat{\avgIII} \ifthenelse{\value{DTLrowi}>13}{}{$\pm$ \dtlformat{\ciIII}}&%
    \dtlformat{\avgIX} \ifthenelse{\value{DTLrowi}>13}{}{$\pm$ \dtlformat{\ciIX}}&%
    \dtlformat{\avgX} \ifthenelse{\value{DTLrowi}>13}{}{$\pm$ \dtlformat{\ciX}}%
  }
  \\\bottomrule
\end{tabular}
\end{center}
\end{table*}

\begin{table*}
\ra{1.2}
\caption{\label{table:more_runs_ftune} The performance of FLUTE when using each of 5 different checkpoints of the dataset classifier. Contrary to Table~\ref{table:more_runs_init}, we now perform the fine-tuning phase too that learns the FiLM parameters for the new task, starting from the Blender's proposed initialization. Table~\ref{table:ftune_chosen_hypers} shows the hyperparameters used for fine-tuning for each of the 5 runs, and their respective validation accuracies. As usual, we report the performance on the test set of each of the seen datasets (ImageNet - Flower) for the weak generalization setting (WG), and the performance on the held-out datasets (Traffic Signs - CIFAR-100) for the strong generalization setting (SG), corresponding to the problem of few-shot dataset generalization that we focus on in this work. The results for the rightmost column (Run 5) are the results we reported for FLUTE in the main paper, since this run achieved the highest validation accuracy as shown in Table~\ref{table:ftune_chosen_hypers}.}
\begin{center}
\scriptsize
\begin{tabular}{@{}lcccccccc@{}}
  \toprule
  Dataset & Run 1 & Run 2 & Run 3 & Run 4 & Run 5 \\
  \midrule
  \DTLforeach{more_runs_ftune_correct_TS}{%
    \task=Column1,%
    \avgI=Column2,%
    \ciI=Column3,%
    \avgII=Column4,%
    \ciII=Column5,%
    \avgIII=Column6,%
    \ciIII=Column7,%
    \avgIX=Column8,%
    \ciIX=Column9,%
    \avgX=Column10,%
    \ciX=Column11%
  }{%
    \ifthenelse{\value{DTLrowi}=1}{}{%
      \ifthenelse{\value{DTLrowi}=9 \OR \value{DTLrowi}=14}{\\\midrule}{\\}%
    }%
    \dtlformat{\task}&%
    \dtlformat{\avgI} \ifthenelse{\value{DTLrowi}>13}{}{$\pm$ \dtlformat{\ciI}}&%
    \dtlformat{\avgII} \ifthenelse{\value{DTLrowi}>13}{}{$\pm$ \dtlformat{\ciII}}&%
    \dtlformat{\avgIII} \ifthenelse{\value{DTLrowi}>13}{}{$\pm$ \dtlformat{\ciIII}}&%
    \dtlformat{\avgIX} \ifthenelse{\value{DTLrowi}>13}{}{$\pm$ \dtlformat{\ciIX}}&%
    \dtlformat{\avgX} \ifthenelse{\value{DTLrowi}>13}{}{$\pm$ \dtlformat{\ciX}}%
  }
  \\\bottomrule
\end{tabular}
\end{center}
\end{table*}

\section*{A closer look at the comparison between FLUTE and SUR-pf} As a reminder, SUR-pf makes the design choice of training the parametric family parameters only on ImageNet, and subsequently training a separate set of FiLM parameters for each other dataset, but without modifying the shared convolutional layer parameters. In this next experiment, we run a variant of SUR-pf that trains on all datasets in the same way as FLUTE (we re-used the parameteric family we trained for FLUTE to achieve this). The only difference between this variant and FLUTE, then, is the algorithm for tackling each test task: FLUTE will learn a new set of batch normalization parameters for the task at hand, as described in the main paper. SUR-pf, on the other hand, creates a `universal representation' by concatenating the activations of the different backbones (that share some but not all of their parameters) and applying SUR's selection mechanism to weigh the universal representation features appropriately for the task at hand \citep{dvornik2020selecting}. The results of this comparison are shown in Table~\ref{table:sur_pf_on_flute_backbone}. While our modified variant of SUR-pf outperforms the original SUR-pf, it still significantly falls short of FLUTE, especially on the strong generalization tasks. This suggests that FLUTE's superiority over SUR is not solely due to training on more data, but also due to its inductive bias that is particularly appropriate for the problem of few-shot dataset generalization.
\begin{table*}
\ra{1.2}
\caption{\label{table:sur_pf_on_flute_backbone} Comparison of FLUTE to SUR-pf and a different variant of SUR-pf that we ran (`All SUR-pf') whose parametric family is trained on all datasets, in the same way as FLUTE, instead of being trained on ImageNet only as SUR-pf is.}
\begin{center}
\scriptsize
\begin{tabular}{@{}lccc@{}}
  \toprule
  Dataset & FLUTE & SUR-pf & All SUR-pf \\
  \midrule
  \DTLforeach{compare_with_sur_correct_TS}{%
    \task=Column1,%
    \avgI=Column2,%
    \ciI=Column3,%
    \avgII=Column4,%
    \ciII=Column5,%
    \avgIII=Column6,%
    \ciIII=Column7%
  }{%
    \ifthenelse{\value{DTLrowi}=1}{}{%
      \ifthenelse{\value{DTLrowi}=9 \OR \value{DTLrowi}=14}{\\\midrule}{\\}%
    }%
    \dtlformat{\task}&%
    \dtlformat{\avgI} \ifthenelse{\value{DTLrowi}>13}{}{$\pm$ \dtlformat{\ciI}}&%
    \dtlformat{\avgII} \ifthenelse{\value{DTLrowi}>13}{}{$\pm$ \dtlformat{\ciII}}&%
    \dtlformat{\avgIII} \ifthenelse{\value{DTLrowi}>13}{}{$\pm$ \dtlformat{\ciIII}}%
  }
  \\\bottomrule
\end{tabular}
\end{center}
\end{table*}

\section*{Shuffled Traffic Signs}
It was recently noticed \footnote{https://github.com/google-research/meta-dataset/issues/54} that in the introduction notebook that comes with the Meta-Dataset code-base\footnote{https://github.com/google-research/meta-dataset}, the usage examples given for the episode input pipeline did not set the parameter that dictates the size of the shuffle buffer, which defaults to not shuffling examples within each class. This led to many previous works on Meta-Dataset using unshuffled datasets, which evidently produced more optimistic results on the Traffic Signs dataset. Specifically, the examples of this dataset are organized as 30-image sequences of pictures from the same physical sign (successive frames from the same video), leading to support and query examples being more frequently really close when not shuffling the examples of each class.

The results we reported in this paper are computed as intended, using the shuffled datasets. For reference, there is also a leaderboard on the Meta-Dataset code-base repository that reflects the shuffled Traffic Signs numbers for different methods. 

However, for completeness, we show here the results computed on the easier variant induced by not shuffling the images. These are in Table~\ref{table:traffic_sings_unshuffled_all} (main results) and Table~\ref{table:traffic_sings_unshuffled} (additional runs of the dataset classifier). The latter displays the results of the same 5 checkpoints as we used in the previous section. The last column represents the model that we used to report FLUTE's results in the main paper.
\begin{table*}
\ra{1.2}
\caption{\label{table:traffic_sings_unshuffled_all} Comparing FLUTE to recent state-of-the-art methods. This is the same table as Table~\ref{table:main} in the main paper, with the exception of the Traffic Signs row that now reflects the easier (unshuffled) variant of Traffic Signs.}
\begin{center}
\scriptsize
\begin{tabular}{@{}lcccccccc@{}}
  \toprule
  Dataset & CNAPs & TaskNorm & SimpleCNAPs & SUR-pf & URT-pf & SUR (x8) & URT (x8) & FLUTE \\
  \midrule
  \DTLforeach{all}{%
    \task=Column1,%
    \avgI=Column2,%
    \ciI=Column3,%
    \avgII=Column4,%
    \ciII=Column5,%
    \avgIII=Column6,%
    \ciIII=Column7,%
    \avgIX=Column8,%
    \ciIX=Column9,%
    \avgX=Column10,%
    \ciX=Column11,%
    \avgXI=Column12,%
    \ciXI=Column13,%
    \avgXII=Column14,%
    \ciXII=Column15,%
    \avgXIII=Column16,%
    \ciXIII=Column17%
  }{%
    \ifthenelse{\value{DTLrowi}=1}{}{%
      \ifthenelse{\value{DTLrowi}=9 \OR \value{DTLrowi}=14}{\\\midrule}{\\}%
    }%
    \dtlformat{\task}&%
    \dtlformat{\avgI} \ifthenelse{\value{DTLrowi}>13}{}{$\pm$ \dtlformat{\ciI}}\%&%
    \dtlformat{\avgII} \ifthenelse{\value{DTLrowi}>13}{}{$\pm$ \dtlformat{\ciII}}\%&%
    \dtlformat{\avgIII} \ifthenelse{\value{DTLrowi}>13}{}{$\pm$ \dtlformat{\ciIII}}\%&%
    \dtlformat{\avgIX} \ifthenelse{\value{DTLrowi}>13}{}{$\pm$ \dtlformat{\ciIX}}\%&%
    \dtlformat{\avgX} \ifthenelse{\value{DTLrowi}>13}{}{$\pm$ \dtlformat{\ciX}}\%&%
    \dtlformat{\avgXI} \ifthenelse{\value{DTLrowi}>13}{}{$\pm$ \dtlformat{\ciXI}}\%&%
    \dtlformat{\avgXII} \ifthenelse{\value{DTLrowi}>13}{}{$\pm$ \dtlformat{\ciXII}}\%&%
    \dtlformat{\avgXIII} \ifthenelse{\value{DTLrowi}>13}{}{$\pm$ \dtlformat{\ciXIII}}\%%
  }
  \\\bottomrule
\end{tabular}
\end{center}
\end{table*}

\begin{table*}
\ra{1.2}
\caption{\label{table:traffic_sings_unshuffled} The performance of FLUTE when using each of 5 different checkpoints of the dataset classifier on the (easier) unshuffled version of the Traffic Signs dataset. The results for the rightmost column (Run 5) are the results produced by the FLUTE variant that we report in the main paper.}
\begin{center}
\scriptsize
\begin{tabular}{@{}lccccc@{}}
  \toprule
  Dataset & Run 1 & Run 2 & Run 3 & Run 4 & Run 5 \\
  \midrule
  \DTLforeach{traffic_signs_unshuffled}{%
    \task=Column1,%
    \avgI=Column2,%
    \ciI=Column3,%
    \avgII=Column4,%
    \ciII=Column5,%
    \avgIII=Column6,%
    \ciIII=Column7,%
    \avgIX=Column8,%
    \ciIX=Column9,%
    \avgX=Column10,%
    \ciX=Column11%
  }{%
    \ifthenelse{\value{DTLrowi}=1}{}{%
      \ifthenelse{\value{DTLrowi}=9 \OR \value{DTLrowi}=14}{\\\midrule}{\\}%
    }%
    \dtlformat{\task}&%
    \dtlformat{\avgI} \ifthenelse{\value{DTLrowi}>13}{}{$\pm$ \dtlformat{\ciI}}\%&%
    \dtlformat{\avgII} \ifthenelse{\value{DTLrowi}>13}{}{$\pm$ \dtlformat{\ciII}}\%&%
    \dtlformat{\avgIII} \ifthenelse{\value{DTLrowi}>13}{}{$\pm$ \dtlformat{\ciIII}}\%&%
    \dtlformat{\avgIX} \ifthenelse{\value{DTLrowi}>13}{}{$\pm$ \dtlformat{\ciIX}}\%&%
    \dtlformat{\avgX} \ifthenelse{\value{DTLrowi}>13}{}{$\pm$ \dtlformat{\ciX}}\%%
  }
  \\\bottomrule
\end{tabular}
\end{center}
\end{table*}

\section*{Hard Blender: using the dataset classifier but without taking a convex combination} An alternative design choice is to use a `Hard' Blender, that instead of taking a convex combination of the training datasets' FiLM parameters, selects only the FiLM parameters of the most likely training dataset (as assessed by the dataset classifier). The comparison with this variant is shown in Table~\ref{table:hard_blender_comparison}. Perhaps unsurprisingly, the two initialization schemes perform similarly. This is expected, especially for the WG tasks, since the Blender, which is based on an accurate dataset classifier, puts most of its probability mass on a single dataset anyway. Taking the convex combination is a more general approach that doesn't suffer, and in fact may slightly be beneficial for some SG tasks. We therefore adopt this initialization scheme as our default one for use with FLUTE.
\begin{table*}[ht]
\ra{1.2}
\caption{\label{table:hard_blender_comparison} Comparing the Blender initialization scheme to the `Hard Blender' variant. Specifically, instead of taking a convex combination of the training datasets' FiLM parameters as Blender does, `Hard Blender' selects only the FiLM parameters of the most likely training dataset (as assessed by the dataset classifier). The columns marked as `fine-tune' train the FiLM parameters using gradient descent from the Blender or Hard Blender initialization, whereas the others use the initialization directly, allowing to more closely inspect the difference between these two initialization schemes.}
\begin{center}
\scriptsize
\begin{tabular}{@{}lcccc@{}}
  \toprule
  Dataset & Blender & Blender (fine-tune) & Hard Blender & Hard Blender (fine-tune) \\
  \midrule
  \DTLforeach{hard_blender_correct_TS}{%
    \task=Column1,%
    \avgI=Column2,%
    \ciI=Column3,%
    \avgII=Column4,%
    \ciII=Column5,%
    \avgIII=Column6,%
    \ciIII=Column7,%
    \avgIX=Column8,%
    \ciIX=Column9%
  }{%
    \ifthenelse{\value{DTLrowi}=1}{}{%
      \ifthenelse{\value{DTLrowi}=9 \OR \value{DTLrowi}=14}{\\\midrule}{\\}%
    }%
    \dtlformat{\task}&%
    \dtlformat{\avgI} \ifthenelse{\value{DTLrowi}>13}{}{$\pm$ \dtlformat{\ciI}}&%
    \dtlformat{\avgII} \ifthenelse{\value{DTLrowi}>13}{}{$\pm$ \dtlformat{\ciII}}&%
    \dtlformat{\avgIII} \ifthenelse{\value{DTLrowi}>13}{}{$\pm$ \dtlformat{\ciIII}}&%
    \dtlformat{\avgIX} \ifthenelse{\value{DTLrowi}>13}{}{$\pm$ \dtlformat{\ciIX}}%
  }
  \\\bottomrule
\end{tabular}
\end{center}
\end{table*}

\end{document}


\twocolumn[
  \icmltitle{\titleprefix Learning a Universal Template for Few-shot Dataset Generalization}
  \icmlsetsymbol{equal}{*}
  
  \begin{icmlauthorlist}
  \icmlauthor{Eleni Triantafillou}{vector,google,atGoogle}
  \icmlauthor{Hugo Larochelle}{google}
  \icmlauthor{Richard Zemel}{vector}
\icmlauthor{Vincent Dumoulin}{google}
  \end{icmlauthorlist}
  
  \icmlaffiliation{google}{Google Research, Brain Team}
  \icmlaffiliation{vector}{University of Toronto, Vector Institute}
  \icmlaffiliation{atGoogle}{Work done at Google.}
  
  \icmlcorrespondingauthor{Eleni Triantafillou}{eleni@cs.toronto.edu}
  
  \icmlkeywords{Machine Learning, ICML, Few-Shot Classification}
  
  \vskip 0.3in
]

\printAffiliationsAndNotice{}

\section{Detailed Accuracies}

\autoref{table:ablations_wo_ci}

{\onecolumn
  \begin{table}[ht]
  \ra{1.2}
  \caption{\label{table:example_table} Example table.}
  \begin{center}
  \begin{tabular}{@{}lcc@{}}
    \toprule
    Data source & Method 1 & Method 2 \\
    \midrule
    \DTLforeach{example_table}{%
      \task=Column1,%
      \avgI=Column2,%
      \stdI=Column3,%
      \avgII=Column4,%
      \stdII=Column5%
    }{%
      \ifthenelse{\value{DTLrowi}=1}{}{%
        \ifthenelse{\value{DTLrowi}=9 \OR \value{DTLrowi}=14}{\\\midrule}{\\}%
      }%
      \dtlformat{\task}&%
      \dtlformat{\avgI} \ifthenelse{\value{DTLrowi}>13}{}{$\pm$ \dtlformat{\stdI}}\%&%
      \dtlformat{\avgII} \ifthenelse{\value{DTLrowi}>13}{}{$\pm$ \dtlformat{\stdII}}\%%
    }
    \\\bottomrule
  \end{tabular}
  \end{center}
  \end{table}
}